\documentclass{article}

\usepackage{amsfonts}
\usepackage{amsmath}
\usepackage{booktabs}
\usepackage{arydshln}
\usepackage{hyperref}
\usepackage{url}
\usepackage{graphicx}
\usepackage{multirow}
\usepackage{multicol}
\usepackage{placeins}
\usepackage{subfigure}

\usepackage{color,soul}
\usepackage{xr}

\DeclareMathOperator*{\argmin}{arg\,min}

\newcommand{\STAB}[1]{\begin{tabular}{@{}c@{}}#1\end{tabular}}

\newcommand{\squishlist}{
 \begin{list}{$\bullet$}
  { \setlength{\itemsep}{0pt}
     \setlength{\parsep}{3pt}
     \setlength{\topsep}{3pt}
     \setlength{\partopsep}{0pt}
     \setlength{\leftmargin}{1.5em}
     \setlength{\labelwidth}{1em}
     \setlength{\labelsep}{0.5em} } }

\newcommand{\squishlisttwo}{
 \begin{list}{$\bullet$}
  { \setlength{\itemsep}{0pt}
     \setlength{\parsep}{0pt}
    \setlength{\topsep}{0pt}
    \setlength{\partopsep}{0pt}
    \setlength{\leftmargin}{2em}
    \setlength{\labelwidth}{1.5em}
    \setlength{\labelsep}{0.5em} } }

\newcommand{\squishend}{
  \end{list}  }

\usepackage{hyperref}

\usepackage[accepted]{icml2020}

\icmltitlerunning{NADS: Neural Architecture Distribution Search for Uncertainty Awareness}

\begin{document}

\twocolumn[
\icmltitle{NADS: Neural Architecture Distribution Search for Uncertainty Awareness}



\icmlsetsymbol{equal}{*}

\begin{icmlauthorlist}
\icmlauthor{Randy Ardywibowo}{to}
\icmlauthor{Shahin Boluki}{to}
\icmlauthor{Xinyu Gong}{goo}
\icmlauthor{Zhangyang Wang}{goo}
\icmlauthor{Xiaoning Qian}{to}
\end{icmlauthorlist}

\icmlaffiliation{to}{Department of Electrical and Computer Engineering, Texas A\&M University, College Station, Texas, USA}
\icmlaffiliation{goo}{Department of Computer Science and Engineering, Texas A\&M University, College Station, Texas, USA}

\icmlcorrespondingauthor{Randy Ardywibowo}{randyardywibowo@tamu.edu}

\icmlkeywords{Machine Learning, ICML}

\vskip 0.3in
]



\printAffiliationsAndNotice{}  

\begin{abstract}
Machine learning (ML) systems often encounter Out-of-Distribution (OoD) errors when dealing with testing data coming from a distribution different from training data. It becomes important for ML systems in critical applications to accurately quantify its predictive uncertainty and screen out these anomalous inputs. However, existing OoD detection approaches are prone to errors and even sometimes assign higher likelihoods to OoD samples. Unlike standard learning tasks, there is currently no well established guiding principle for designing OoD detection architectures that can accurately quantify uncertainty. To address these problems, we first seek to identify guiding principles for designing uncertainty-aware architectures, by proposing \textit{Neural Architecture Distribution Search} (NADS). NADS searches for a distribution of architectures that perform well on a given task, allowing us to identify common building blocks among all uncertainty-aware architectures. With this formulation, we are able to optimize a stochastic OoD detection objective and construct an ensemble of models to perform OoD detection. We perform multiple OoD detection experiments and observe that our NADS performs favorably, with up to 57\% improvement in accuracy compared to state-of-the-art methods among 15 different testing configurations.
\end{abstract}

\section{Introduction}

Detecting anomalous data is crucial for safely applying machine learning in autonomous systems for critical applications and for AI safety \citep{amodei2016concrete}. Such anomalous data can come in settings such as in autonomous driving \citep{kendall2017uncertainties,autonomous}, disease monitoring~\citep{hendrycks2016baseline,ardywibowo2019adaptive,ardywibowo2018switching}, and fault detection \citep{hendrycks2018deep}. In these situations, it is important for these systems to reliably detect abnormal inputs so that their occurrence can be overseen by a human, or the system can proceed using a more conservative policy.

The widespread use of deep learning models within these autonomous systems have aggravated this issue. Despite having high performance in many predictive tasks, deep networks tend to give high confidence predictions on Out-of-Distribution (OoD) data \citep{goodfellow2014explaining,nguyen2015deep}. Moreover, commonly used OoD detection approaches are prone to errors and even assign higher likelihoods to samples from other datasets \citep{lee2018training,hendrycks2016baseline}. 

Unlike common machine learning tasks such as image classification, segmentation, and speech recognition, there are currently no well established guidelines for designing architectures that can accurately screen out OoD data and quantify its predictive uncertainty. Such a gap makes Neural Architecture Search (NAS) a promising option to explore the better design of uncertainty-aware models~\citep{elsken2018neural}. NAS algorithms attempt to find an optimal neural network architecture for a specific task. Existing efforts have primarily focused on searching for architectures that perform well on image classification or segmentation. However, it is unclear whether architecture components that are beneficial for image classification and segmentation models would also lead to better uncertainty quantification~(UQ) and thereafter be effective for OoD detection. 

Because of this, it is necessary to tailor the search objective in order to find the  architectures that can accurately detect OoD data. However, designing an optimization objective that leads to uncertainty-aware models is also not straightforward. With no access to labels for OoD data, unsupervised/self-supervised generative models maximizing the likelihood of in-distribution data become the primary tools for UQ \citep{hendrycks2019using}. However, these models counter-intuitively assign high likelihoods to OoD data~\citep{nalisnick2018do, choi2018generative, hendrycks2019using, shafaeiless}. Because of this, maximizing the log-likelihood is inadequate for OoD detection. On the other hand, \citet{choi2018generative} proposed using the Widely Applicable Information Criterion (WAIC) \citep{watanabe2013widely}, a penalized likelihood score, as the OoD detection criterion, showing that it was robust for OoD detection. However, the score was approximated using an ensemble of models that was trained on maximizing the likelihood and did not directly optimize the WAIC score. In line with this, previous work on deep uncertainty quantification show that ensembles can help calibrate OoD classifier based methods, as well as improve OoD detection performance of likelihood estimation models \citep{lakshminarayanan2017simple}. Based on these findings, one might consider finding a distribution of well-performing architectures for uncertainty awareness, instead of searching for a single best performing architecture, as is typically done in existing NAS methods.


To this end, we propose \textit{Neural Architecture Distribution Search} (\textbf{NADS}) to identify common building blocks that naturally incorporate model uncertainty quantification and compose good OoD detection models. NADS searches for a \textbf{distribution} of well-performing architectures, instead of a single best architecture, by formulating the architecture search problem as a stochastic optimization problem. We optimize the WAIC score of the architecture distribution, a score that was shown to be robust towards estimating model uncertainty. By taking advantage of weight sharing between different architectures, as well as through a particular parameterization of the architecture distribution, the discrete search problem for NADS can be efficiently solved by a continuous relaxation \cite{xie2018snas, chang2019differentiable}. Using the learned posterior architecture distribution, we construct a Bayesian ensemble of deep models to perform OoD detection, demonstrating state-of-the-art performance in multiple OoD detection experiments. Specifically, our main contributions with NADS include:
\squishlist
  \item NADS learns a posterior distribution on the architecture search space to enable UQ for better OoD detection, instead of providing a maximum-likelihood point estimate to the best model. 
  \item We design a novel generative search space that is inspired by Glow \cite{kingma2018glow}, which is different from previous NAS methods.
  \item We use the WAIC score as the reward to guide the architecture search and provide a method to estimate this score for architecture search.
  \item NADS yields state-of-the-art performance in multiple OoD detection experiments, making likelihood estimation based OoD detection competitive against multi-class classifier based approaches. Notably, our method yields consistent improvements in accuracy among 15 different in-distribution -- out-of-distribution testing pairs, with an improvement of up to 57\% accuracy against existing state-of-the-art methods.
\squishend

\section{Background}

\subsection{Neural Architecture Search}

Neural Architecture Search (NAS) algorithms aim to automatically discover an optimal neural network architecture instead of using a hand-crafted one for a specific task. Previous work on NAS has achieved successes in image classification \citep{pham2018efficient}, image segmentation \citep{liu2019auto}, object detection~\citep{ghiasi2019fpn}, structured prediction \citep{chen2018searching}, and generative adversarial networks \citep{Gong_2019_ICCV}. However, there has been no NAS algorithm developed for uncertainty quantificaton and OoD detection.

NAS consists of three components: the proxy task, the search space, and the optimization algorithm. Prior work in specifying the search space either searches for an entire architecture directly, or searches for small cells and arrange them in a pre-defined way. Optimization algorithms that have been used for NAS include reinforcement learning \citep{baker2016designing, zoph2018learning, zhong2018practical, zoph2016neural}, Bayesian optimization \citep{jin2018auto}, random search \citep{chen2018searching}, Monte Carlo tree search \citep{negrinho2017deeparchitect}, and gradient-based optimization methods \citep{liu2018darts,ahmed2018maskconnect, xie2018snas, chang2019differentiable}. To efficiently evaluate the performance of discovered architectures and guide the search, the design of the proxy task is critical. Existing proxy tasks include leveraging shared parameters \citep{pham2018efficient}, predicting performance using a surrogate model \citep{liu2018progressive}, and early stopping \citep{zoph2018learning,chen2018searching}.

To the best of our knowledge, all existing NAS algorithms seek a single best performing architecture. In comparison, searching for a distribution of architectures allows us to analyze the common building blocks that all of the candidate architectures have. Moreover, this technique can also complement ensemble methods by creating a more diverse set of models tailored to optimize the ensemble objective, an important ingredient for deep uncertainty quantification \citep{lakshminarayanan2017simple, choi2018generative}. 

\subsection{Uncertainty Quantification and OoD Detection}

Prior work on uncertainty quantification and OoD detection for deep models can be divided into model-dependent \citep{lakshminarayanan2017simple, gal2016dropout,boluki2020learnable,liang2017enhancing}, and model-independent techniques \citep{dinh2016density, germain2015made, oord2016wavenet}. Model-dependent techniques aim to yield confidence measures $p(y | \boldsymbol{x})$ for a model's prediction $y$ when given input data $\boldsymbol{x}$. However, a limitation of model-dependent OoD detection is that they may discard information regarding the data distribution $p(\boldsymbol{x})$ when learning the task specific model $p(y | \boldsymbol{x})$. This could happen when certain features of the data are irrelevant for the predictive task, causing information loss regarding the data distribution $p(\boldsymbol{x})$.  Moreover, existing methods to calibrate model uncertainty estimates assume access to OoD data during training \citep{lee2018training,hendrycks2018deep}. Although the OoD data may not come from the testing distribution, this approach assumes that the structure of OoD data is known ahead of time, which can be incorrect in settings such as active/online learning where new training distributions are regularly encountered.



On the other hand, model-independent techniques seek to estimate the likelihood of the data distribution $p(\boldsymbol{x})$. These techniques include Variational Autoencoders (VAEs) \citep{kingma2013auto}, Generative Adversarial Networks (GANs) \citep{goodfellow2014generative}, autoregressive models \citep{germain2015made,oord2016wavenet},  and invertible flow-based models \citep{dinh2016density,kingma2018glow}. Among these techniques, invertible models offer exact computation of the data likelihood, making them attractive for likelihood estimation. Moreover, they do not require OoD samples during training, making them applicable to any OoD detection scenario. Thus in this paper, we focus on searching for invertible flow-based architectures, though the presented techniques are also applicable to other likelihood estimation models.

Along this direction, recent work has discovered that likelihood-based models can assign higher likelihoods to OoD data compared to in-distribution data \citep{nalisnick2018do, choi2018generative} (see Figure~\ref{celeba_motivating} of the supplementary material for an example). One hypothesis for such a phenomenon is that most data points lie within the typical set of a distribution, instead of the region of high likelihood \citep{nalisnick2019detecting}. Thus, \citet{nalisnick2019detecting} recommend to estimate the entropy using multiple data samples to screen out OoD data instead of using the likelihood. Other uncertainty quantification formulations can also be related to entropy estimation \citep{choi2018generative, lakshminarayanan2017simple}. However, it is not always realistic to test multiple data points in practical data streams, as testing data often come one sample at a time and are never well-organized into in-distribution or out-of-distribution groups.

With this in mind, model ensembling becomes a natural consideration to formulate entropy estimation. Instead of computing the entropy by averaging over multiple data points, model ensembles produce multiple estimates of the data likelihood, thus ``augmenting'' one data point into as many data points as needed to reliably estimate the entropy. However, care must be taken to ensure that the model ensemble produces likelihood estimates that agree with one another on in-distribution data, while also being diverse enough to discriminate OoD data likelihoods. 

In what follows, we propose NADS as a method that can identify distributions of architectures for uncertainty quantification. Using a loss function that accounts for the diversity of architectures within the distribution, NADS allows us to construct an ensemble of models that can reliably detect OoD data.

\section{Neural Architecture Distribution Search}

\begin{figure*}[t!]
  \centering
  \includegraphics[width=0.72\textwidth]{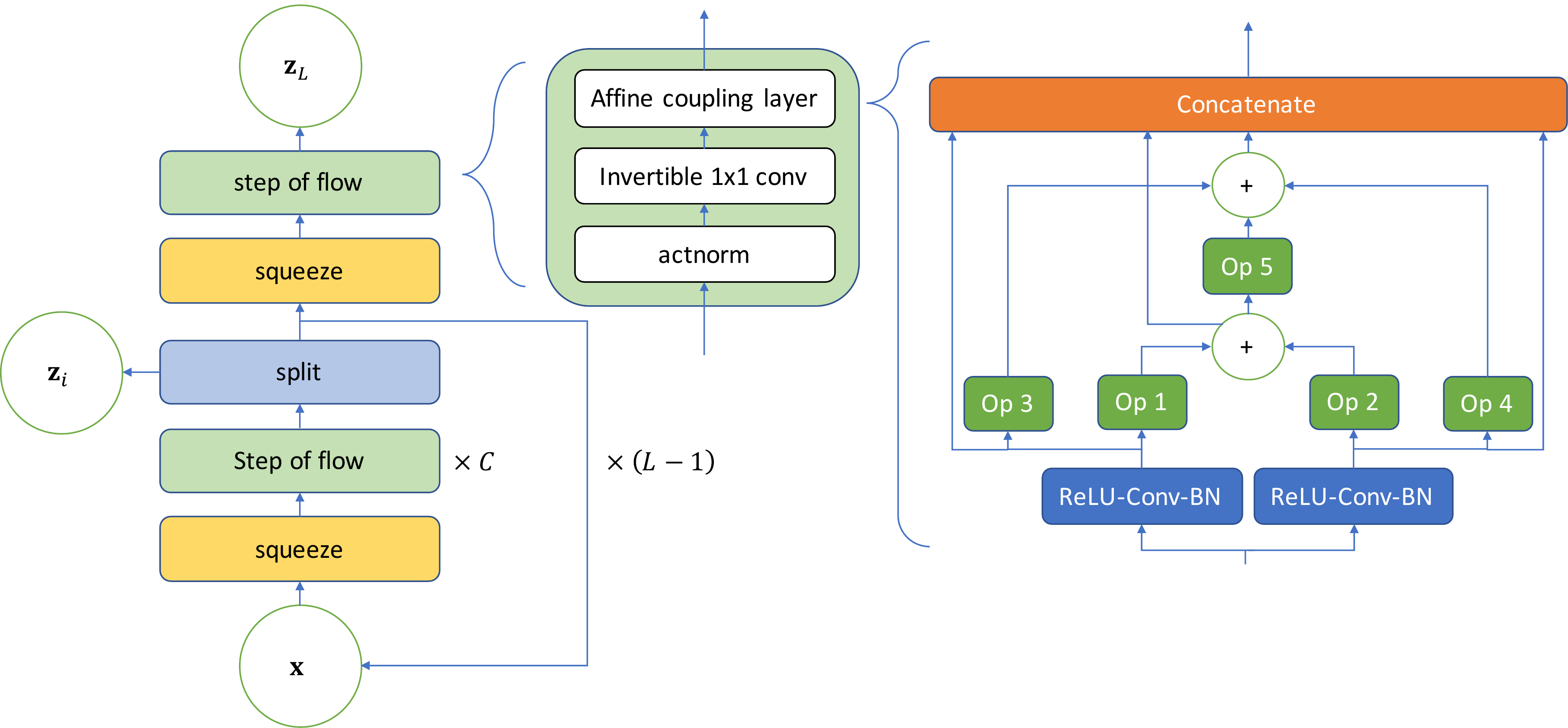}
  \caption{Search space of a single block in the architecture}
  \label{nasops}
\end{figure*}

Putting Neural Architecture Distribution Search~(NADS) under a common NAS framework~\citep{elsken2018neural}, we break down our search formulation into three main components: the proxy task, the search space, and the optimization method. Specifying these components for NADS with the ultimate goal of uncertainty quantification for OoD detection is not immediately obvious. For example, naively using data likelihood maximization as a proxy task would run into the issues pointed out by \citet{nalisnick2018do}, with models assigning higher likelihoods to OoD data. On the other hand, the search space needs to be large enough to include a diverse range of architectures, yet still allowing a search algorithm to traverse it efficiently. In the following sections, we motivate our decision on these three choices and describe these components for NADS in detail. 






\subsection{Proxy Task}

The first component of NADS is the training objective that guides the neural architecture search. Different from existing NAS methods, our aim is to derive an ensemble of deep models to improve model uncertainty quantification and OoD detection. To this end, instead of searching for architectures that maximize the likelihood of in-distribution data, which tends to cause models to incorrectly assign high likelihoods to OoD data, we instead seek architectures that can perform entropy estimation by maximizing the Widely Applicable Information Criteria~(WAIC) of the training data. The WAIC score is a Bayesian adjusted metric to calculate the marginal likelihood \citep{watanabe2013widely}. This metric has been shown by \citet{choi2018generative} to be robust towards the pitfall causing likelihood estimation models to assign high likelihoods to OoD data. The score is defined as follows:
\begin{align}
\begin{split}
    \text{WAIC} & (X) = \mathbb{E}_{\alpha \sim p(\alpha) }\bigg[ \mathbb{E}_{p(\boldsymbol{x})}[\log p(\boldsymbol{x} | \alpha)] \bigg] \\ & - \mathbb{V}_{\alpha \sim p(\alpha)} \bigg[ \mathbb{E}_{p(\boldsymbol{x})}[\log p(\boldsymbol{x} | \alpha)] \bigg]
\label{WAIC_equation}
\end{split}
\end{align}

Here, $\mathbb{E}[\cdot]$ and $\mathbb{V}[\cdot]$ denote expectation and variance respectively, which are taken over all architectures $\alpha$ sampled from the posterior architecture distribution $p(\alpha)$. Such a strategy captures model uncertainty in a Bayesian fashion, improving OoD detection while also converging to the true data likelihood as the number of data points increases \cite{gelman2014understanding}. Intuitively, minimizing the variance of training data likelihoods allows its likelihood distribution to remain tight which, by proxy, minimizes the overlap of in-distribution and out-of-distribution likelihoods, thus making them separable. 



Under this objective function, we search for an optimal distribution of network architectures $p(\alpha)$ by deriving the corresponding parameters that characterize $p(\alpha)$. Because the score requires aggregating the results from multiple architectures $\alpha$, optimizing such a score using existing search methods can be intractable, as they typically only consider a single architecture at a time. Later, we will show how to circumvent this problem in our optimization formulation.

\subsection{Search Space}

NADS constructs a layer-wise search space with a pre-defined macro-architecture, where each layer can have a different architecture component. Such a search space has been studied by \citet{zoph2016neural, liu2018darts, real2019regularized}, where it shows to be both expressive and scalable/efficient.

The macro-architecture closely follows the Glow architecture presented in \citet{kingma2018glow}. Here, each layer consists of an actnorm, an invertible $1\times1$ convolution, and an affine coupling layer. Instead of pre-defining the affine coupling layer, we allow it to be optimized by our architecture search. The search space can be viewed in Figure~\ref{nasops}. Here, each operational block of the affine coupling layer is selected from a list of candidate operations that include $3\times3$ average pooling, $3\times3$ max pooling, skip-connections, $3\times3$ and $5\times5$ separable convolutions, $3\times3$ and $5\times5$ dilated convolutions, identity, and zero. We choose this search space to answer the following questions towards better architectures for OoD detection:
\squishlist
  \item What topology of connections between layers is best for uncertainty quantification? Traditional likelihood estimation architectures focus only on feedforward connections without adding any skip-connection structures. However, adding skip-connections may improve optimization speed and stability.
  \item Are more features/filters better for OoD detection? More feature outputs of each layer should lead to a more expressive model. However, if many of those features are redundant, it may slow down learning, overfitting nuisances and resulting in sub-optimal models.
  \item Which operations are best for OoD detection? Intuitively, operations such as max/average pooling should not be preferred, as they discard information of the original data point ``too aggressively". However, this intuition remains to be confirmed.
\squishend

\begin{figure*}[t!]
  \centering
  \includegraphics[width=0.75\textwidth]{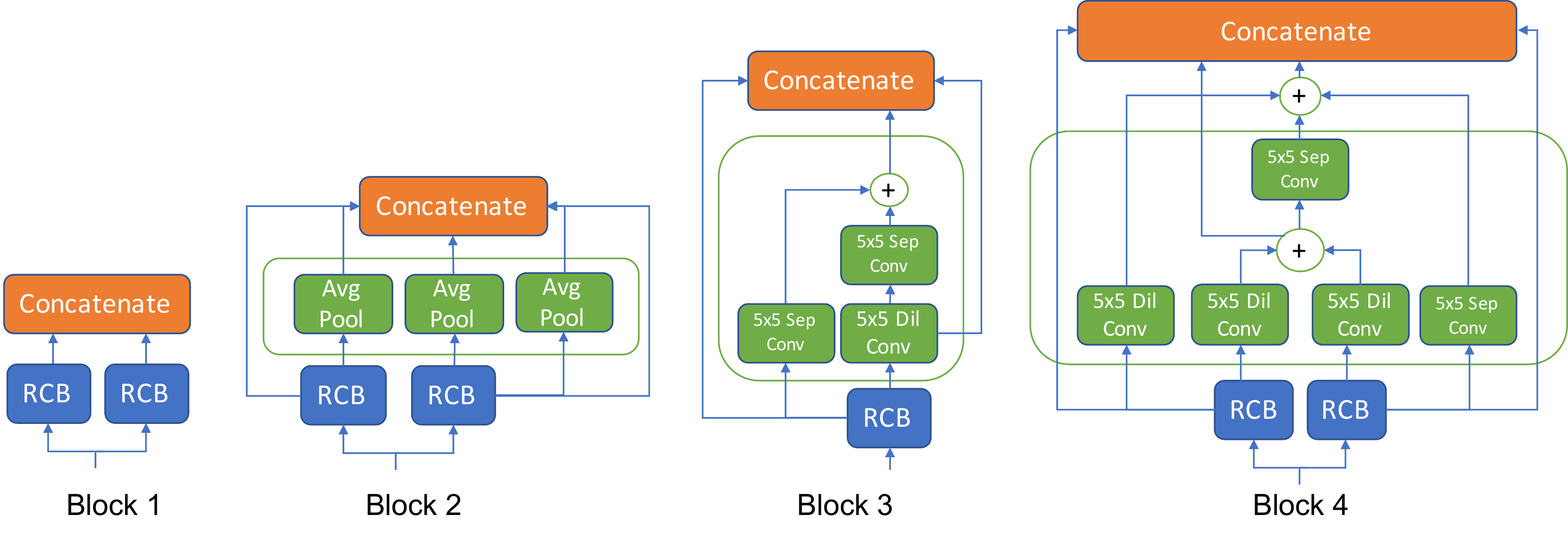}
  \caption{Summary of our architecture search findings: the most likely architecture structure for each block $K$ found by NADS.}
  \label{nas_blocks_summary}
\end{figure*}

\subsection{Optimization}

Having specified our proxy task and search space, we now describe our optimization method for NADS. Specifically, let $\mathcal{A}$ denote our discrete architecture search space and $\alpha \in \mathcal{A}$ be an architecture in this space. Let $ l_{\theta^{*}}(\alpha) $ be the loss function of architecture $ \alpha $ with its parameters set to $\theta^{*}$ such that it satisfies $\theta^{*} = \argmin_{\theta}l{(\theta|\alpha)}$ for some loss function $l(\cdot)$. We are interested in finding a distribution $p_\phi(\alpha)$ parameterized by $\phi$ that minimizes the expected loss of an architecture $\alpha$ sampled from it. We denote this loss function as $ L(\phi) = \mathbb{E}_{\alpha \sim p_\phi(\alpha)}[l_{\theta^{*}}(\alpha)] $. For our NADS, this loss function is the negative WAIC score of in-distribution data $L(\phi) = - \text{WAIC}(X) $.


Several difficulties arise when naively attempting to optimize this setup. Firstly, the objective function involves computing an expectation and variance over all possible discrete architectures. We alleviate this problem by approximating the WAIC objective through Monte Carlo sampling. Specifically, we can sample $M$ architectures from $p_\phi(\alpha)$ and approximate the WAIC score expectation and variance terms as
\begin{align}
\begin{split}
    \text{WAIC} & (X) \approx \sum_{i = 1}^N \Bigg[ \sum_{j=1}^{M} \log p(\boldsymbol{x}_i | \alpha_j) - \\ 
    & \bigg( \sum_{j=1}^{M} (\log p(\boldsymbol{x}_i | \alpha_j))^2 - \Big( \sum_{j=1}^{M} \log p(\boldsymbol{x}_i | \alpha_j) \Big)^2 \bigg) \Bigg]
\end{split}
\label{eq:WAICApprox}
\end{align}
Despite this approximation, optimizing  (\ref{eq:WAICApprox}) with respect to $p_\phi(\alpha)$, a distribution over high-dimensional discrete random variables $\alpha$, is still intractable, as we would still need to search for the optimal network parameters for each newly sampled architecture. To circumvent this, we utilize a continuous relaxation for the discrete search space, allowing us to approximately optimize the discrete architectures through backpropagation and weight sharing between common architecture blocks, as similarly implemented by \citet{xie2018snas} and \citet{chang2019differentiable}. Other potential possibilities for directly optimizing the discrete variables \cite{yin2019arsm,icassp_arsm,dadaneh2020pairwise} are prohibitively computationally expensive for our setup.

For clarity of exposition, we first focus on sampling an architecture with a single hidden layer. In this setting, we intend to find a probability vector $\boldsymbol{\phi} = [\phi_1, \dots, \phi_K]$ with which we randomly pick a single operation from a list of $K$ different operations $ [o_1, \dots, o_K] $. Let $\boldsymbol{b} = [b_1, \dots, b_K]$ denote the random categorical indicator vector sampled from $\boldsymbol{\phi}$, where $b_i$ is $1$ if the $i^{th}$ operation is chosen, and zero otherwise. Note that $\boldsymbol{b}$ is equivalent to the discrete architecture variable $\alpha$ in this setting. With this, we can write the random output $\boldsymbol{y}$ of the hidden layer given input $\boldsymbol{x}$ as
\begin{equation*}
    \boldsymbol{y} = \sum_{i=1}^{K} b_i \cdot o_i(\boldsymbol{x}).
\end{equation*}
To make optimization tractable, we relax the discrete mask $\boldsymbol{b}$ to be a continuous random variable $\tilde{\boldsymbol{b}}$ using the Gumbel-Softmax reparameterization \citep{gumbel1954statistical, maddison2014sampling} as follows:
\begin{equation*}
    \tilde{b}_i = \frac{\exp((\log(\phi_i) + g_i)/\tau)}{\sum_{j=1}^{k}\exp((\log(\phi_i) + g_i)/\tau)} \quad \text{for} \quad i = 1, \dots, K.
\end{equation*}
Here, $g_1 \dots g_k \sim -\log(-\log(u))$ where $u \sim \text{Unif}(0,1)$, and $\tau$ is a temperature parameter. For low values of $\tau$, $\tilde{\boldsymbol{b}}$ approaches a sample of a categorical random variable, recovering the original discrete problem. While for high values, $\tilde{\boldsymbol{b}}$ will equally weigh the $K$ operations \citep{jang2016categorical}. Using this, we can compute backpropagation by approximating the gradient of the discrete architecture $\alpha$ with the gradient of the continuously relaxed categorical random variable $\tilde{\boldsymbol{b}}$, as $\nabla_{\theta, \phi} \alpha = \nabla_{\theta, \phi} \boldsymbol{b} \approx \nabla_{\theta, \phi} \tilde{\boldsymbol{b}}$. With this backpropagation gradient defined, generalizing the above setting to architectures with multiple layers simply involves recursively applying the above gradient relaxation to each layer. We can gradually remove the continuous relaxation and sample discrete architectures by annealing the temperature parameter $\tau$, allowing us to perform architecture search without using a validation set. 


\begin{figure*}[t!]
\centering
\subfigure{
\includegraphics[width=0.23\textwidth]{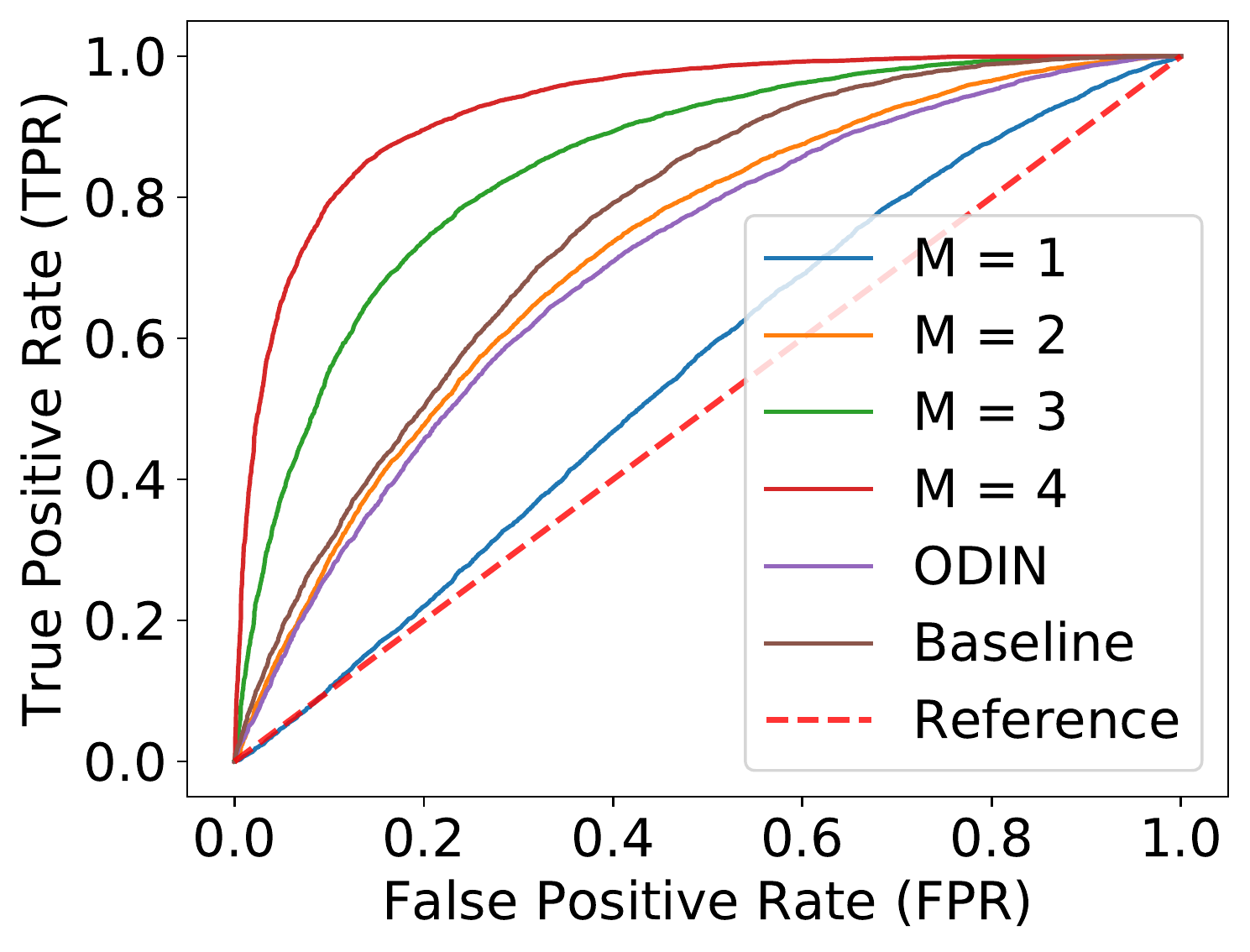}
\label{cifar100:fpr}
}
\subfigure{
\includegraphics[width=0.23\textwidth]{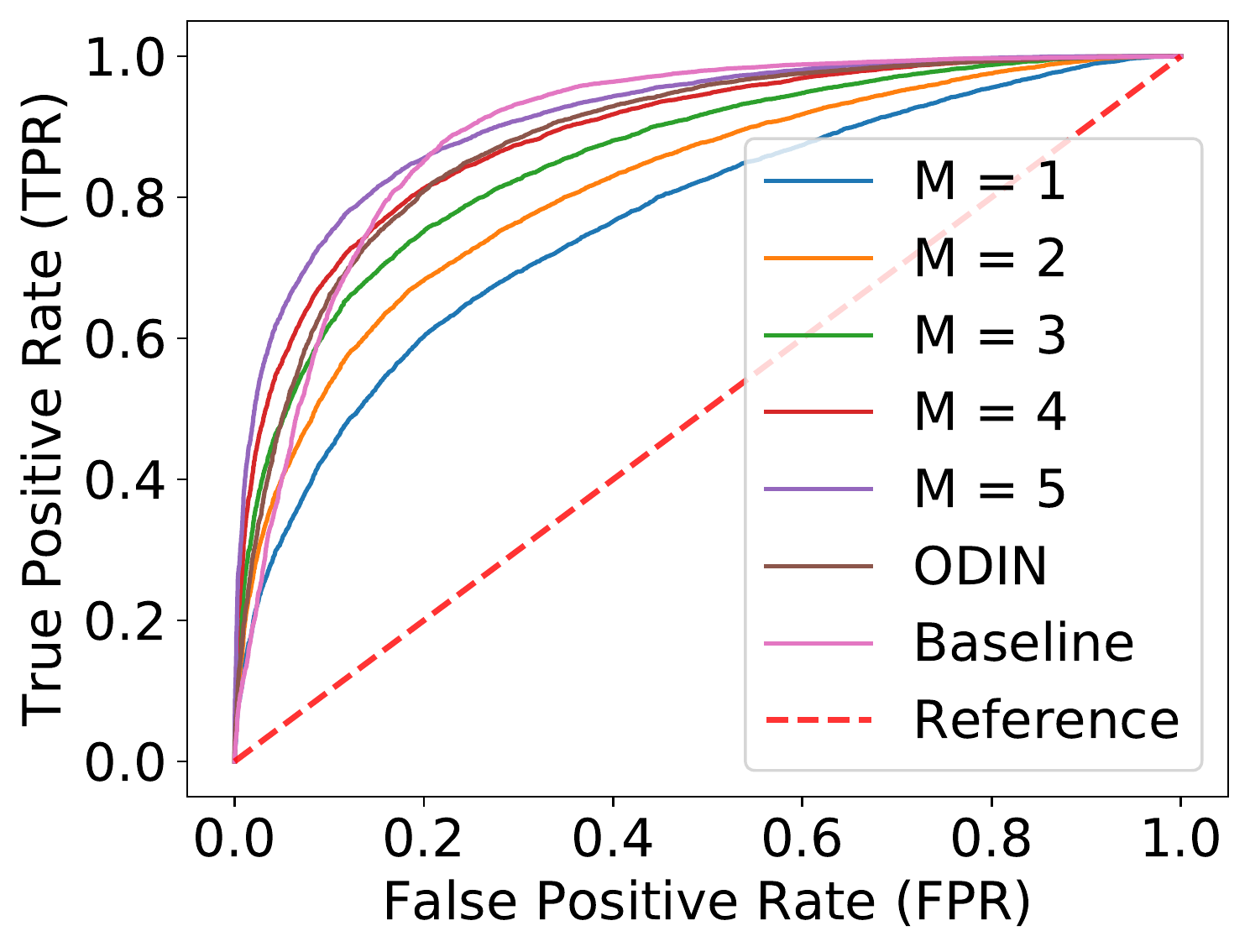}
\label{cifar:fpr}
}
\subfigure{
\includegraphics[width=0.23\textwidth]{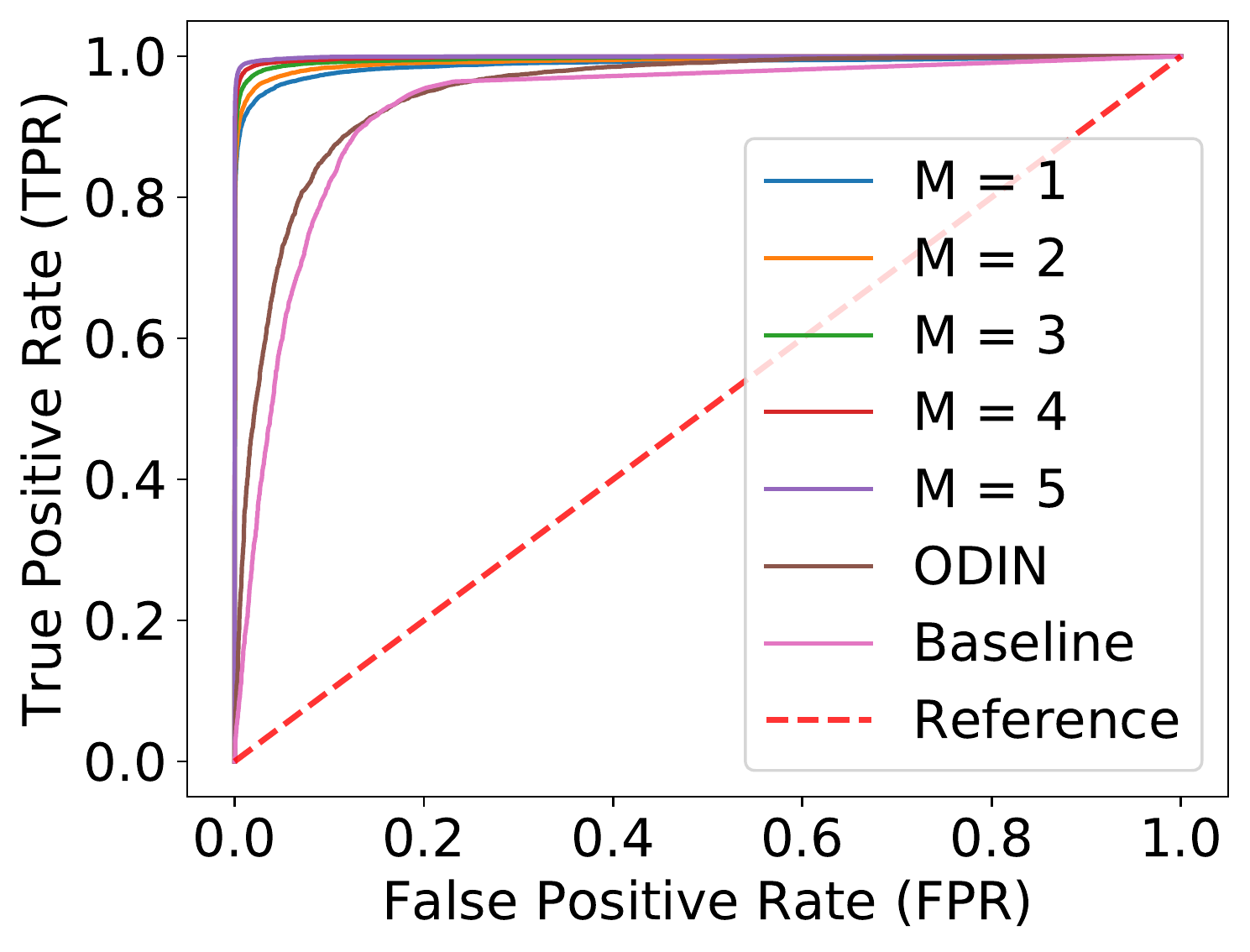}
\label{svhn:fpr}
}
\subfigure{
\includegraphics[width=0.23\textwidth]{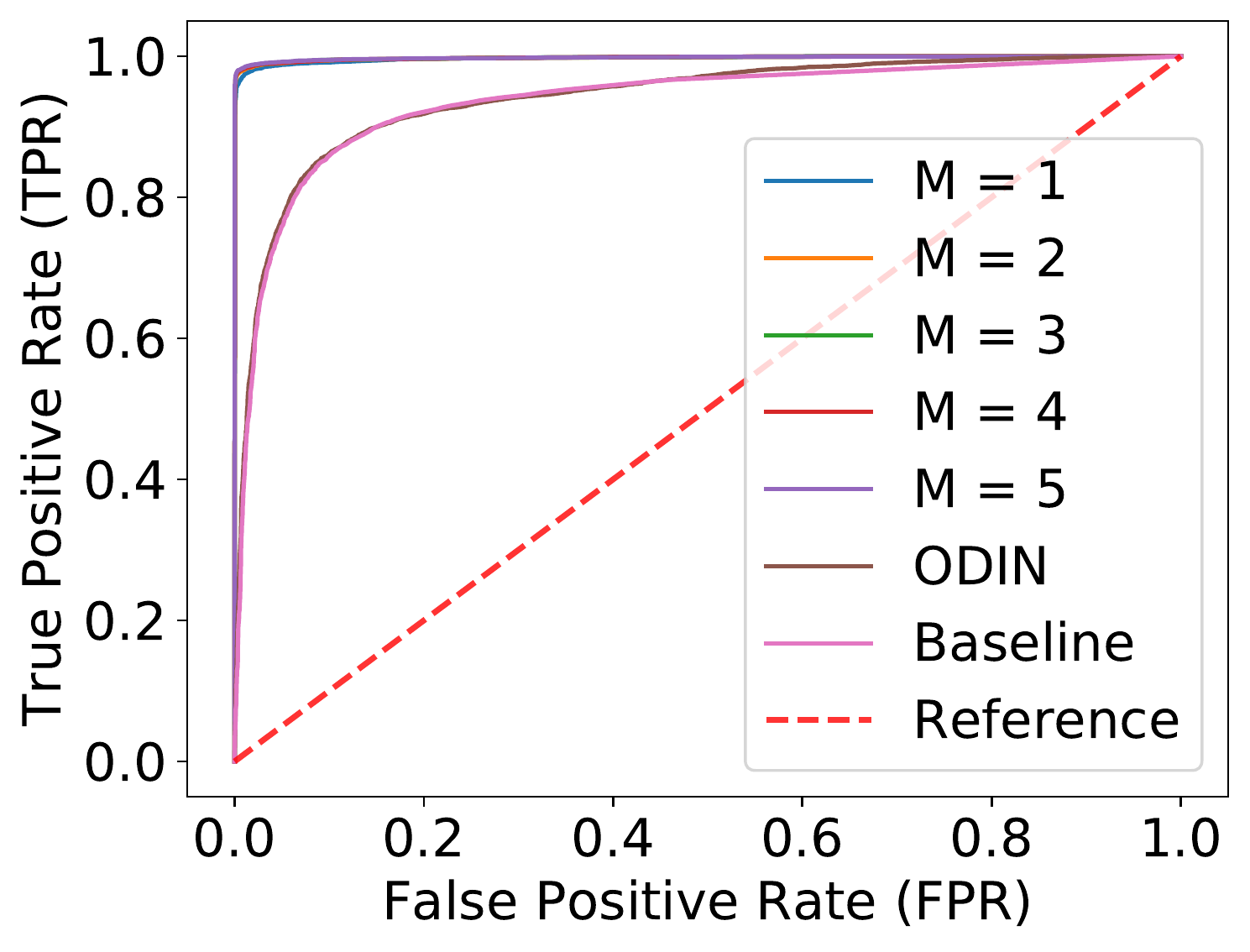}
\label{mnist:fpr}
}
\setcounter{subfigure}{0}
\subfigure[CIFAR100-CIFAR10]{
\includegraphics[width=0.23\textwidth]{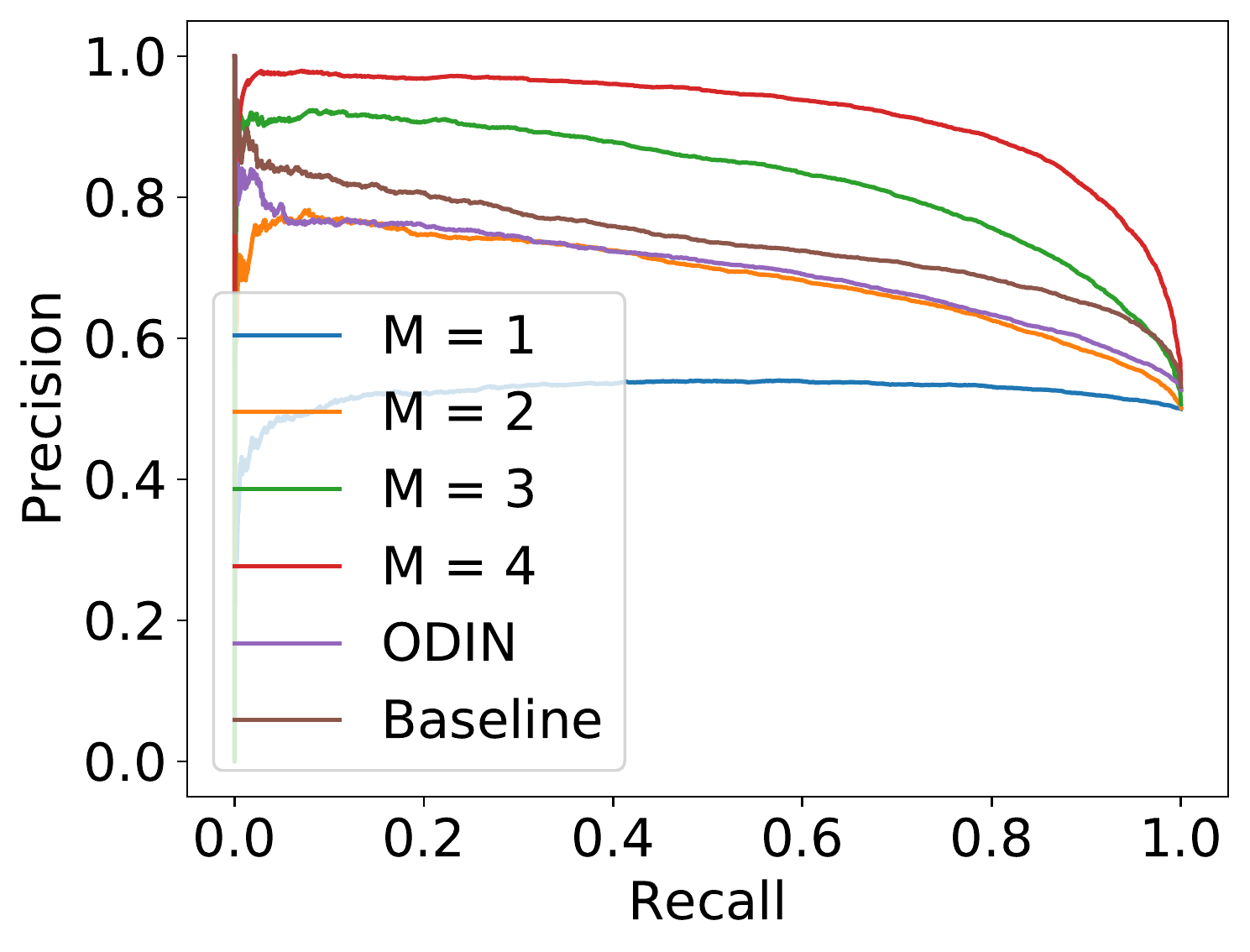}
\label{cifar100:prerec}
}
\subfigure[CIFAR10-CIFAR100]{
\includegraphics[width=0.23\textwidth]{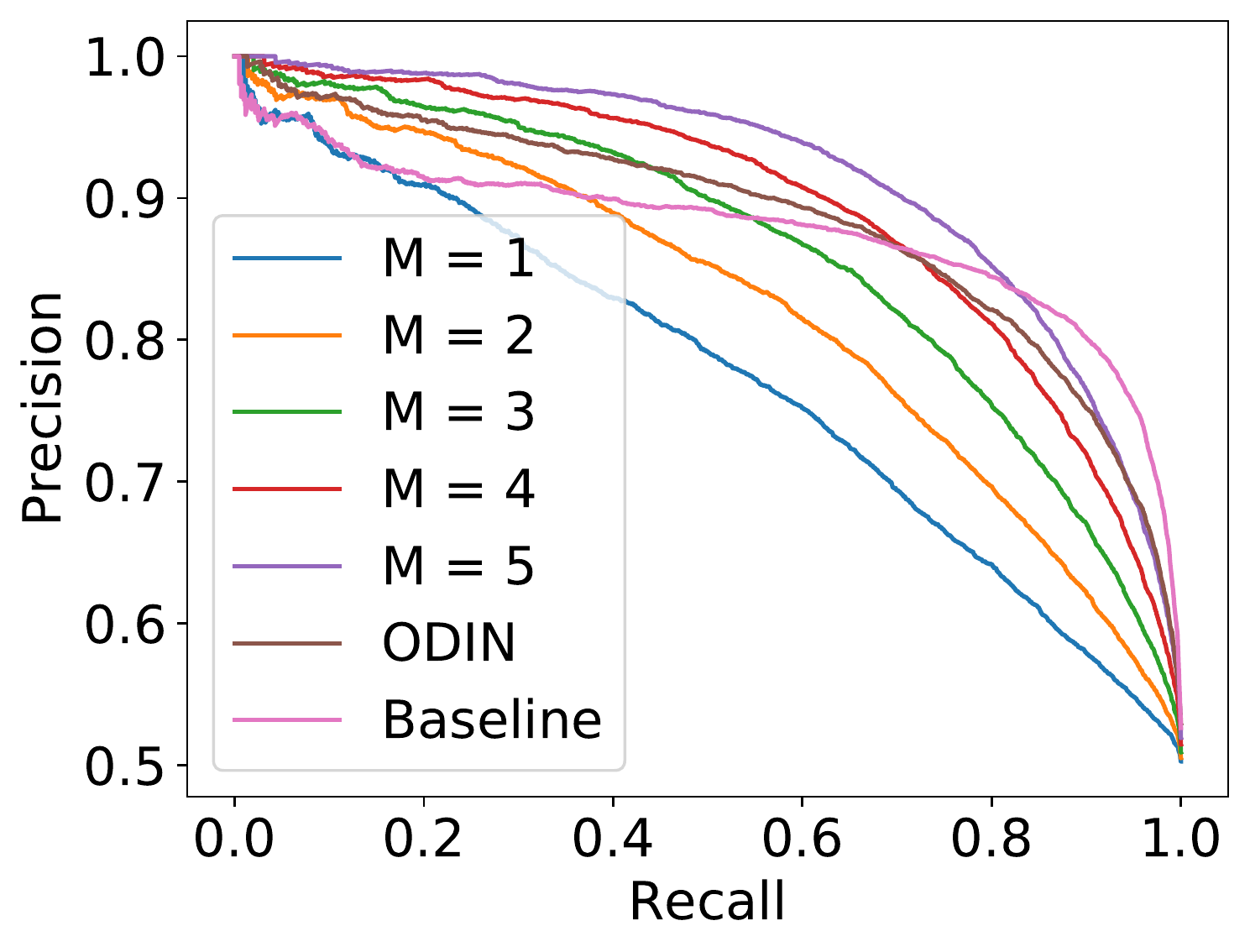}
\label{cifar:prerec}
}
\subfigure[SVHN-CIFAR10]{
\includegraphics[width=0.23\textwidth]{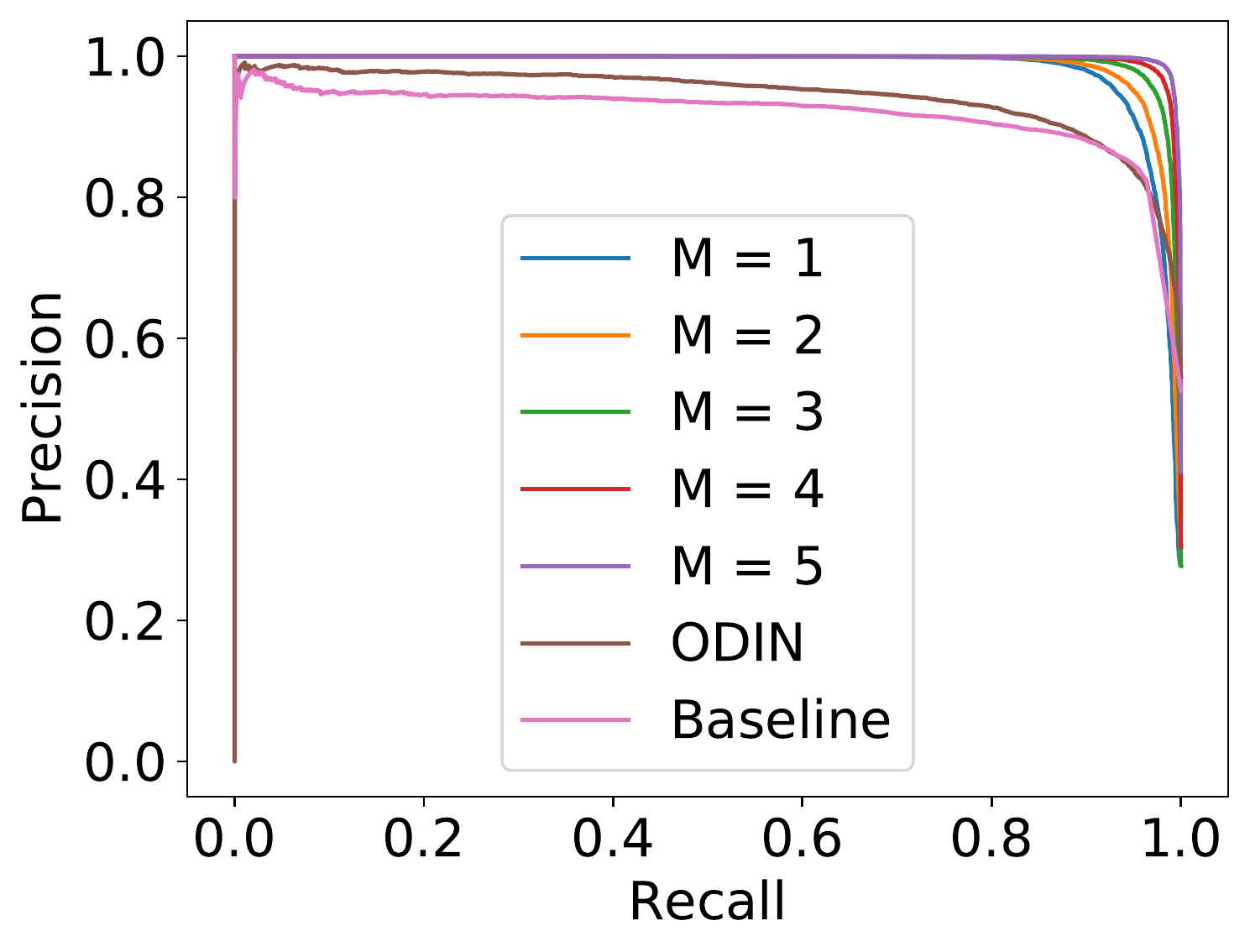}
\label{svhn:prerec}
}
\subfigure[MNIST-KMNIST]{
\includegraphics[width=0.23\textwidth]{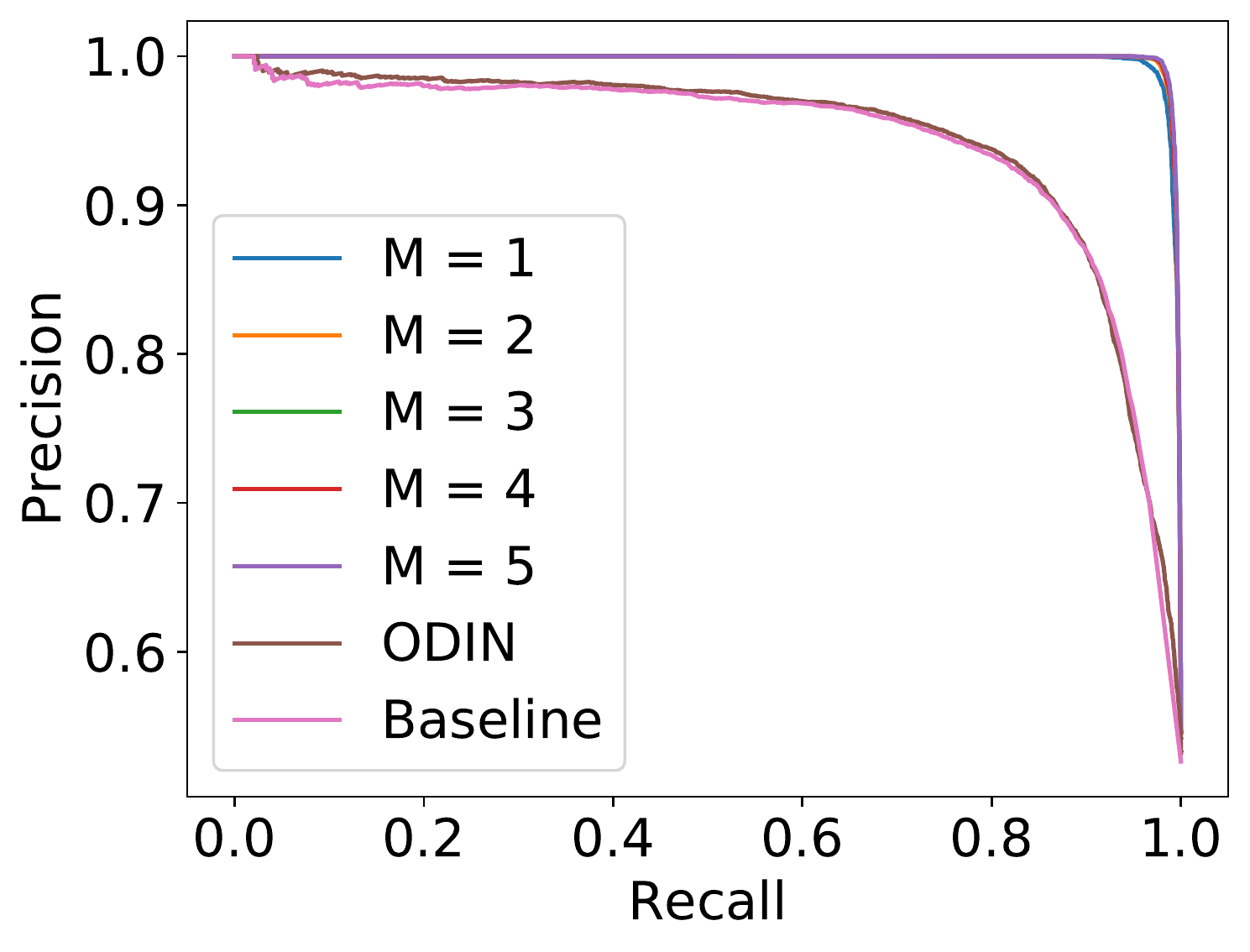}
\label{mnist:prerec}
}
\caption{ROC and PR curve comparison of the most challenging evaluation setups for our NADS ensemble. Here, `Baseline' denotes the method proposed by \cite{hendrycks2016baseline}. Subcaptions denote training-testing set pairs. Additional figures are provided in Section~\ref{append:additional-fpr} of the supplementary material.}
\label{fpr_roc}
\vspace{-0.1in}
\end{figure*}

\subsection{Search Results}


We applied our architecture search on five datasets: CelebA \citep{liu2018large}, CIFAR-10, CIFAR-100, \citep{krizhevsky2009learning}, SVHN \citep{netzer2011reading}, and MNIST \citep{lecun1998mnist}. In all experiments, we used the Adam optimizer with a fixed learning rate of $1\times10^{-5}$ with a batch size of 4 for 10000 iterations. We approximate the WAIC score using $M=4$ architecture samples, and set the temperature parameter $\tau=1.5$. The number of layers and latent dimensions is the same as in the original Glow architecture~\citep{kingma2018glow}, with 4 blocks and 32 flows per block. Images were resized to $64\times64$ as inputs to the model. With this setup, we found that we are able to identify neural architectures in less than 1 GPU day on an Nvidia RTX 2080 Ti graphics card.

Our findings are summarized in Figure~\ref{nas_blocks_summary}, while more samples from our architecture search can be seen in Section~\ref{append:architecture_samples} of the supplementary material. Observing the most likely architecture components found on all of the datasets, a number of notable observations can be made:
\squishlist
  \item The first few layers have a simple feedforward structure, with either only a few convolutional operations or average pooling operations. On the other hand, more complicated structures with skip connections are preferred in the deeper layers of the network. We hypothesize that in the first few layers, simple feature extractors are sufficient to represent the data well. Indeed, recent work on analyzing neural networks for image data have shown that the first few layers have filters that are very similar to SIFT features or wavelet bases~\citep{ zeiler2014visualizing,lowe1999object}.
  \item The max pooling operation is almost never selected by the architecture search. This confirms our hypothesis that operations that discard information about the data is unsuitable for OoD detection. However, to our surprise, average pooling is preferred in the first layers of the network. We hypothesize that average pooling has a less severe effect in discarding information, as it can be thought of as a convolutional filter with uniform weights.
  \item The deeper layers prefer a more complicated structure, with some components recovering the skip connection structure of ResNets \citep{he2016deep}. We hypothesize that deeper layers may require more skip connections in order to feed a strong signal for the first few layers. This increases the speed and stability of training. Moreover, a larger number of features can be extracted using the more complicated architecture.
\squishend


Interestingly enough, we found that the architectures that we sample from our NADS perform well in image generation without further retraining, as shown in Section~\ref{append:image_generation_samples} of the supplementary material.

\begin{figure*}[ht!]
\centering
\subfigure[MNIST]{
\includegraphics[width=0.48\textwidth]{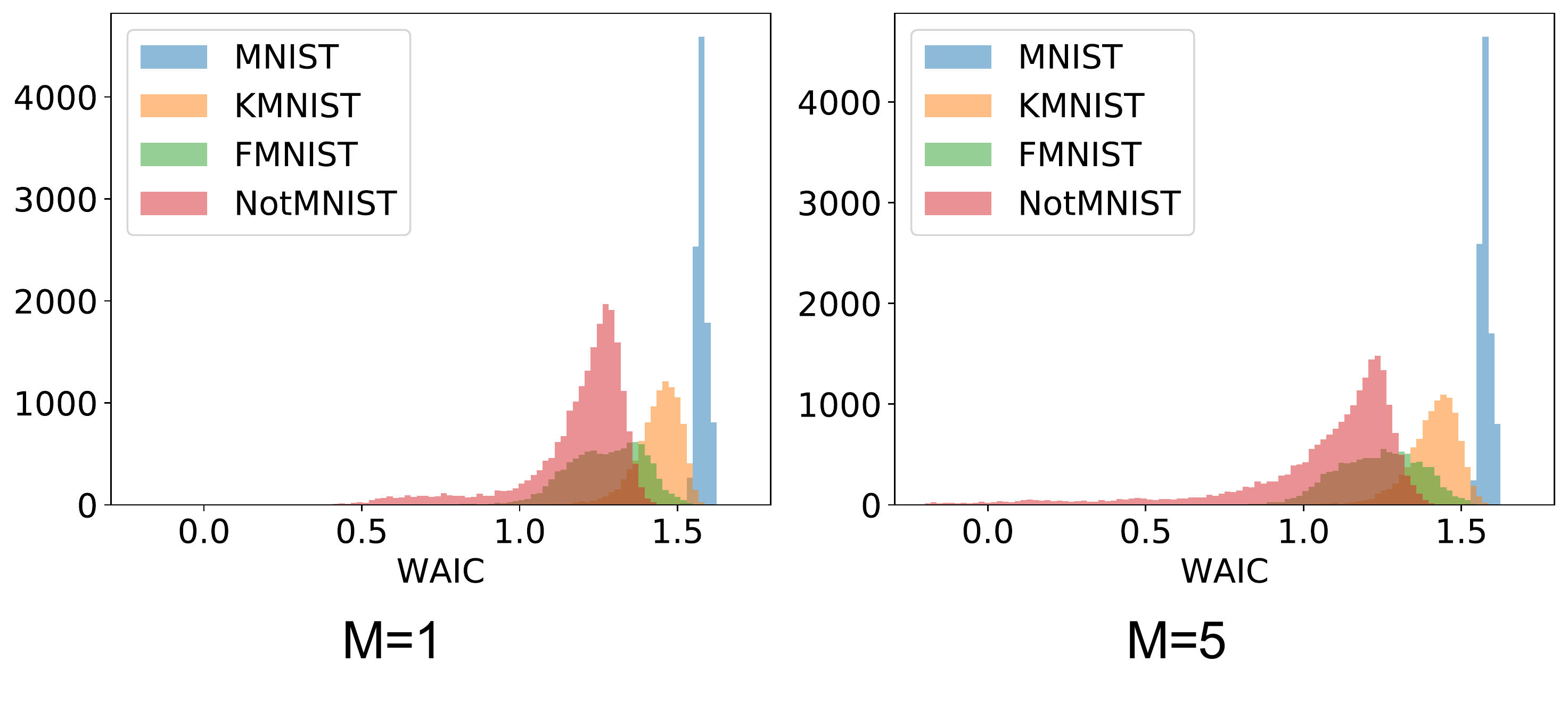}
\label{cifar10-hist}
}
\subfigure[SVHN]{
\includegraphics[width=0.48\textwidth]{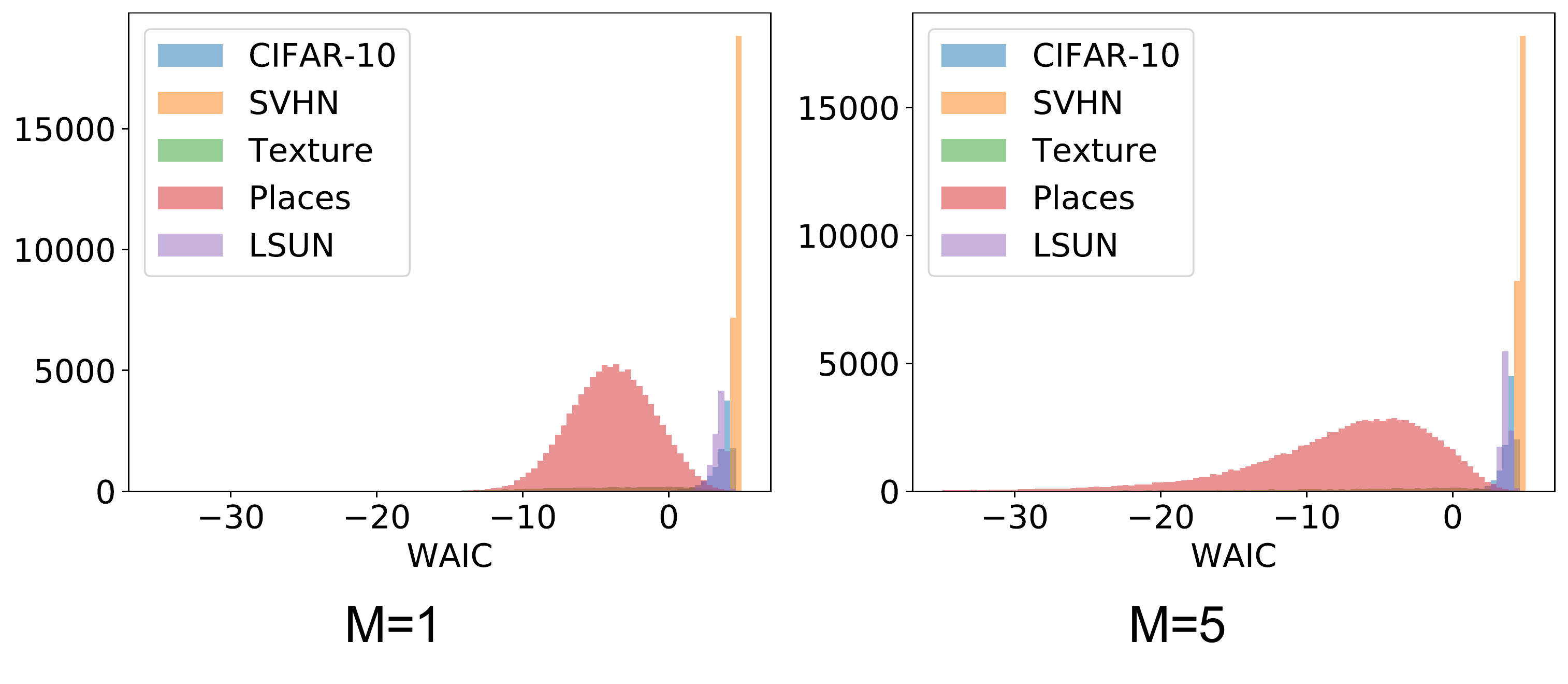}
\label{svhn-hist}
}
\vspace{-0.1in}
\subfigure[CIFAR-10]{
\includegraphics[width=0.98\textwidth]{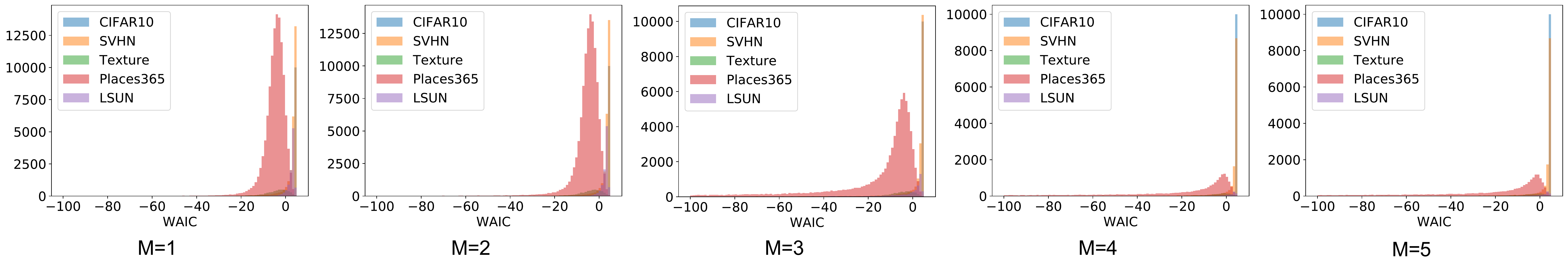}
\label{svhn-hist}
}
\caption{Effect of ensemble size to the distribution of WAIC scores estimated by model ensembles trained on different datasets. Larger ensemble sizes causes the WAIC score likelihood estimate of OoD data to be lower. Additional histograms for different ensemble sizes in Section~\ref{append:ensemble-effect} of the supplementary material are with higher resolution.}
\label{ensemble_comparison_small}
\vspace{-0.1in}
\end{figure*}

\section{Bayesian Model Ensemble of Neural Architectures}

\subsection{Model Ensemble Formulation}

Using the architectures sampled from our search, we create a Bayesian ensemble of models to estimate the WAIC score. Each model of our ensemble is weighted according to its probability as in~\citet{hoeting1999bayesian}. The log-likelihood estimate as well as the variance of this model ensemble is given as follows:
\begin{small}
\begin{align*}
\begin{split}
    \mathbb{E}_{\alpha \sim p_\phi(\alpha)}[\log p(\boldsymbol{x})] & = \sum_{\alpha \in \mathcal{A}} p_\phi(\alpha) \log p(\boldsymbol{x} | \alpha) \\
    & \approx \sum_{i=1}^{M} \frac{p_\phi(\alpha_i)}{\sum_{j=1}^{M} p_\phi(\alpha_j)} \log p(\boldsymbol{x} | \alpha_i)
\end{split}
\end{align*}
\begin{align*}
\begin{split}
    \mathbb{V}_{\alpha \sim p_\phi(\alpha)} & [\log p(\boldsymbol{x})] \approx \sum_{i=1}^{M} \frac{p_\phi(\alpha_i)}{\sum_{j=1}^{M} p_\phi(\alpha_j)} \Big( \mathbb{V}[\log p(\boldsymbol{x} | \alpha_i)] \\ & + (\log p(\boldsymbol{x} | \alpha_i))^2 \Big) - \mathbb{E}_{\alpha \sim p_\phi(\alpha)}[\log p(\boldsymbol{x})]^2
\end{split}
\end{align*}
\end{small}

Intuitively, we are weighing each member of the ensemble by their posterior architecture distribution $p_\phi(\alpha)$, a measure of how likely each architecture is in optimizing the WAIC score. We note that for our setup, $ \mathbb{V}[\log p_{\alpha_i}(x)] $ is zero for each model in our ensemble; however, for models which do have variance estimates, such as models that incorporate variational dropout~\citep{gal2017concrete,boluki2020learnable,kingma2015variational, gal2016dropout}, this term may be nonzero. Using these estimates, we are able to approximate the WAIC score in Equation~(\ref{WAIC_equation}). 

\subsection{Ensemble Results}


\begin{table*}
\centering
\caption{OoD detection results on various evaluation setups. We compared our method with MSP (Baseline) \citep{hendrycks2016baseline}, NAS following the DARTS search design (DARTS) \cite{liu2018darts}, and Outlier Exposure (OE) \citep{hendrycks2018deep}.}
{\footnotesize
\begin{tabular}{cccccc|cccc|cccc}
\hline
\multirow{2}{*}{$D_{in}$} & \multirow{2}{*}{$D_{out}$} & \multicolumn{4}{c}{FPR\% at TPR 95\%} & \multicolumn{4}{c}{AUROC\%} & \multicolumn{4}{c}{AUPR\%} \\
\cline{3-14}
& & Base & DARTS & OE & Ours & Base & DARTS & OE & Ours & Base & DARTS & OE & Ours \\
\hline
\multirow{3}{*}{\STAB{\rotatebox[origin=c]{90}{MNIST}}} 
& not-MNIST & 10.3 & 23.07 & 0.25 & \textbf{0.00} & 97.2 & 94.62 & 99.86 & \textbf{100}  & 97.4 & 96.81 & 99.86 & \textbf{100}    \\
& F-MNIST & 61.1 & 8.29 & 0.99 & \textbf{0.00} & 88.8 & 97.59 & 99.83 & \textbf{100}  & 90.8 & 97.06 & 99.83 & \textbf{100}    \\
& K-MNIST & 29.6 & 9.37 & \textbf{0.03} & 0.76 & 93.6 & 97.39 & 97.60 & \textbf{99.80}  & 94.3 & 96.90 & 97.05 & \textbf{99.84}    \\
\hline
\multirow{4}{*}{\STAB{\rotatebox[origin=c]{90}{SVHN}}} 
& Texture & 33.9 & 23.43 & 1.04 & \textbf{0.07} & 89.3 & 94.25 & \textbf{99.75} & 99.26  & 86.8 & 80.98 & \textbf{99.09} & 97.75    \\
& Places365 & 22.2 & 16.17 & 0.02 & \textbf{0.00} & 92.8 & 95.74 & \textbf{99.99} & \textbf{99.99}  & 99.7 & 99.57 & \textbf{99.99} & \textbf{99.99}    \\
& LSUN & 26.8 & 16.16 & 0.05 & \textbf{0.02} & 88.2 & 95.44 & 99.98 & \textbf{99.99}  & 90.4 & 87.36 & 99.95 & \textbf{99.99}    \\
& CIFAR10 & 23.2 & 16.82 & 3.11 & \textbf{0.37} & 91.1 & 95.36 & 99.26 & \textbf{99.92}  & 91.9 & 87.45 & 97.88 & \textbf{99.83}    \\
\hline
\multirow{7}{*}{\STAB{\rotatebox[origin=c]{90}{CIFAR10}}} 
& SVHN & 30.5 & 19.47 & \textbf{8.41} & 17.05 & 89.5 & 93.58 & \textbf{98.20} & 97.65  & 94.9 & 96.25 & 97.97 & \textbf{99.07}    \\
& Texture & 39.8 & 24.25 & 14.9 & \textbf{0.25} & 87.7 & 92.18 & 96.7 & \textbf{99.81}  & 79.8 & 83.51 & 94.39 & \textbf{99.86}    \\
& Places365 & 36.0 & 41.64 & 19.07 & \textbf{0.00} & 88.1 & 87.65 & 95.41 & \textbf{100}  & 99.5 & 99.42 & 95.32 & \textbf{100}    \\
& LSUN & 14.6 & 30.02 & 15.20 & \textbf{0.44} & 95.4 & 90.11 & 96.43 & \textbf{99.83}  & 96.1 & 86.88 & 96.01 & \textbf{99.89}    \\
& CIFAR100 & 33.1 & 35.72 & \textbf{26.59} & 36.36 & 88.7 & 88.43 & \textbf{92.93} & 91.23  & 87.7 & 72.95 & \textbf{92.13} & 91.60    \\
& Gaussian & 6.3 & 11.67 & 0.7 & \textbf{0.00} & 97.7 & 95.55 & 99.6 & \textbf{100}  & 93.6 & 87.46 & 94.3 & \textbf{100}    \\
& Rademacher & 6.9 & 10.73 & 0.5 & \textbf{0.00}  & 96.9 & 95.26 & 99.8 & \textbf{100}  & 89.7 & 84.10 & 97.4 & \textbf{100}    \\
\hline
\multirow{7}{*}{\STAB{\rotatebox[origin=c]{90}{CIFAR100}}} 
& SVHN & 46.2 & 53.81 & \textbf{42.9} & 45.92 & 82.7 & 79.30 & 86.9 & \textbf{94.35}  & 91.3 & 88.52 & 80.21 & \textbf{96.01}    \\
& Texture & 74.3 & 62.49 & 55.97 & \textbf{0.42} & 72.6 & 75.00 & 84.23 & \textbf{99.76}  & 60.1 & 57.77 & 75.76 & \textbf{99.81}    \\
& Places365 & 63.2 & 64.91 & 57.77 & \textbf{0.012}  & 76.2 & 75.72 & 82.65 & \textbf{99.99} & 98.9 & 98.78 & 81.47 & \textbf{99.99}    \\
& LSUN & 69.4 & 56.01 & 57.5 & \textbf{38.85} & 83.7 & 77.57 & 83.4 & \textbf{90.65} & 70.1 & 72.94 & 77.85 & \textbf{90.61}    \\
& CIFAR10 & 62.5 & 61.62 & 59.96 & \textbf{45.62} & 75.8 & 76.15 & 77.53 & \textbf{83.27}  & 74.0 & 71.41 & 72.82 & \textbf{81.48}    \\
& Gaussian & 29.3 & 26.70 & 12.1 & \textbf{0.00} & 86.5 & 87.82 & 95.7 & \textbf{100} & 66.1 & 69.05 & 71.1 & \textbf{100}    \\
& Rademacher & 59.4 & 16.19 & 17.1 & \textbf{0.00} & 51.7 & 92.05 & 93.0 & \textbf{100} & 32.7 & 73.02 & 56.9 & \textbf{100}    \\
\hline
\end{tabular}
}
\label{ood_table}
\vspace{-0.1in}
\end{table*}

We trained our proposed method on 4 datasets $D_{in}$: CIFAR-10, CIFAR-100 \citep{krizhevsky2009learning}, SVHN \citep{netzer2011reading}, and MNIST \citep{lecun1998mnist}. In all experiments, we randomly sampled an ensemble of $M=5$ models from the posterior architecture distribution $p_{\phi^{*}}(\alpha)$ found by NADS. We then retrained each architecture for 150000 iterations using Adam with a learning rate of $1\times10^{-5}$. 

We first show the effects of increasing the ensemble size in Figure~\ref{ensemble_comparison_small} and Section~\ref{append:ensemble-effect} of the supplementary material. Here, we can see that increasing the ensemble size causes the OoD WAIC scores to decrease as their corresponding histograms shift away from the training data WAIC scores, thus improving OoD detection performance. Next, we compare our ensemble search method against a traditional ensemble method that uses a single Glow \cite{kingma2018glow} architecture trained with multiple random initializations. We find that our method is superior for OoD detection compared to the traditional ensemble method, as shown in Table~\ref{ensemble_table} of the supplementary material.

We evaluate our NADS ensemble OoD detection method for screening out samples from datasets that the original model was not trained on ($D_{out}$). For SVHN, we used the Texture, Places, LSUN, and CIFAR-10 as the OoD dataset. For CIFAR-10 and CIFAR-100, we used the SVHN, Texture, Places, LSUN, CIFAR-100 (CIFAR-10 for CIFAR-100) datasets, as well as the Gaussian and Rademacher distributions as the OoD dataset. Finally, for MNIST, we used the not-MNIST, F-MNIST, and K-MNIST datasets. We compared our method against a baseline method that uses maximum softmax probability (MSP) \citep{hendrycks2016baseline}, as well as two popular OoD detection methods: ODIN \citep{liang2017enhancing} and Outlier Exposure (OE) \citep{hendrycks2018deep}.

ODIN attempts to calibrate the uncertainty estimates of an existing model by reweighing its output softmax score using a temperature parameter and through random perturbations of the input data. For this, we use DenseNet as the base model as described in \citet{liang2017enhancing}. On the other hand, OE models are trained to minimize a loss regularized by an outlier exposure loss term, a loss term that requires access to OoD samples, although they are not required to be from the tested OoD distribution.

We also show the improvements made by our design of the search space and the optimization objective by comparing our method to applying architecture search without taking these factors into consideration. To do this, we applied neural architecture search with the goal of maximizing classification accuracy on in-distribution data. Here, our search formulation closely follows the Differentiable Architecture Search (DARTS) method \cite{liu2018darts}. After identifying the optimal architecture, we screen out OoD data using the maximum softmax probability (MSP) \cite{hendrycks2016baseline}, a score that gives classification architectures the ability to screen out OoD data.

As shown in Tables~\ref{ood_table} and~\ref{odin_table} in the supplementary material, our method outperforms the baseline MSP and ODIN significantly while performing better or comparably with OE, which requires OoD data during training, albeit not from the testing distribution. Notably, our method was able to achieve an improvement of 57\% FPR on the CIFAR100 -- Places365 setup compared to OE. Comparing the original architecture used by MSP and the identified architecture by DARTS, we can see that there is an improvement in OoD detection performance, however, because the architectures are not tailored to perform OoD detection, our NADS was also able to outperform it in our experiments.

We plot Receiver Operating Characteristic (ROC) and Precision-Recall (PR) curves in Figure~\ref{fpr_roc} and Section~\ref{append:additional-fpr} of the supplementary material for more comprehensive comparison. In particular, our method consistently achieves high area under PR curve (AUPR\%), showing that we are especially capable of screening out OoD data in settings where their occurrence is rare. Such a feature is important in situations where anomalies are sparse, yet have disastrous consequences. Notably, ODIN underperforms in screening out many OoD datasets, despite being able to reach the original reported performance when testing on LSUN using a CIFAR10 trained model. This suggests that ODIN may not be stable for use on different anomalous distributions.

\section{Conclusion}
\label{conclusion}

Unlike NAS for common learning tasks, specifying a model and an objective to optimize for uncertainty estimation and outlier detection is not straightforward. Moreover, using a single model may not be sufficient to accurately quantify uncertainty and successfully screen out OoD data. We developed a novel neural architecture distribution search (NADS) formulation to identify a random ensemble of architectures that perform well on a given task. Instead of seeking to maximize the likelihood of in-distribution data which may cause OoD samples to be mistakenly given a higher likelihood, we developed a search algorithm to optimize the WAIC score, a Bayesian adjusted estimation of the data entropy. Using this formulation, we have identified several key features that make up good uncertainty quantification architectures, namely a simple structure in the shallower layers, use of information preserving operations, and a larger, more expressive structure with skip connections for deeper layers to ensure optimization stability. Using the architecture distribution learned by NADS, we then constructed an ensemble of models to estimate the data entropy using the WAIC score. We demonstrated the superiority of our method to existing OoD detection methods and showed that our method has highly competitive performance without requiring access to OoD samples. Overall, NADS as a new uncertainty-aware architecture search strategy enables model uncertainty quantification that is critical for more robust and generalizable deep learning, a crucial step in safely applying deep learning to healthcare, autonomous driving, and disaster response.

\section*{Acknowledgement}
The presented materials are based upon the work supported by the National Science Foundation under Grants CCF-1553281, IIS-1812641, and CCF-1934904; and the Defense Advanced Research Projects Agency under grand FA8750-18-2-0027. We also thank Texas A\&M High Performance Research Computing and Texas Advanced Computing Center for providing computational resources to perform experiments in this work.


\bibliography{example_paper}
\bibliographystyle{icml2020}

\newpage
\onecolumn
\appendix

\section{Fixed Model Ablation Study}
\label{append:ablation}

\begin{table}[!h]
\centering\vspace{-0.05in}
\caption{OoD detection results on various training and testing experiments comparing our method with a baseline ensembling method that uses a fixed architecture trained multiple times with different random initializations.}
\begin{tabular}{cccc|cc|cc}
\hline
\multirow{2}{*}{$D_{in}$} & \multirow{2}{*}{$D_{out}$} & \multicolumn{2}{c}{FPR\% at TPR 95\%} & \multicolumn{2}{c}{AUROC\%} & \multicolumn{2}{c}{AUPR\%} \\
\cline{3-8}
&    & Base Ensemble & Ours  & Base Ensemble & Ours & Base Ensemble & Ours                \\
\hline
\multirow{7}{*}{\STAB{\rotatebox[origin=c]{90}{CIFAR10}}} 
& SVHN & 50.07 & \textbf{17.05}           & 93.48 & \textbf{97.65}  & 95.98 & \textbf{99.07}    \\
& Texture & 6.22 & \textbf{0.25}       & 97.68 & \textbf{99.81}  & 97.44 & \textbf{99.86}    \\
& Places365 & 1.03 & \textbf{0.00}           & 99.59 & \textbf{100}  & 99.97 & \textbf{100}    \\
& LSUN & 34.35 & \textbf{0.44}           & 91.55 & \textbf{99.83}  & 92.15 & \textbf{99.89}    \\
& CIFAR100 & 65.13 & \textbf{36.36} & 78.44 & \textbf{91.23}  & 79.44 & \textbf{91.60}    \\
& Gaussian & \textbf{0.00} & \textbf{0.00} & \textbf{100} & \textbf{100}  & \textbf{100} & \textbf{100}    \\
& Rademacher & \textbf{0.00} & \textbf{0.00}  & \textbf{100} & \textbf{100}  & \textbf{100} & \textbf{100}    \\
\hline
\end{tabular}
\label{ensemble_table}
\end{table}

\section{OoD Detection Performance Comparison with ODIN}
\label{append:ODIN}

\begin{table}[!h]
\centering\vspace{-0.05in}
\caption{OoD detection results on various training and testing experiments comparing our method with ODIN \citep{liang2017enhancing}.}
\begin{tabular}{cccc|cc|cc}
\hline
\multirow{2}{*}{$D_{in}$} & \multirow{2}{*}{$D_{out}$} & \multicolumn{2}{c}{FPR\% at TPR 95\%} & \multicolumn{2}{c}{AUROC\%} & \multicolumn{2}{c}{AUPR\%} \\
\cline{3-8}
                  &                    & ODIN & Ours                   & ODIN & Ours                  & ODIN & Ours                \\
\hline
\multirow{3}{*}{\STAB{\rotatebox[origin=c]{90}{MNIST}}} &
not-MNIST & 8.7 & \textbf{0.00}          & 98.2 & \textbf{100}  & 98.0 & \textbf{100}    \\
& F-MNIST & 65 & \textbf{0.00}           & 88.6 & \textbf{100}  & 90.5 & \textbf{100}    \\
 & K-MNIST & 36.5 & \textbf{0.76}           & 94.0 & \textbf{99.80}  & 94.6 & \textbf{99.84}    \\
\hline
\multirow{4}{*}{\STAB{\rotatebox[origin=c]{90}{SVHN}}} 
& Texture & 33.9 & \textbf{0.07}          & 92.4 & \textbf{99.26}  & 88.2 & \textbf{97.75}    \\
& Places365 & 22.2 & \textbf{0.00}           & 94.9 & \textbf{99.99}  & 99.8 & \textbf{99.99}    \\
& LSUN & 26.8 & \textbf{0.02}           & 93.5 & \textbf{99.99}  & 93.1 & \textbf{99.99}    \\
& CIFAR10 & 21.6 & \textbf{0.37}           & 94.8 & \textbf{99.92}  & 94.4 & \textbf{99.83}    \\
\hline
\multirow{7}{*}{\STAB{\rotatebox[origin=c]{90}{CIFAR10}}} 
& SVHN & 36.5 & \textbf{17.05}           & 89.7 & \textbf{97.65}  & 95.6 & \textbf{99.07}    \\
& Texture & 76.2 & \textbf{0.25}           & 81.4 & \textbf{99.81}  & 76.7 & \textbf{99.86}    \\
& Places365 & 44.0 & \textbf{0.00}           & 89.0 & \textbf{100}  & 99.6 & \textbf{100}    \\
& LSUN & 3.9 & \textbf{0.44}           & 99.2 & \textbf{99.83}  & 99.2 & \textbf{99.89}    \\
& CIFAR100 & 45.4 & \textbf{36.36} & 88.3 & \textbf{91.23}  & 88.5 & \textbf{91.60}    \\
& Gaussian & 0.1 & \textbf{0.00} & \textbf{100} & \textbf{100}  & 99.9 & \textbf{100}    \\
& Rademacher & 0.3 & \textbf{0.00}  & 99.9 & \textbf{100}  & 99.8 & \textbf{100}    \\
\hline
\multirow{7}{*}{\STAB{\rotatebox[origin=c]{90}{CIFAR100}}} 
& SVHN & \textbf{32.8} & 45.92           & 90.3 & \textbf{94.35}  & 95.3 & \textbf{96.01}    \\
& Texture & 78.9 & \textbf{0.42}           & 75.7 & \textbf{99.76}  & 64.5 & \textbf{99.81}    \\
& Places365 & 63.3 & \textbf{0.012}           & 79.0 & \textbf{99.99}  & 99.1 & \textbf{99.99}    \\
& LSUN & \textbf{17.6} & 38.85           & \textbf{96.8} & 90.65  & \textbf{96.5} & 90.61    \\
& CIFAR10 & 78.2 & \textbf{45.62}           & 70.6 & \textbf{83.27}  & 69.7 & \textbf{81.48}    \\
& Gaussian & 1.3 & \textbf{0.00}           & 99.5 & \textbf{100}  & 97.8 & \textbf{100}    \\
& Rademacher & 13.8 & \textbf{0.00}           & 92.7 & \textbf{100}  & 75.0 & \textbf{100}    \\
\hline
\end{tabular}
\label{odin_table}
\end{table}

\clearpage

\section{Additional Sample Architectures}
\label{append:architecture_samples}

\begin{figure}[h!]
\centering
\subfigure[Block 1]{
\includegraphics[width=0.40\textwidth]{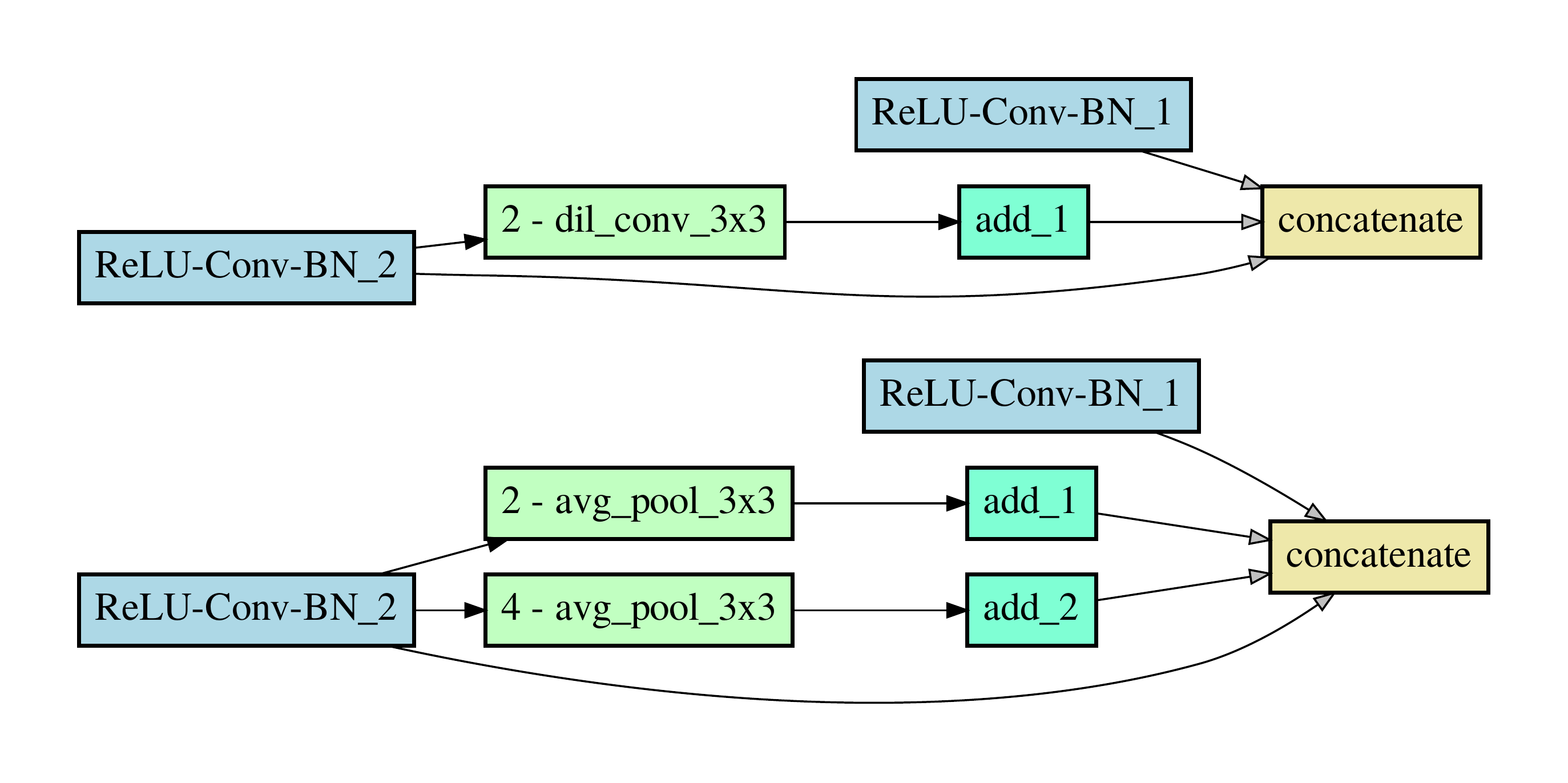}
}
\subfigure[Block 2]{
\includegraphics[width=0.40\textwidth]{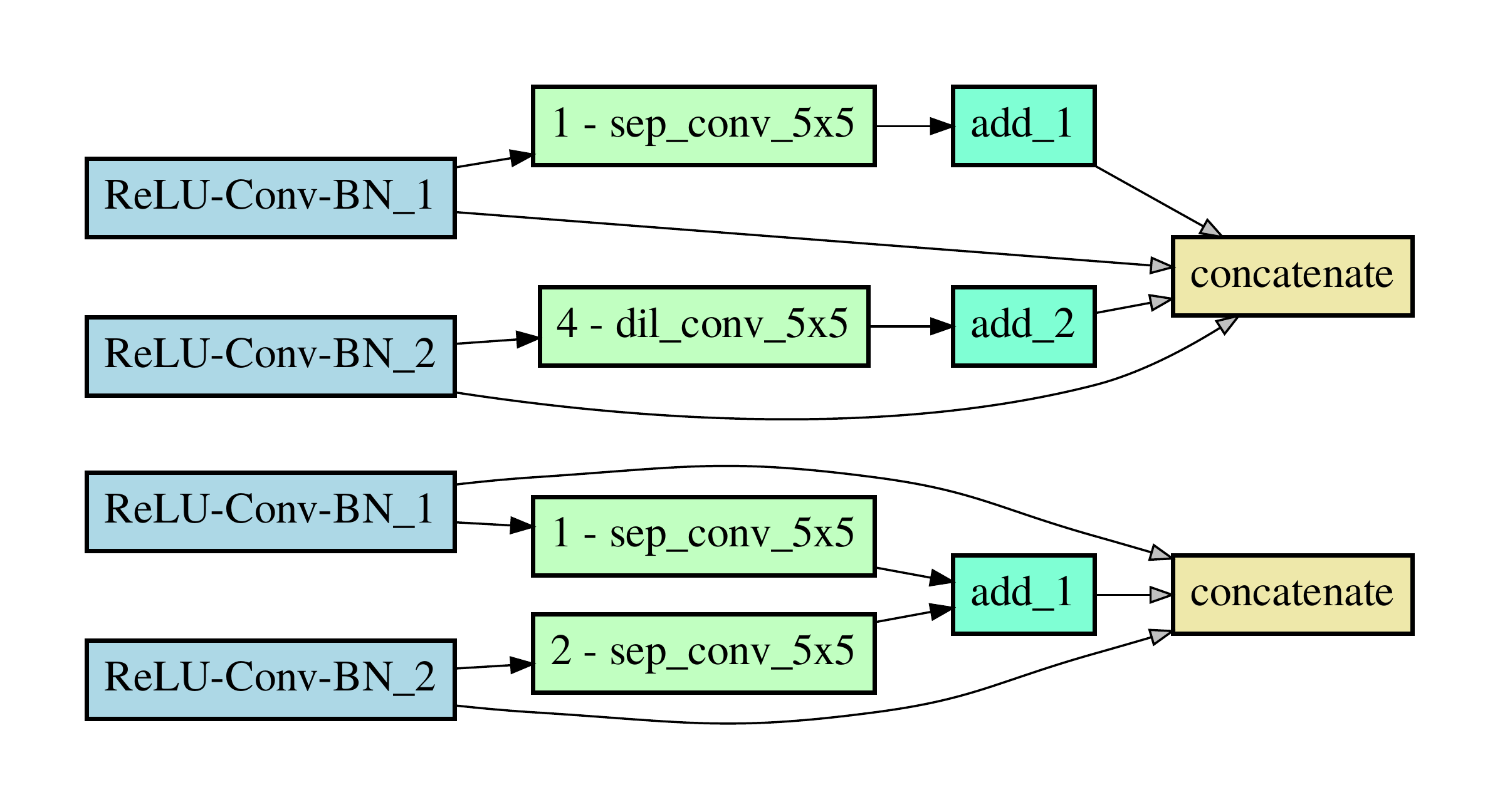}
}
\subfigure[Block 3]{
\includegraphics[width=0.40\textwidth]{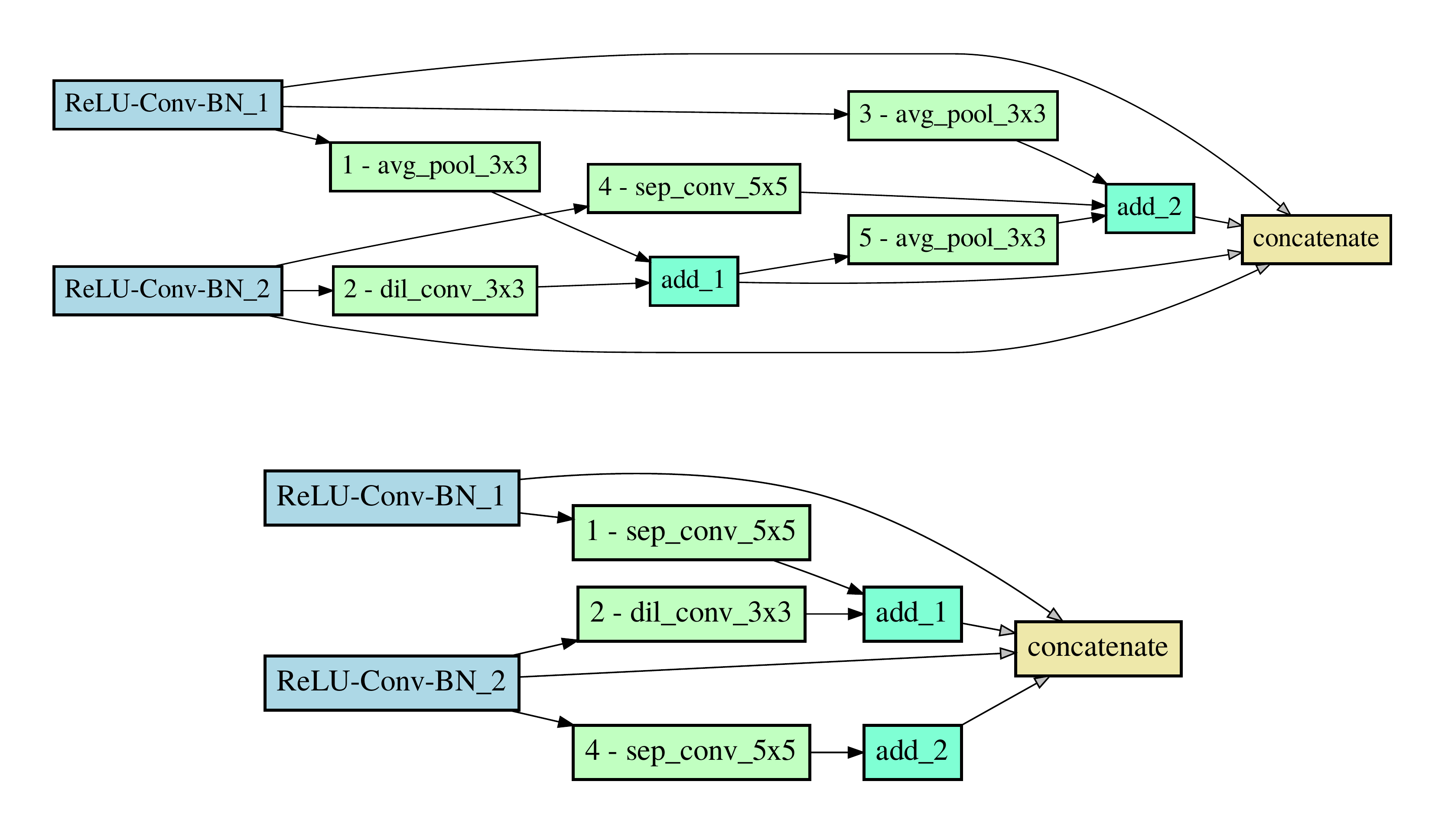}
}
\subfigure[Block 4]{
\includegraphics[width=0.40\textwidth]{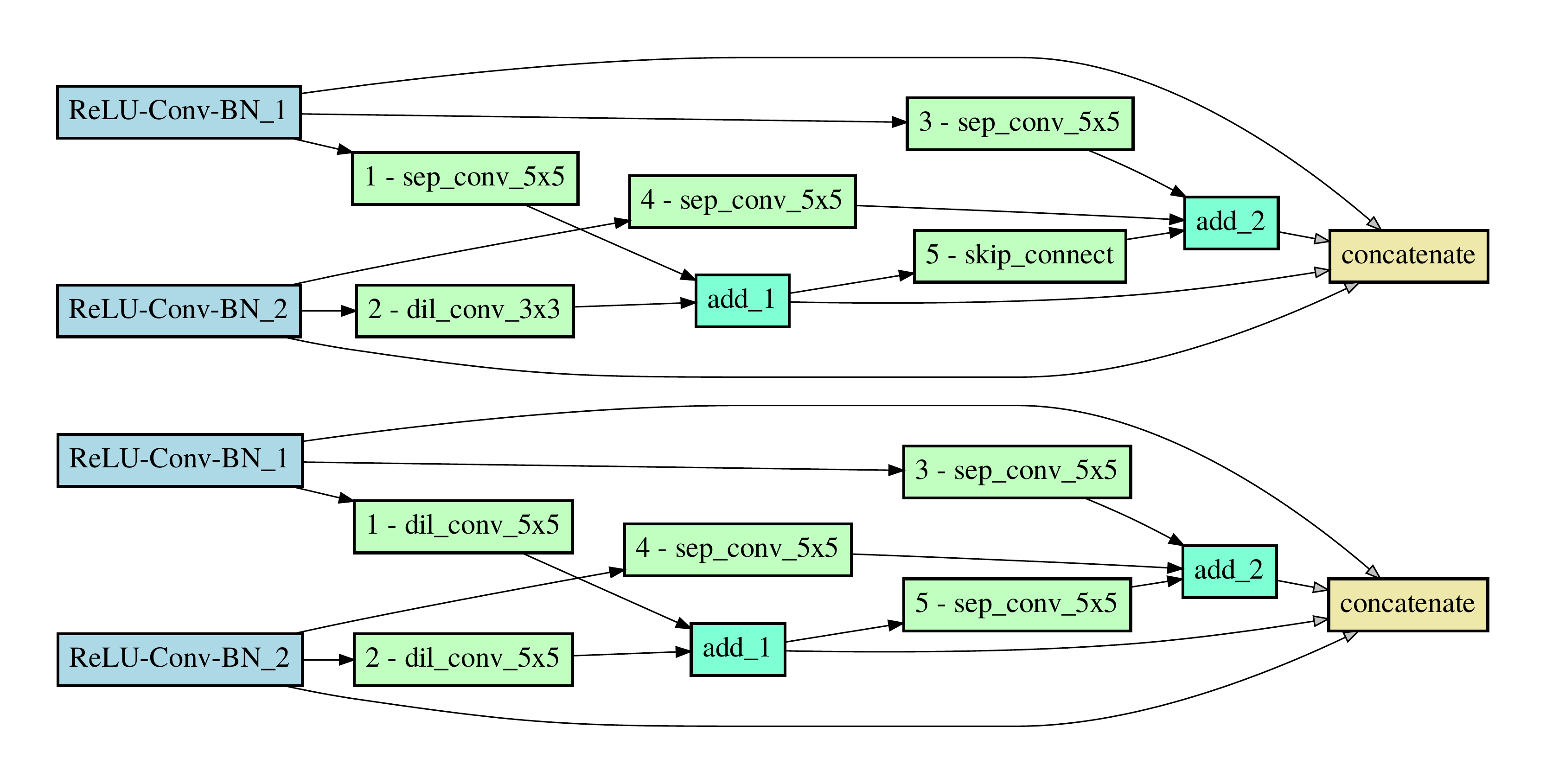}
}
\caption{Maximum likelihood architectures inferred by our search algorithm on CelebA. Shown are two samples taken from each block.}
\label{CelebA_NASSamples}
\end{figure}

\begin{figure}[h!]
\centering
\subfigure[Block 1]{
\includegraphics[width=0.40\textwidth]{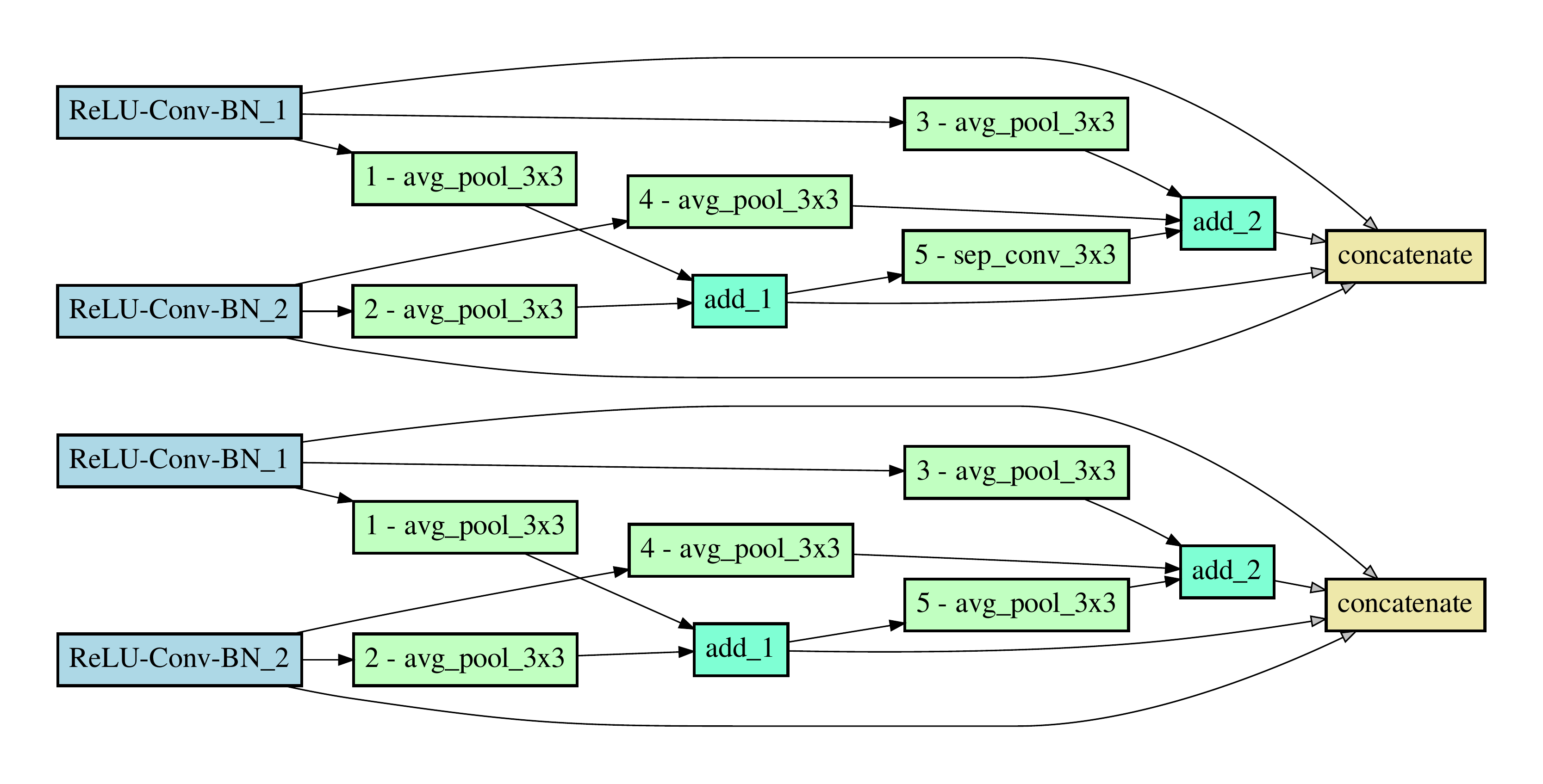}
}
\subfigure[Block 2]{
\includegraphics[width=0.40\textwidth]{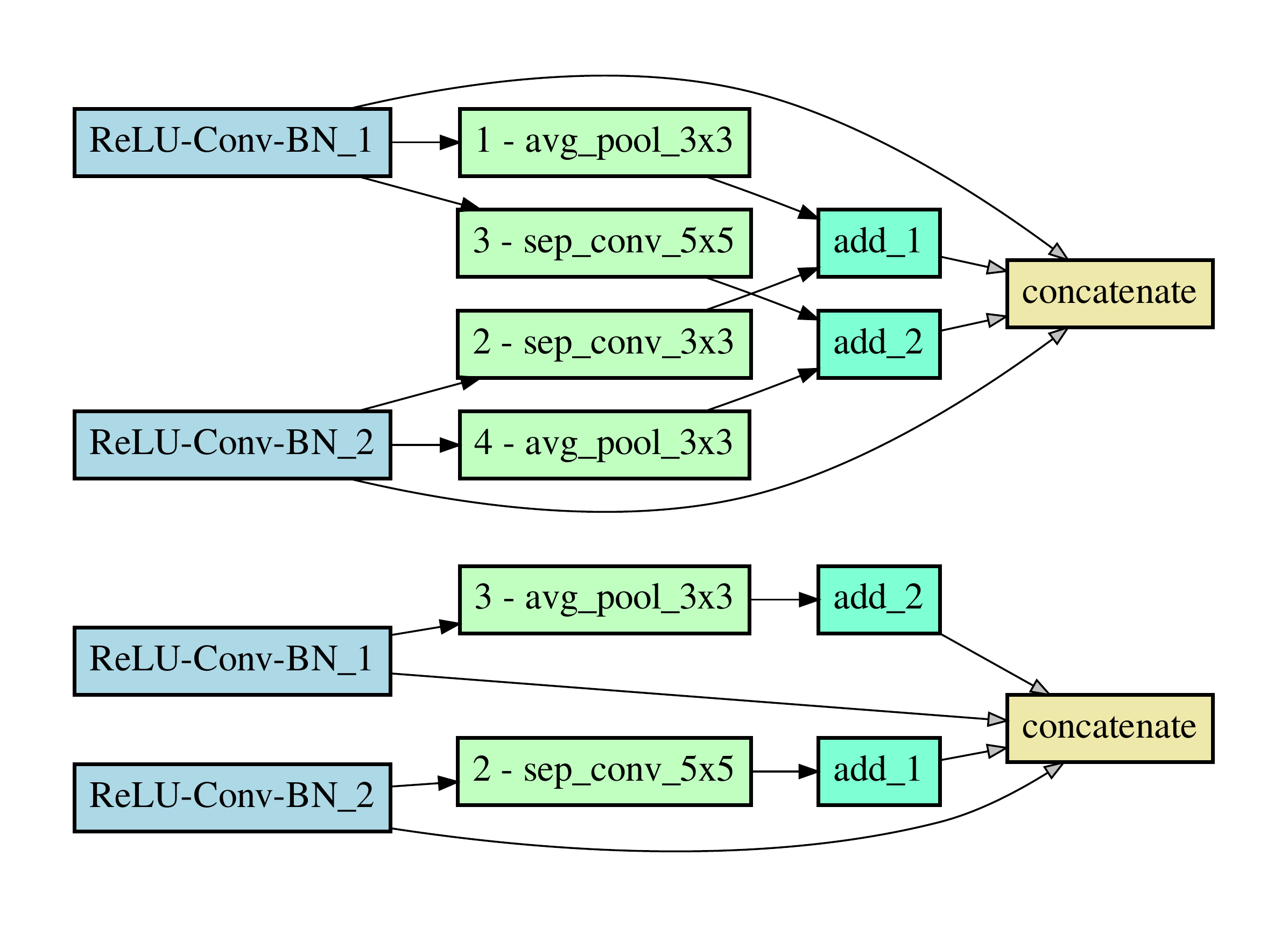}
}
\subfigure[Block 3]{
\includegraphics[width=0.40\textwidth]{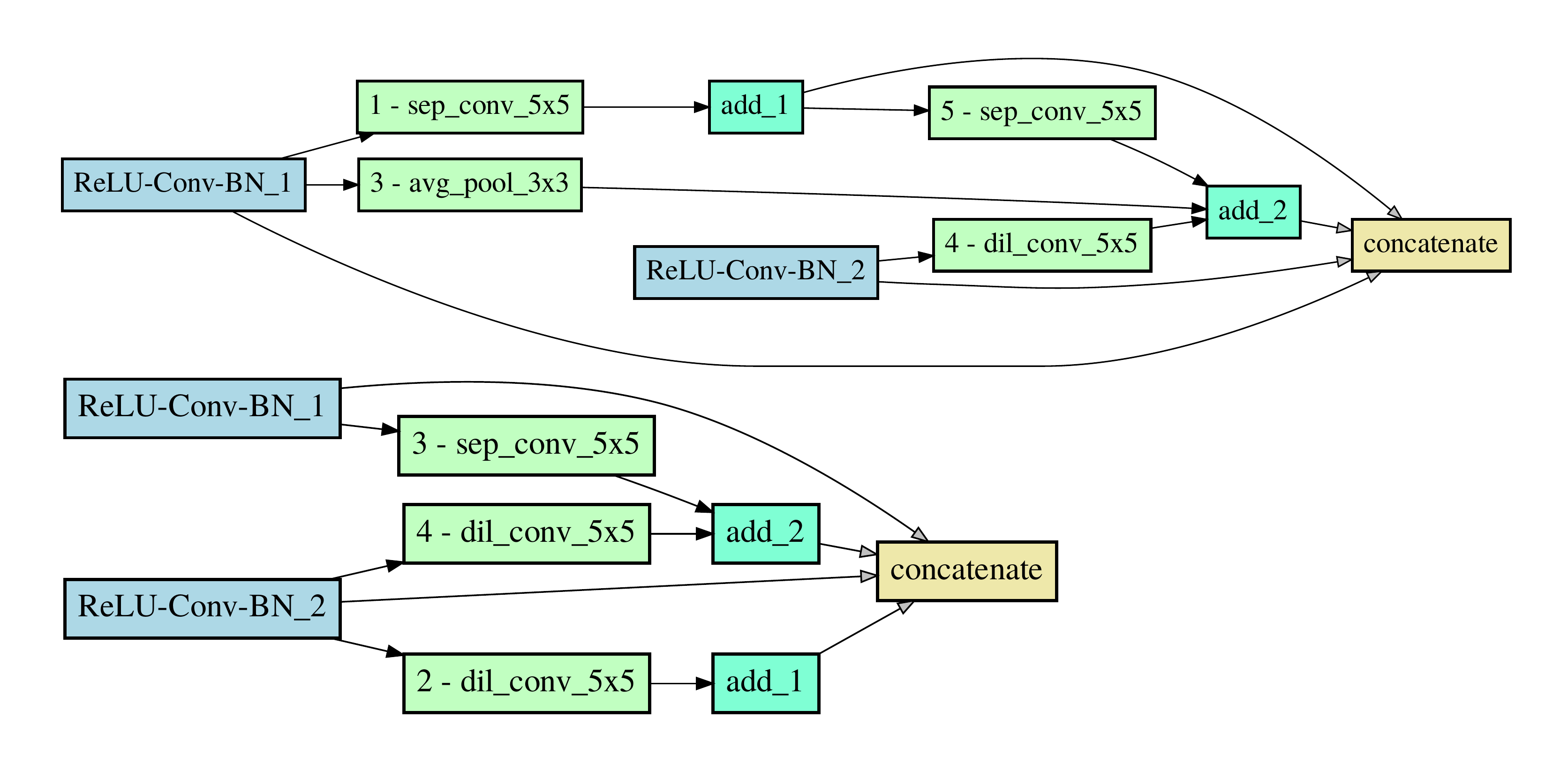}
}
\subfigure[Block 4]{
\includegraphics[width=0.40\textwidth]{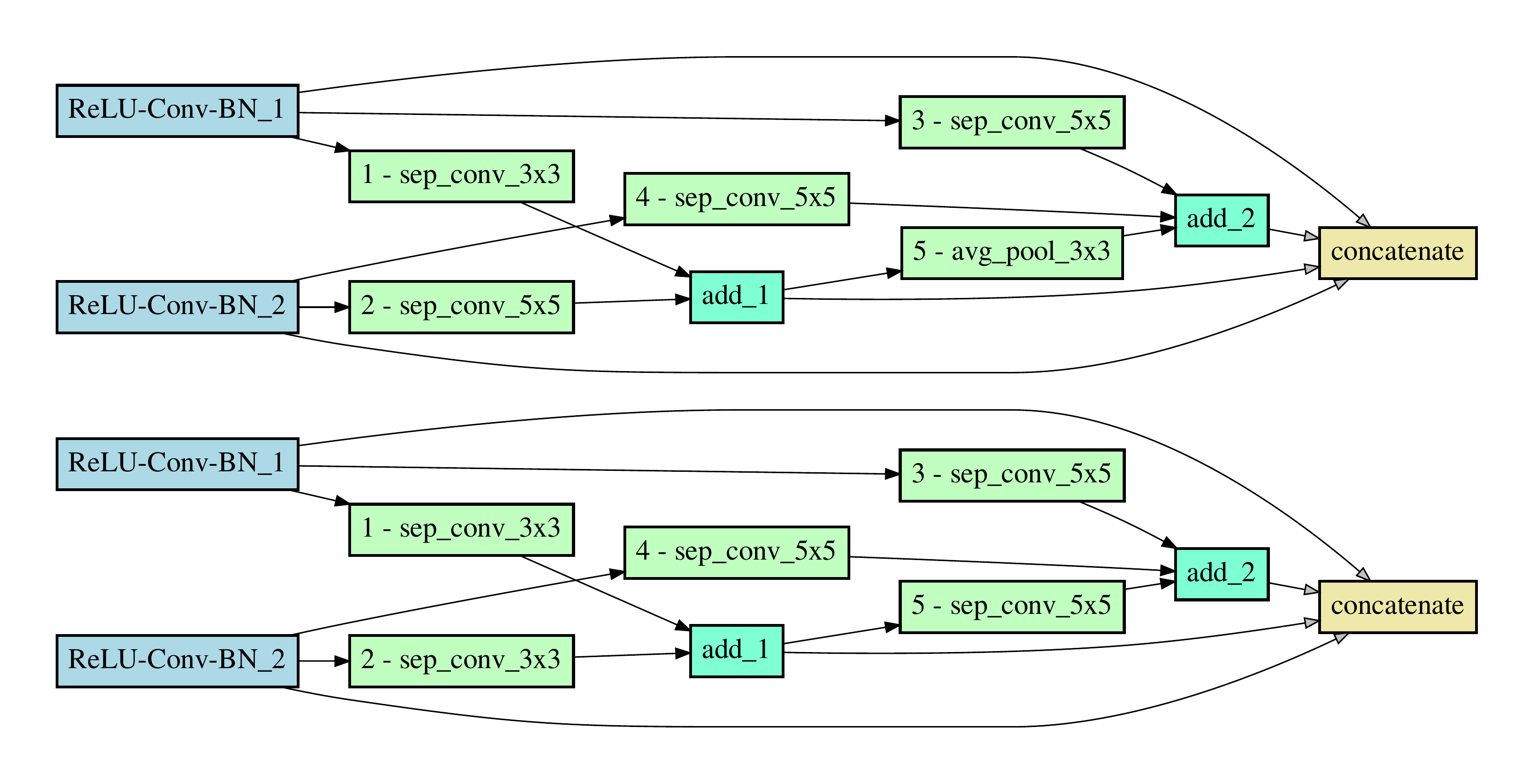}
}
\caption{Maximum likelihood architectures inferred by our search algorithm on MNIST. Shown are two samples taken from each block.}
\label{MNIST_NASSamples}
\end{figure}

\begin{figure}[h!]
\centering
\subfigure[Block 1]{
\includegraphics[width=0.40\textwidth]{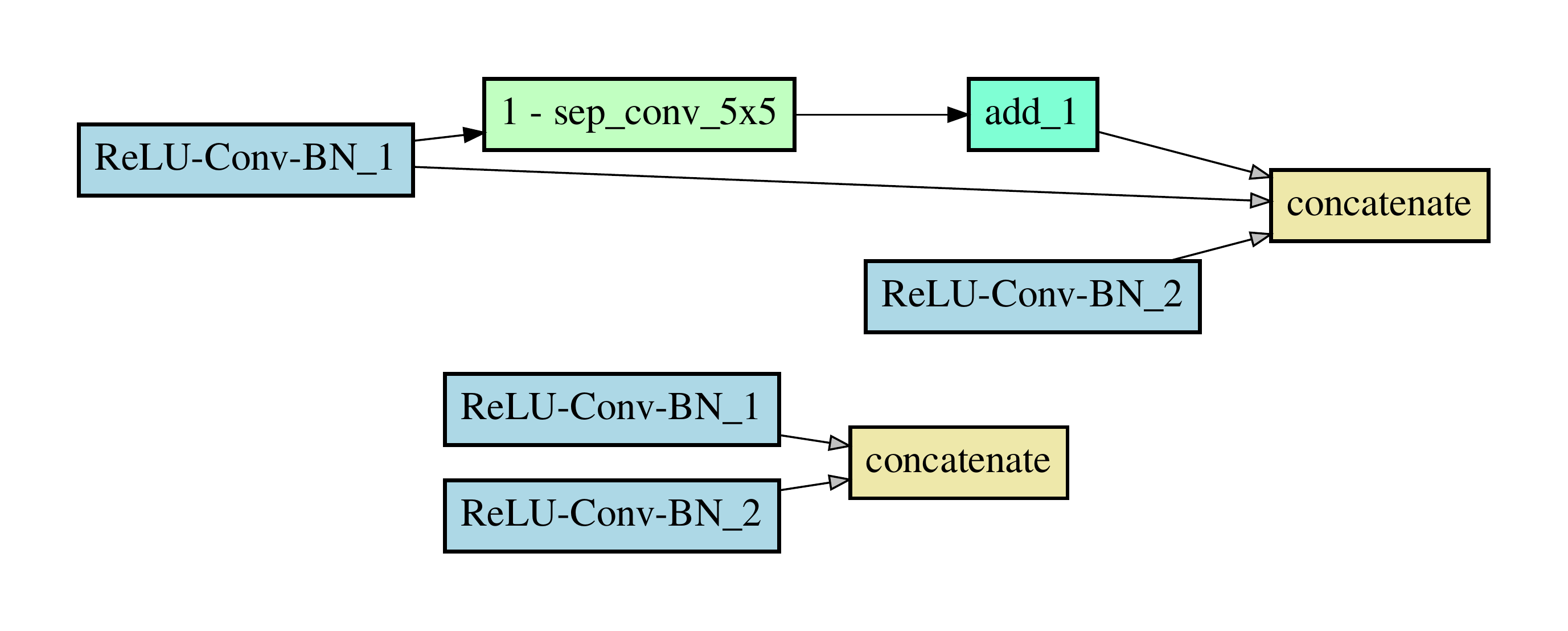}
}
\subfigure[Block 2]{
\includegraphics[width=0.40\textwidth]{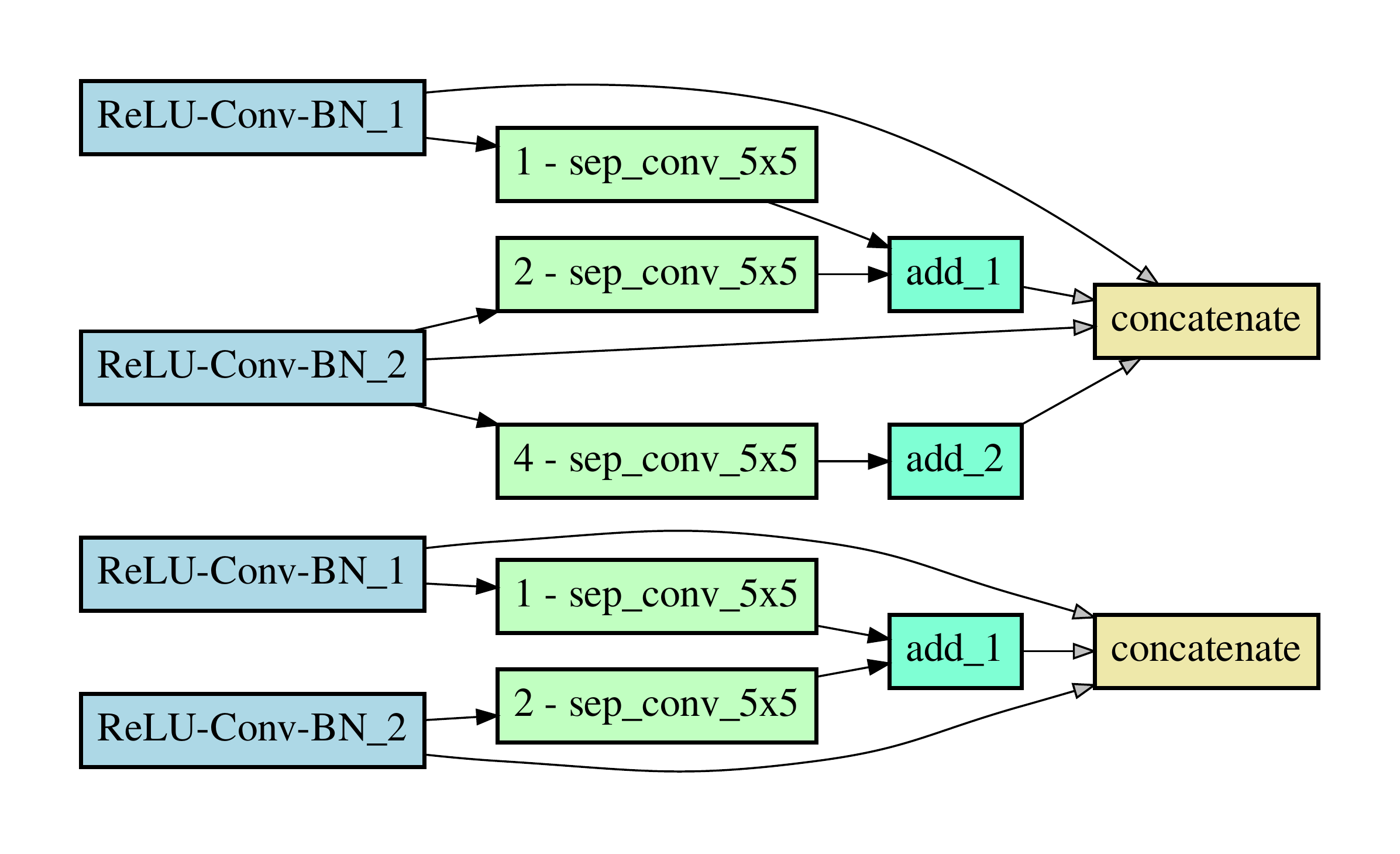}
}
\subfigure[Block 3]{
\includegraphics[width=0.40\textwidth]{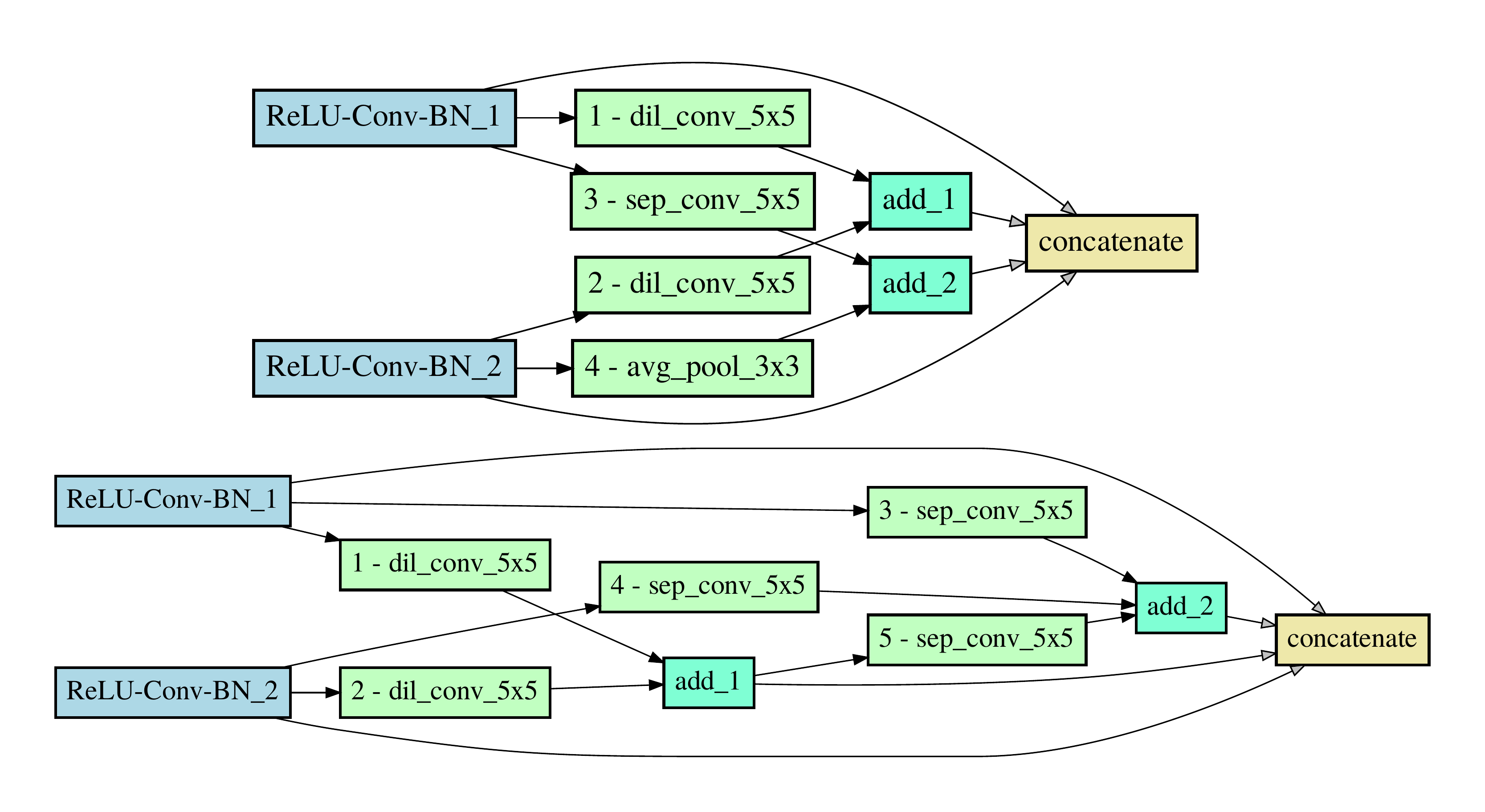}
}
\subfigure[Block 4]{
\includegraphics[width=0.40\textwidth]{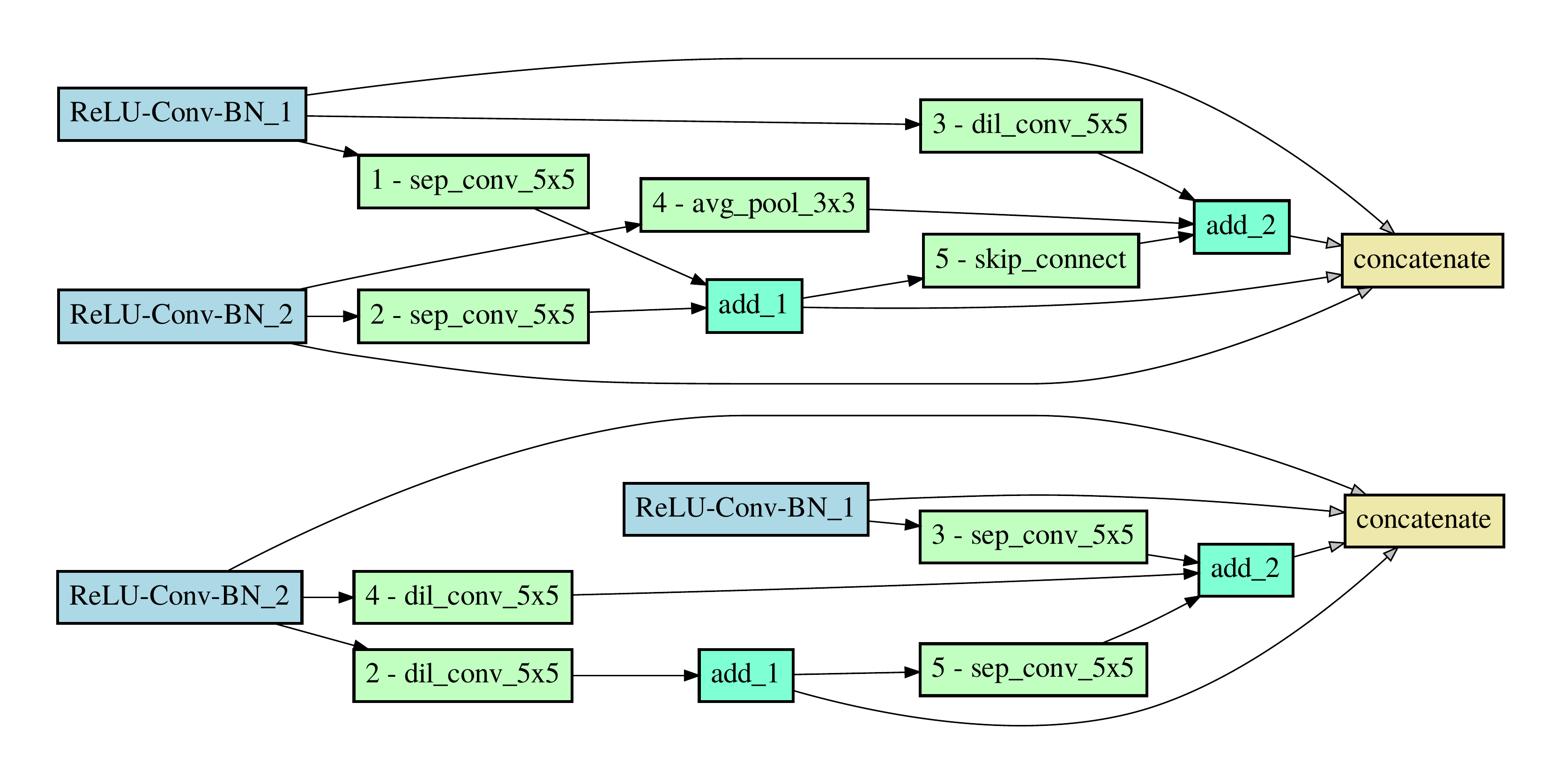}
}
\caption{Maximum likelihood architectures inferred by our search algorithm on SVHN. Shown are two samples taken from each block.}
\label{SVHN_NASSamples}
\end{figure}

\begin{figure}[h!]
\centering
\subfigure[Block 1]{
\includegraphics[width=0.40\textwidth]{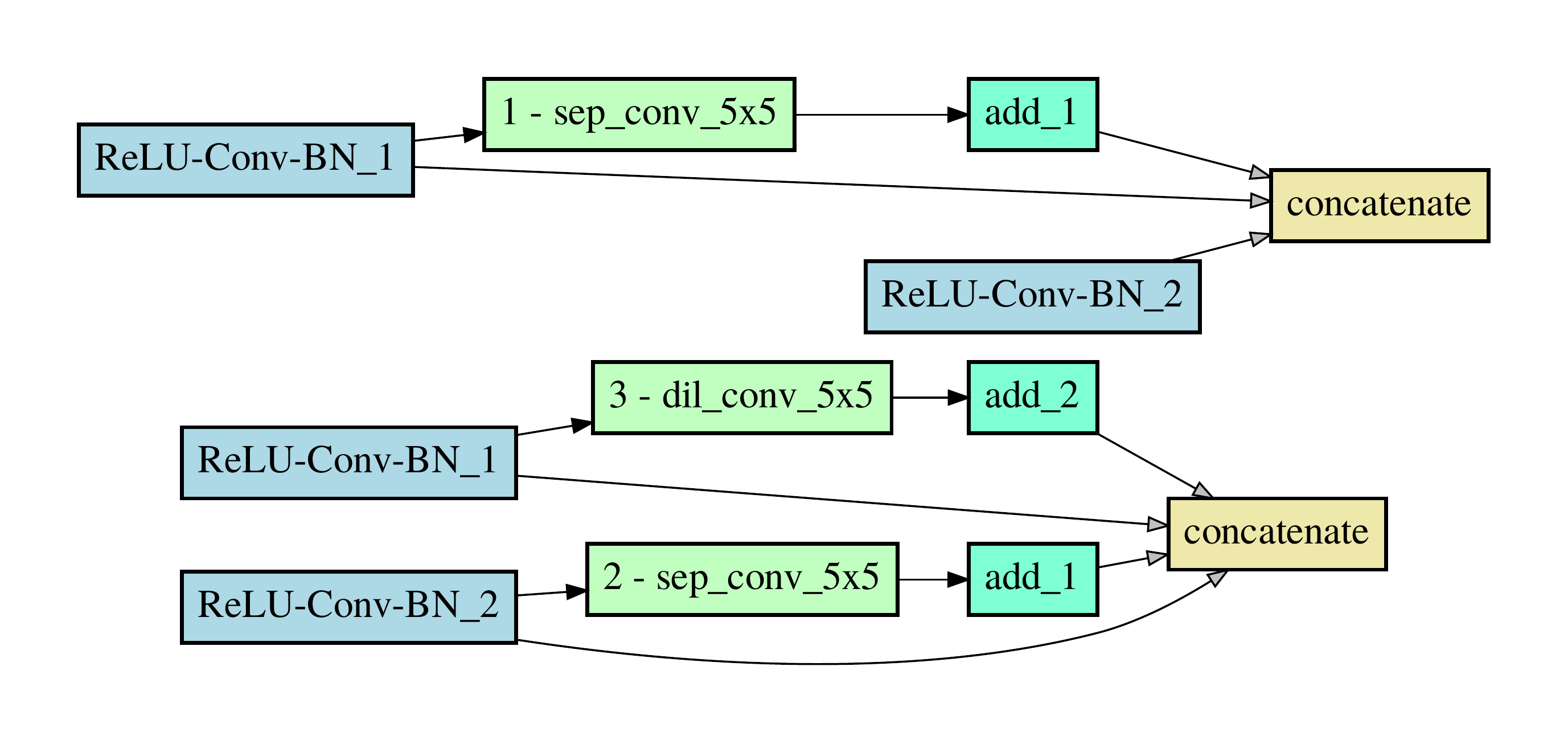}
}
\subfigure[Block 2]{
\includegraphics[width=0.40\textwidth]{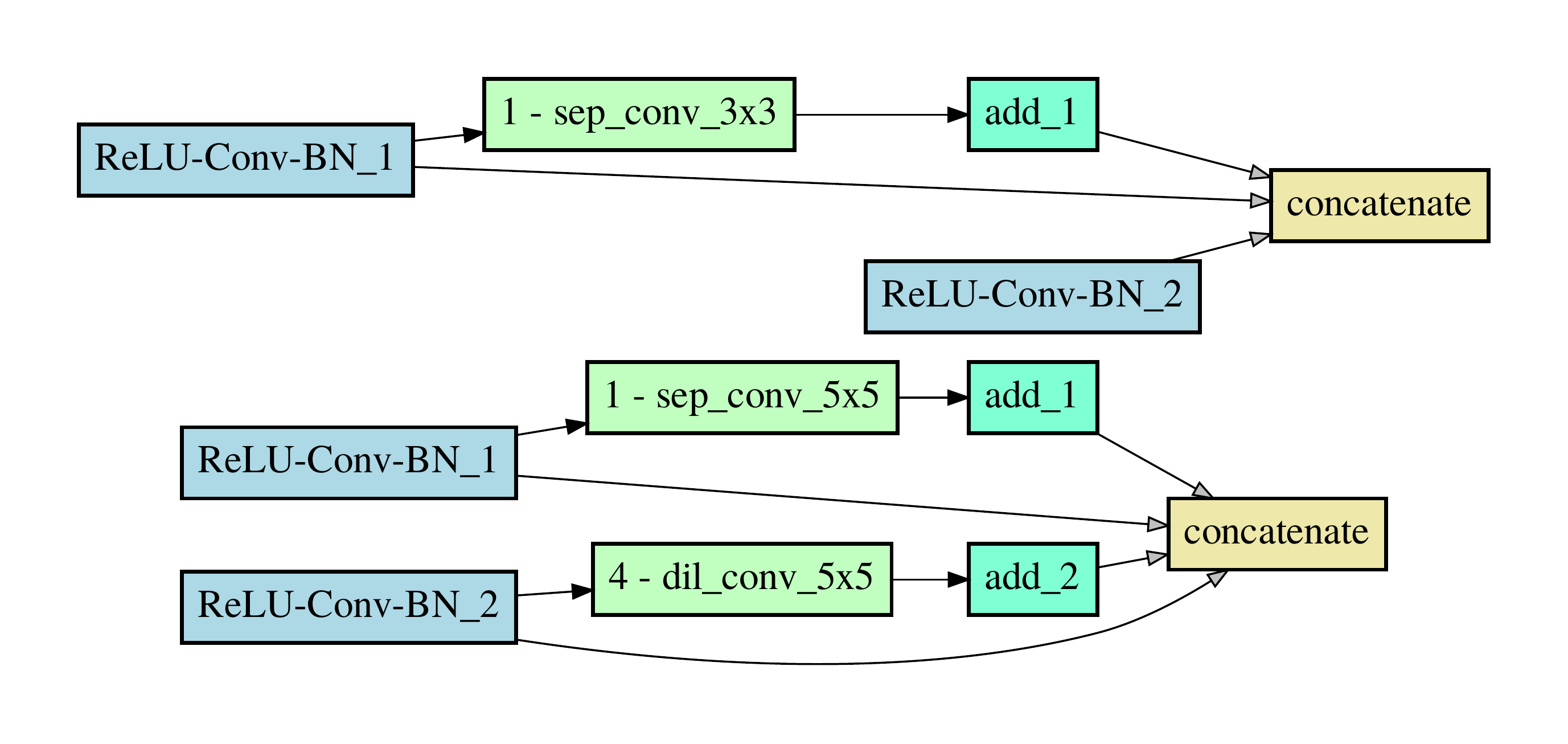}
}
\subfigure[Block 3]{
\includegraphics[width=0.40\textwidth]{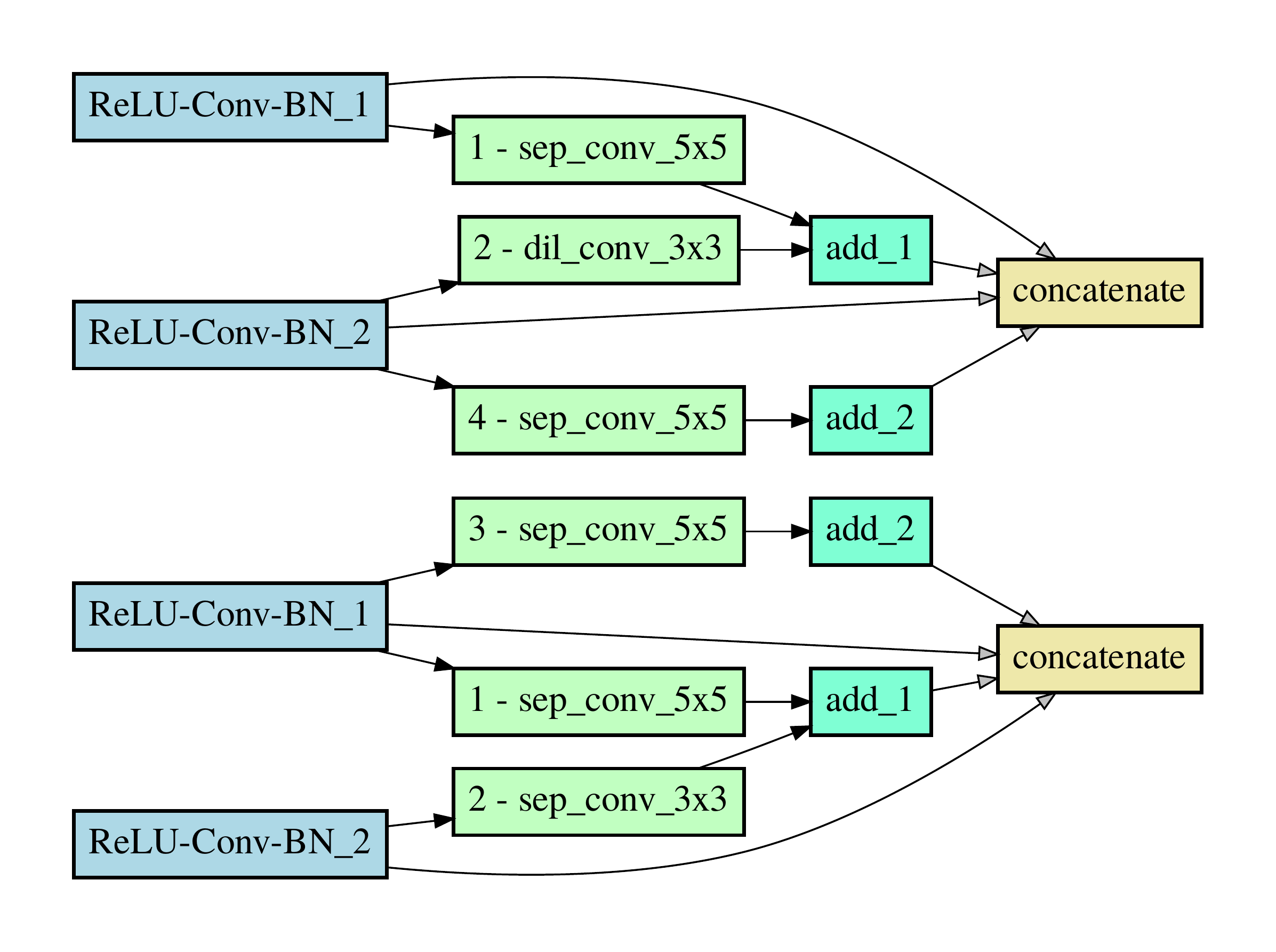}
}
\subfigure[Block 4]{
\includegraphics[width=0.40\textwidth]{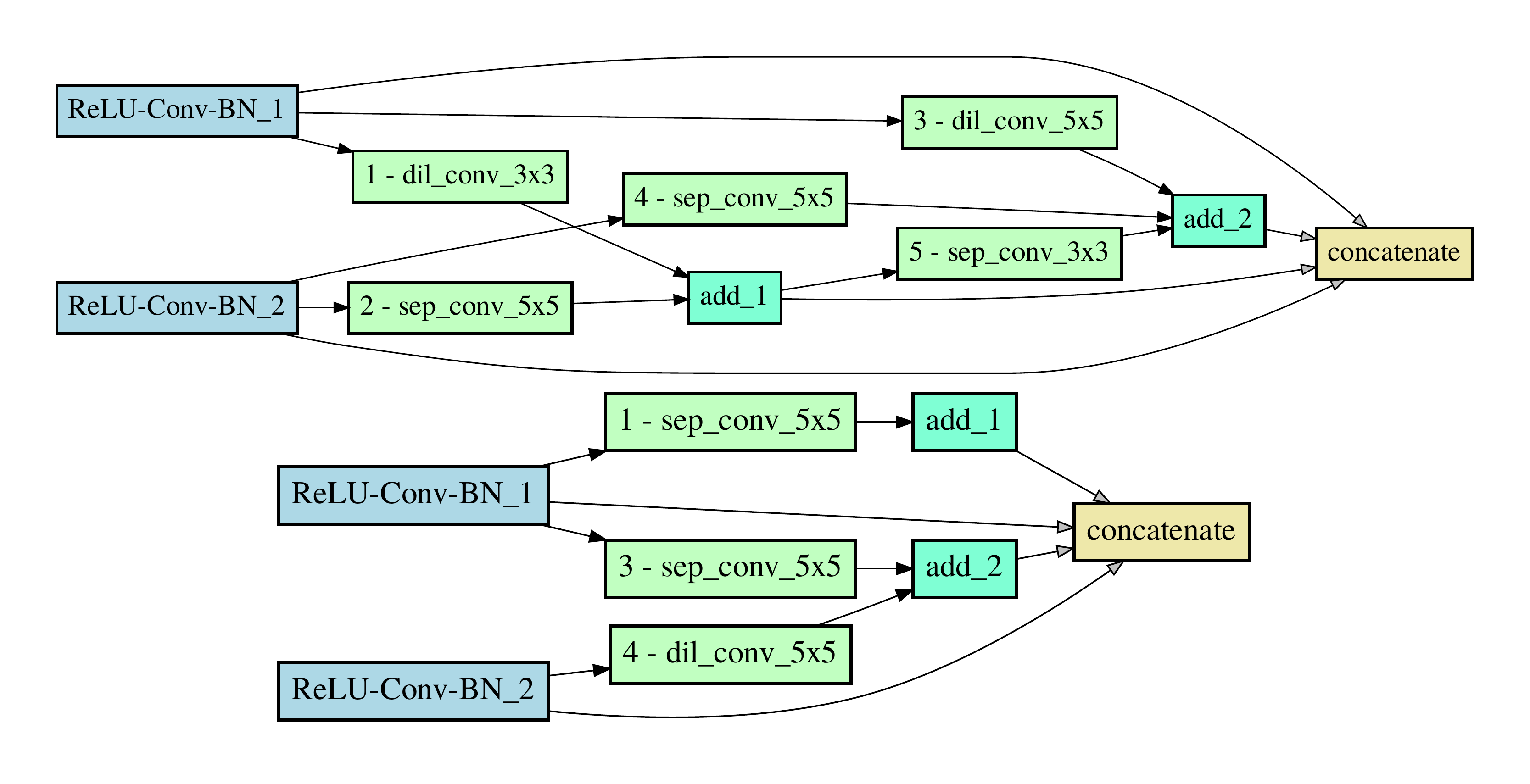}
}
\caption{Maximum likelihood architectures inferred by our search algorithm on CIFAR-10. Shown are two samples taken from each block.}
\label{CIFAR_NASSamples}
\end{figure}

\clearpage
\FloatBarrier

\section{Image generation samples}
\label{append:image_generation_samples}

\begin{figure}[ht!]
  \centering
  \includegraphics[width=0.8\textwidth]{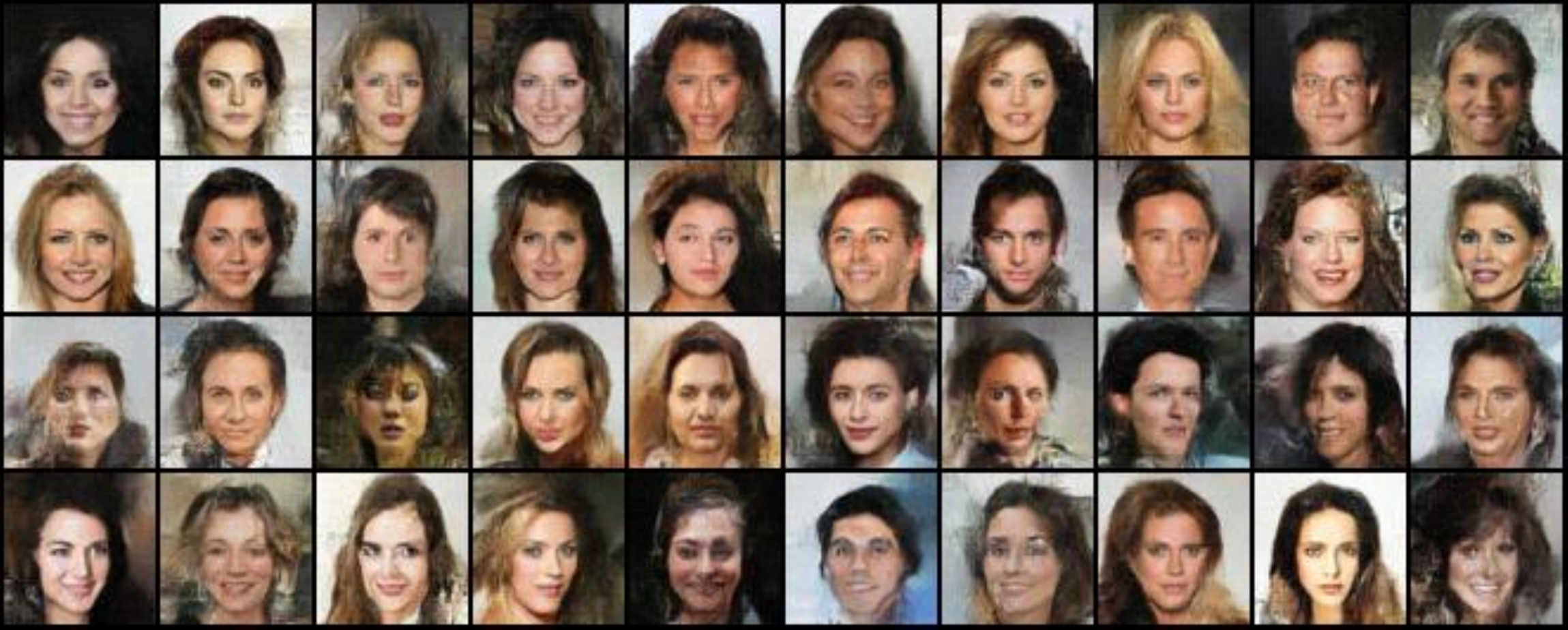}
  \caption{Samples taken from randomly sampled NADS architectures searched on CelebA. Images were not cherry-picked and the architectures were sampled without further retraining.}
  \label{sample_celeba}
\end{figure}

\begin{figure}[ht!]
  \centering
  \includegraphics[width=0.8\textwidth]{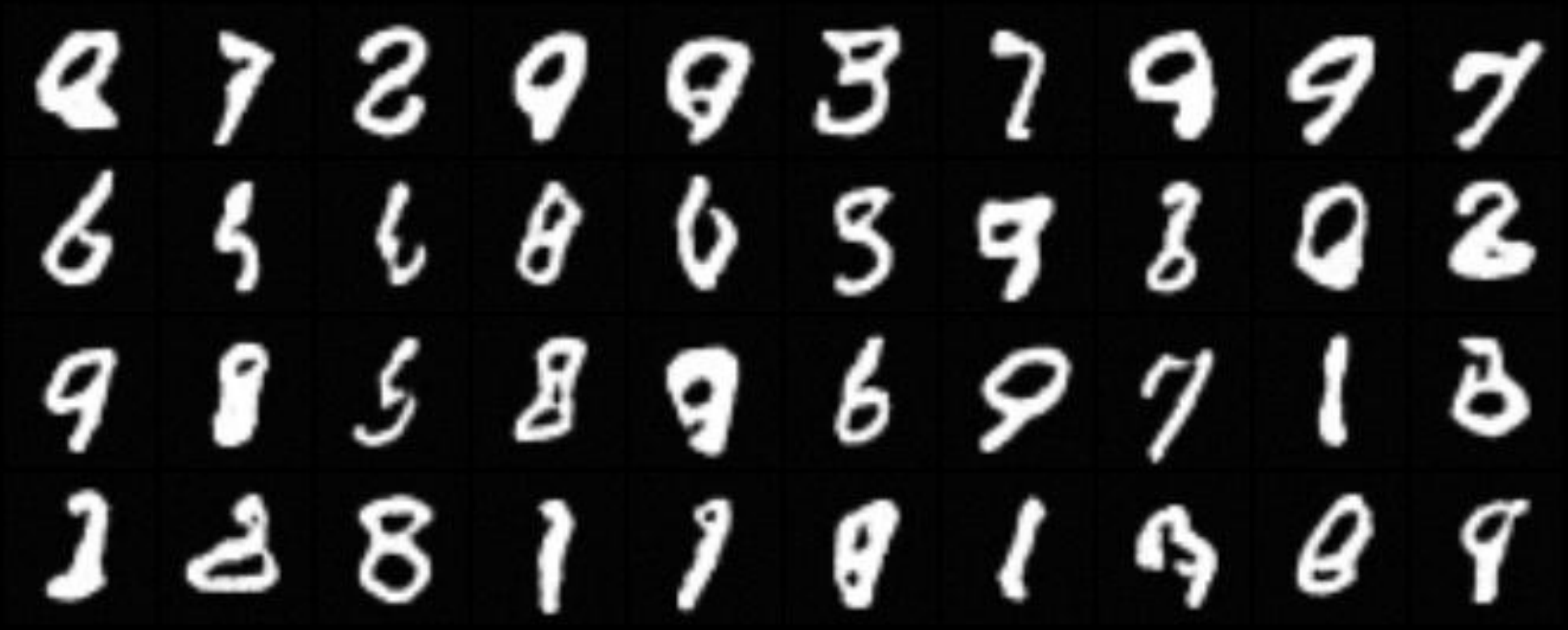}
  \caption{Samples taken from randomly sampled NADS architectures searched on MNIST. Images were not cherry-picked and the architectures were sampled without further retraining.}
  \label{sample_mnist}
\end{figure}

\begin{figure}[ht!]
  \centering
  \includegraphics[width=0.8\textwidth]{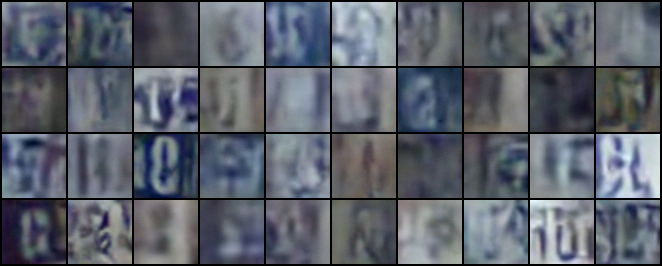}
  \caption{Samples taken from randomly sampled NADS architectures searched on SVHN. Images were not cherry-picked and the architectures were sampled without further retraining.}
  \label{sample_svhn}
\end{figure}

\begin{figure}[ht!]
  \centering
  \includegraphics[width=0.8\textwidth]{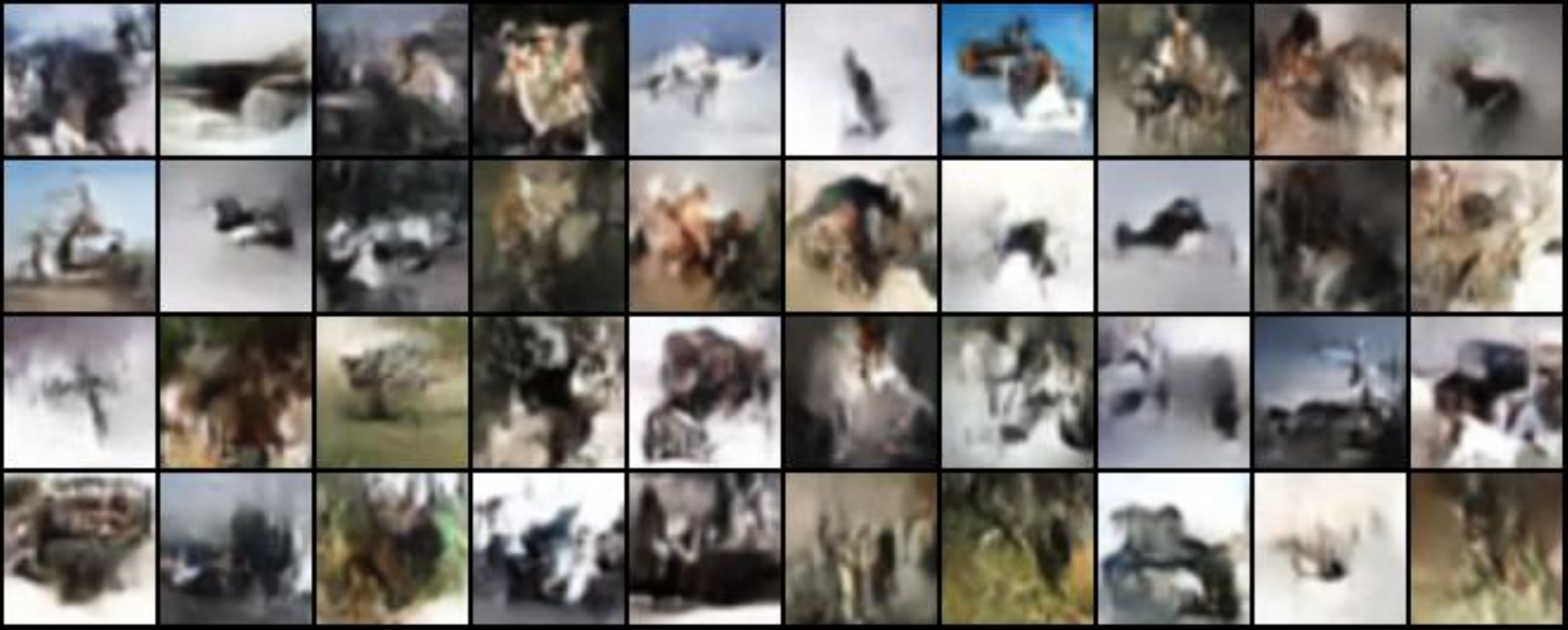}
  \caption{Samples taken from randomly sampled NADS architectures searched on CIFAR-10. Images were not cherry-picked and the architectures were sampled without further retraining.}
  \label{cifar10_samples}
\end{figure}

\FloatBarrier

\section{Likelihood Estimation Models Assign Higher Likelihood to OoD Data}

\begin{figure}[h!]
  \centering
  \includegraphics[width=0.8\textwidth]{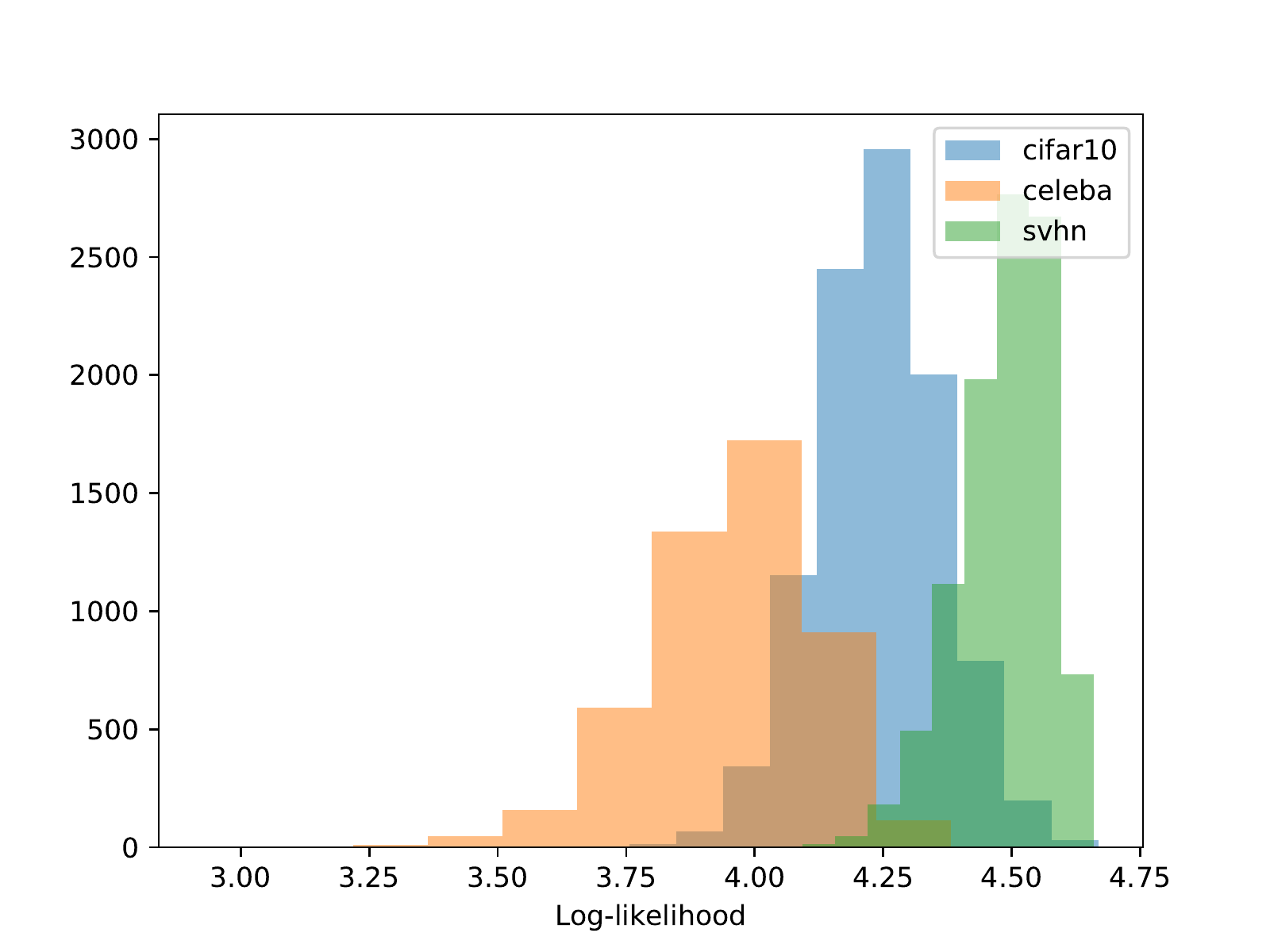}
  \caption{Likelihood distributions of different datasets evaluated on a Glow model trained on CelebA. The model assigns higher likelihood to OoD samples from CIFAR-10 and SVHN.}
  \label{celeba_motivating}
\end{figure}

\clearpage

\section{Effect of Ensemble Size}
\label{append:ensemble-effect}

\begin{figure}[h!]
\centering
\includegraphics[width=0.775\textwidth]{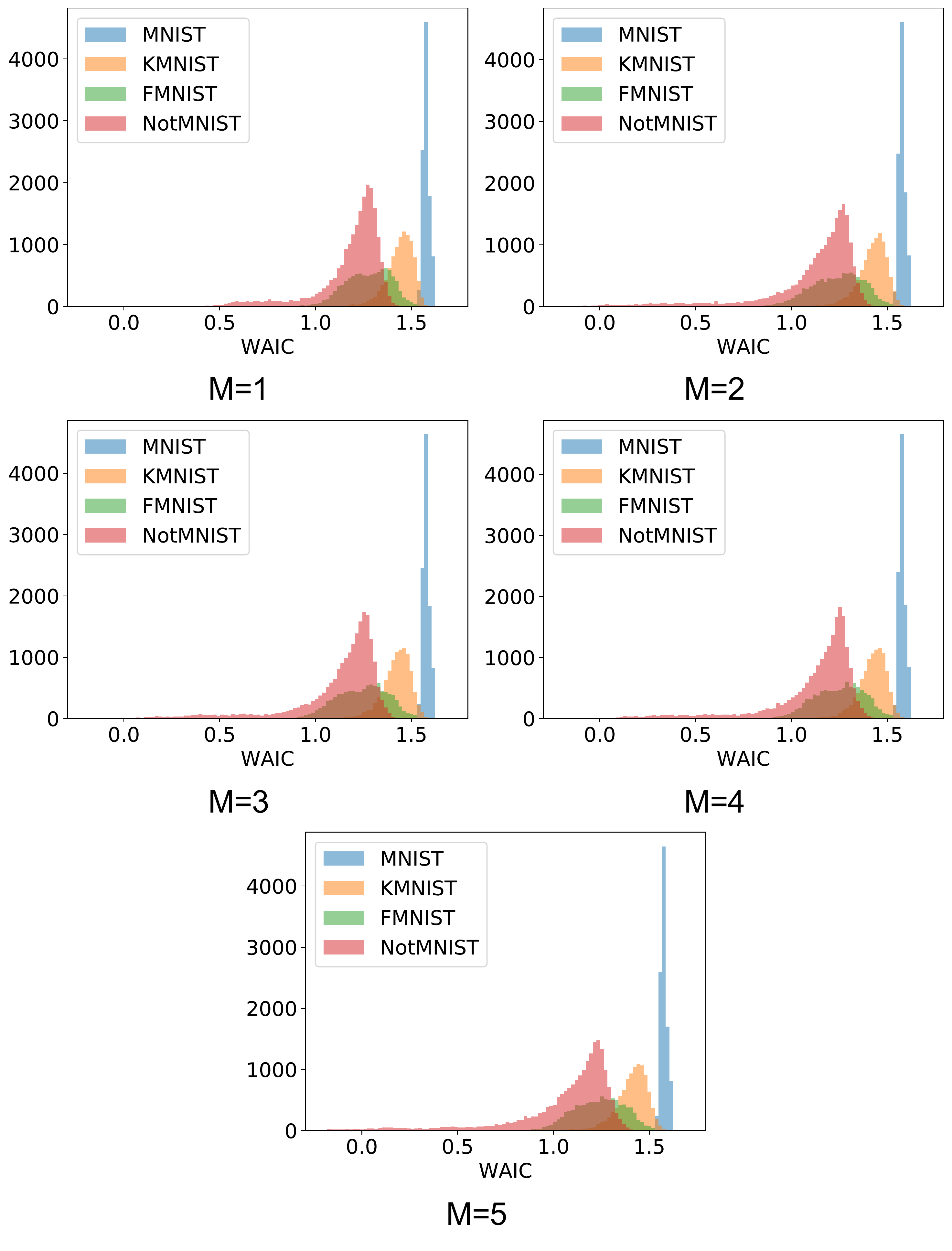}
\label{mnist-hist}
\caption{Effect of ensemble size to the distribution of WAIC scores estimated by model ensembles trained on MNIST.}
\end{figure}

\begin{figure}[h!]
\centering
\includegraphics[width=0.975\textwidth]{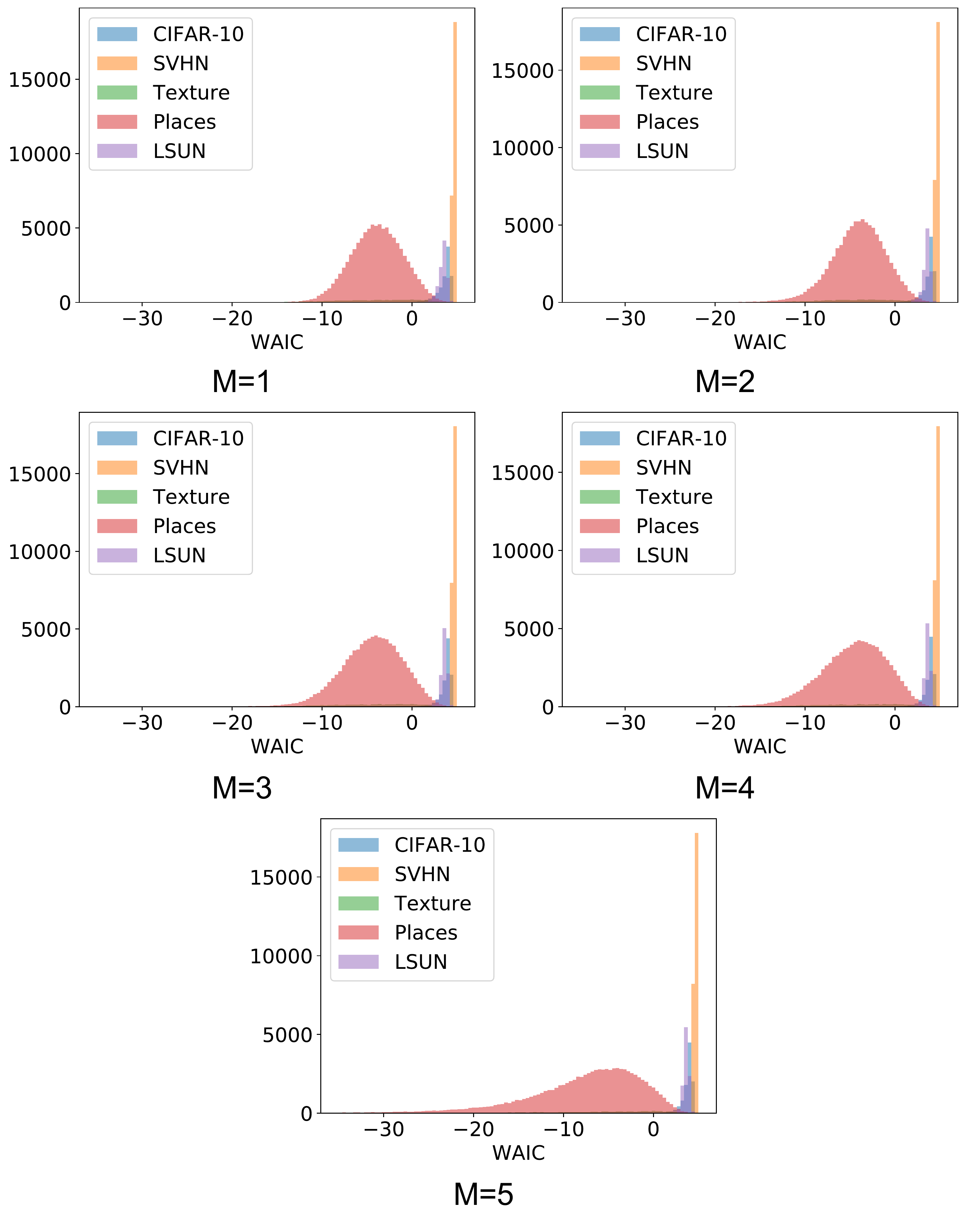}
\label{svhn-hist}
\caption{Effect of ensemble size to the distribution of WAIC scores estimated by model ensembles trained on SVHN.}
\end{figure}

\begin{figure}[h!]
\centering
\includegraphics[width=0.975\textwidth]{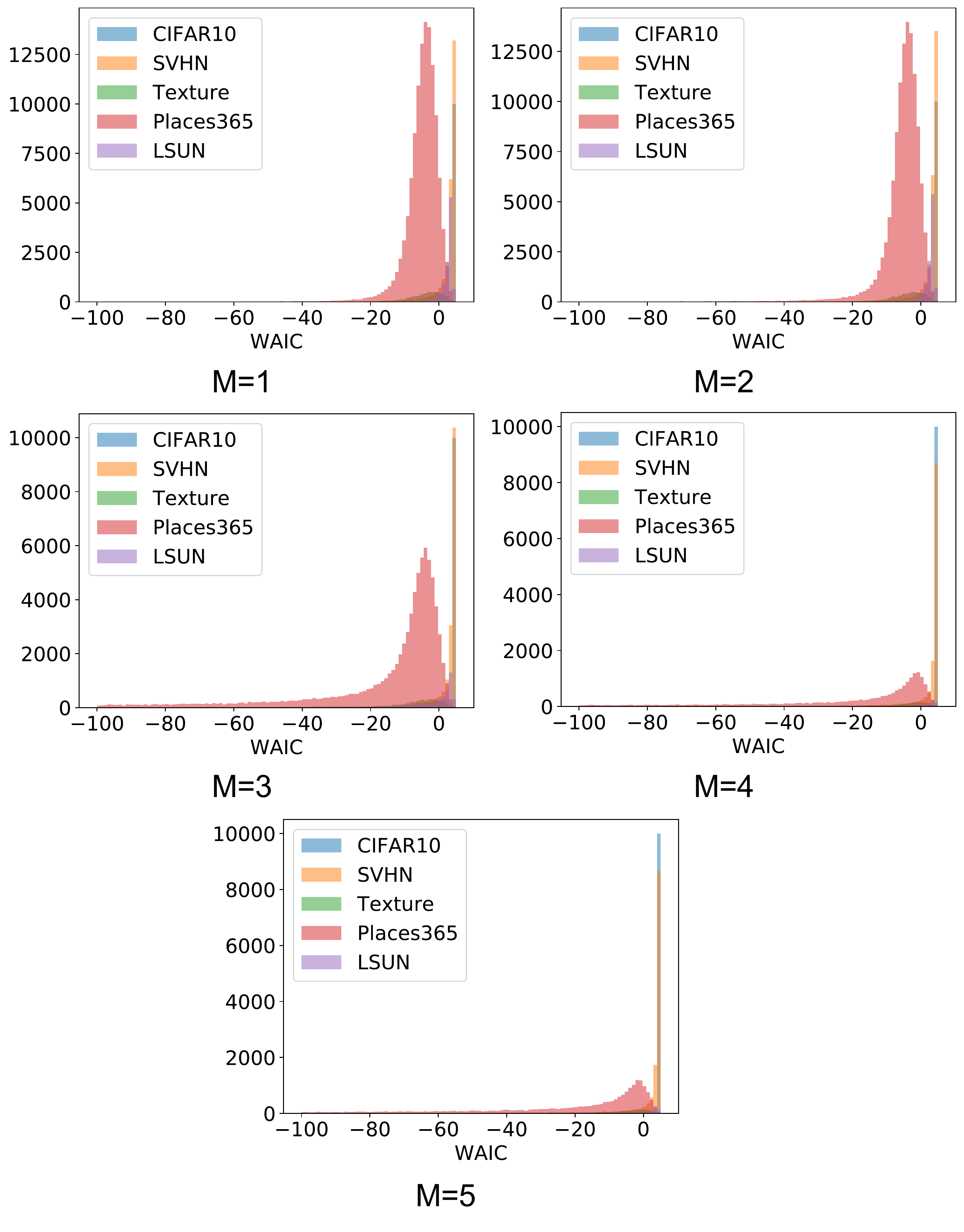}
\label{cifar10-hist}
\caption{Effect of ensemble size to the distribution of WAIC scores estimated by model ensembles trained on CIFAR-10.}
\end{figure}



\FloatBarrier

\section{Additional ROC and Precision-Recall Curves}
\label{append:additional-fpr}

\begin{figure}[h!]
\centering
\subfigure{
\includegraphics[width=0.23\textwidth]{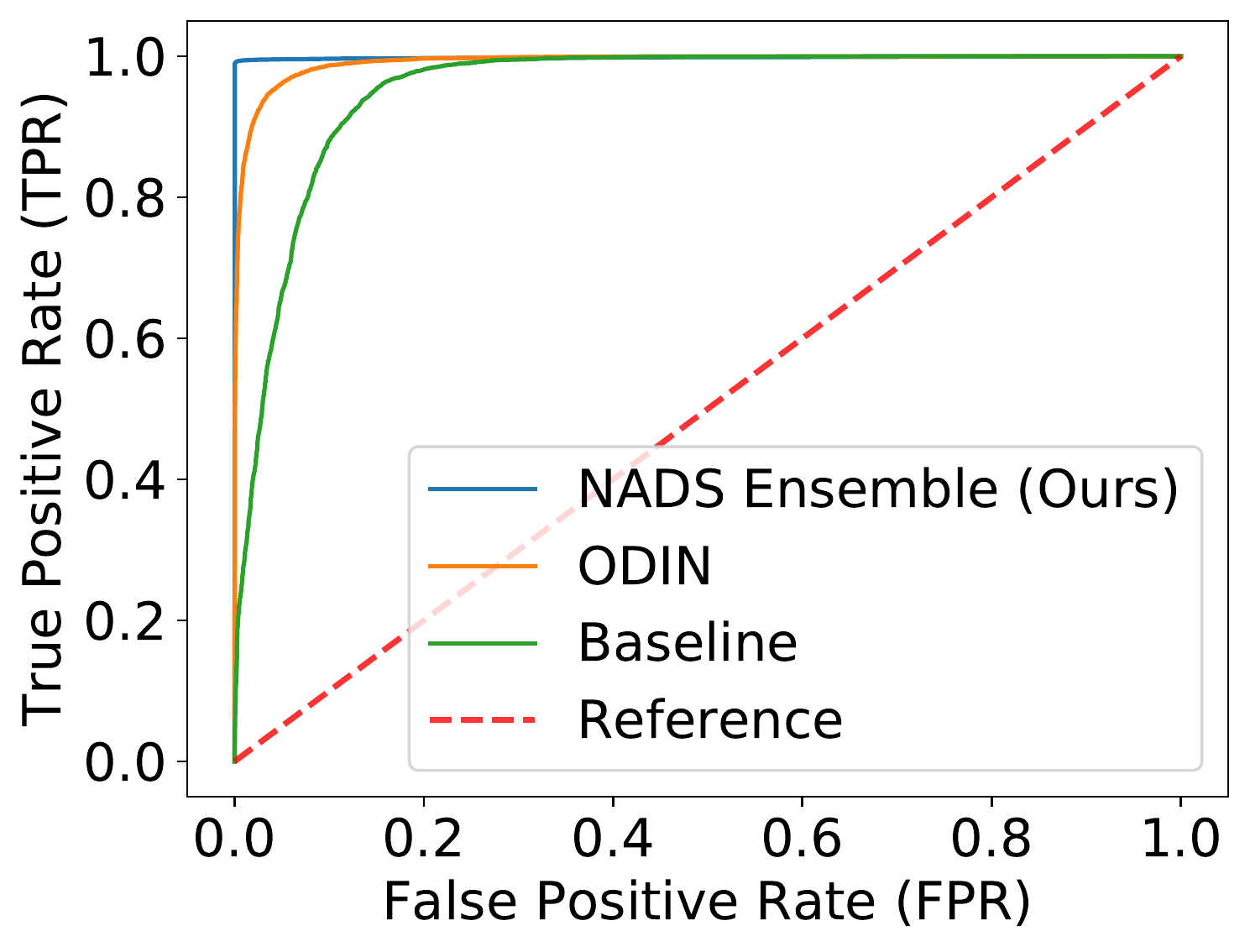}
}
\subfigure{
\includegraphics[width=0.23\textwidth]{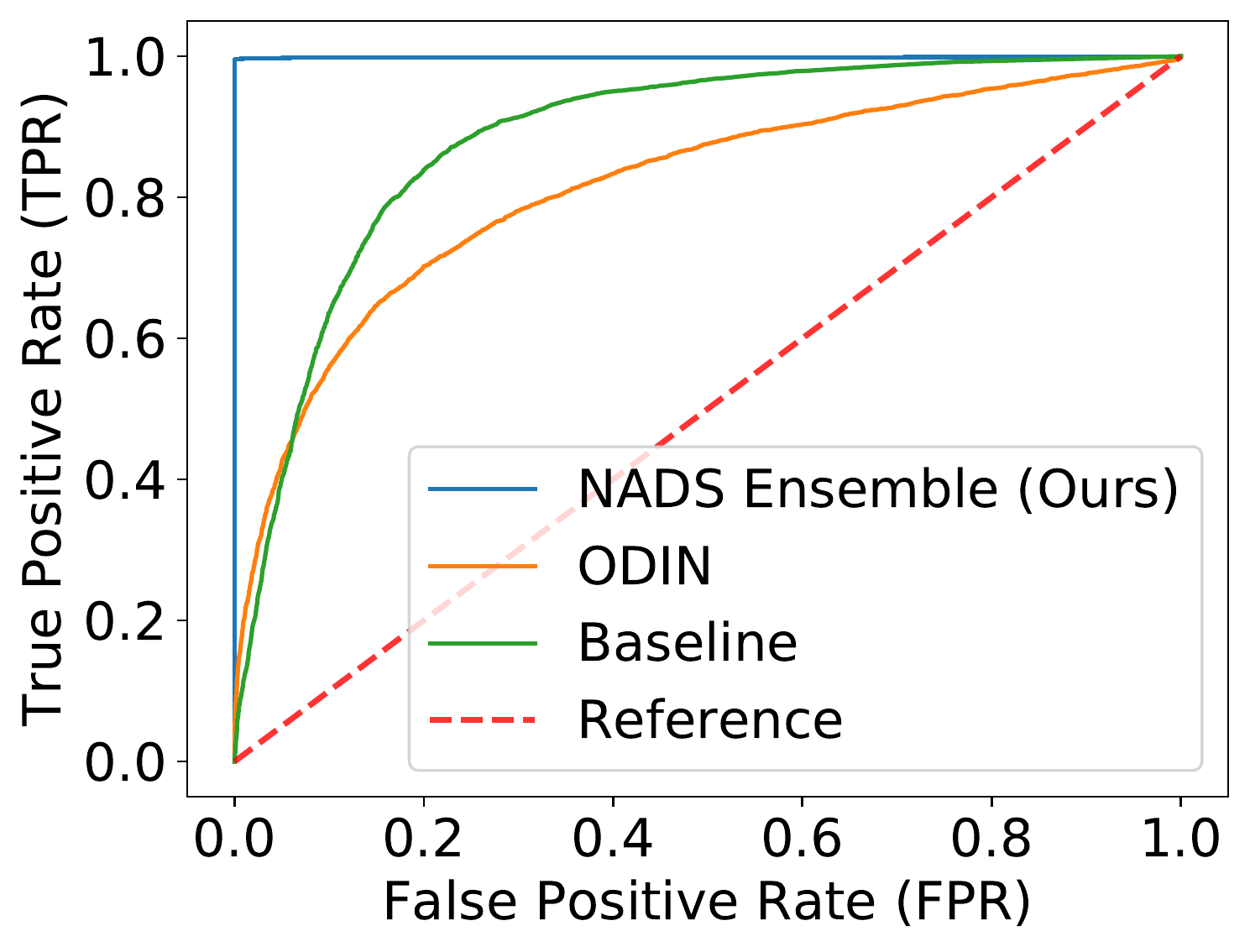}
}
\subfigure{
\includegraphics[width=0.23\textwidth]{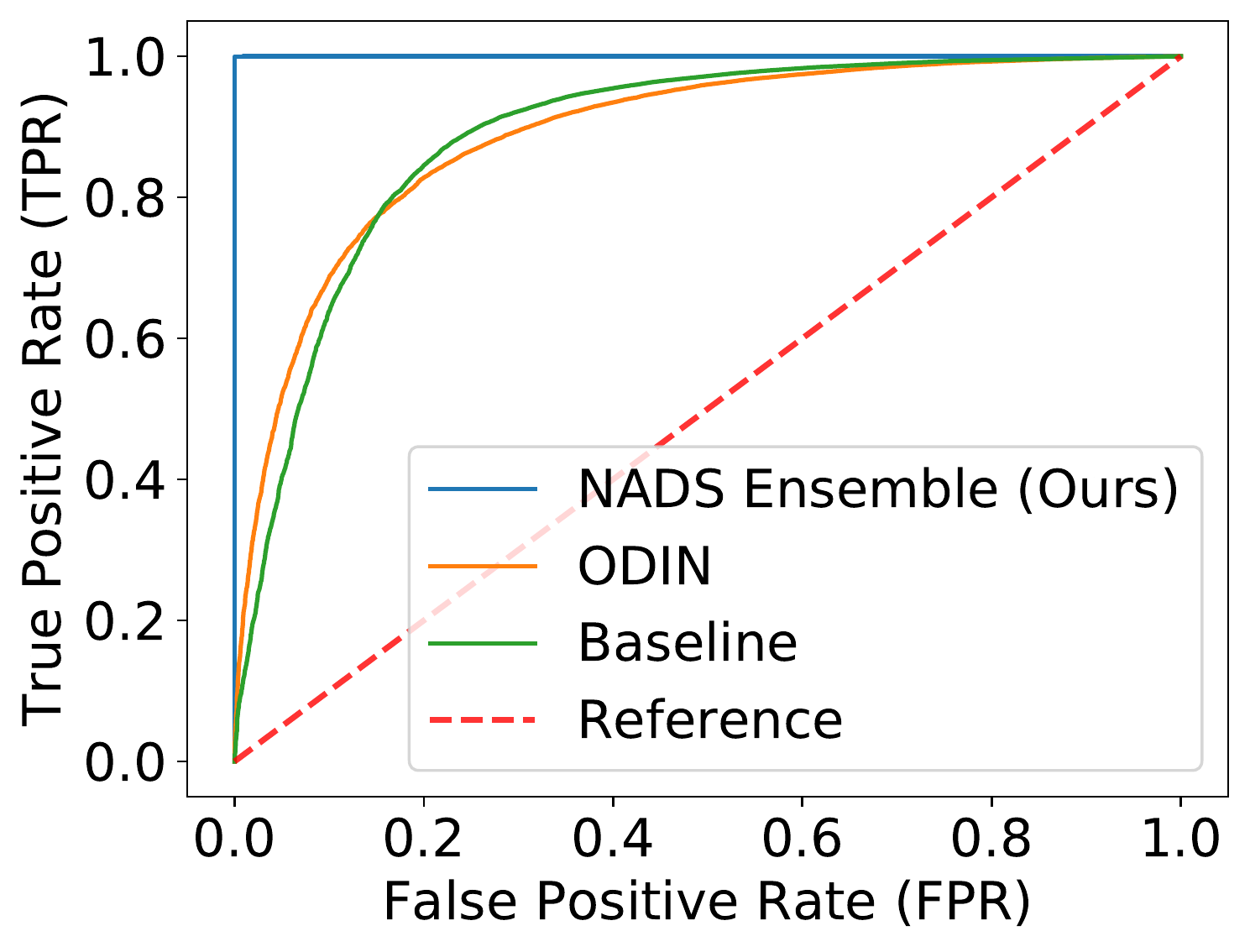}
}
\subfigure{
\includegraphics[width=0.23\textwidth]{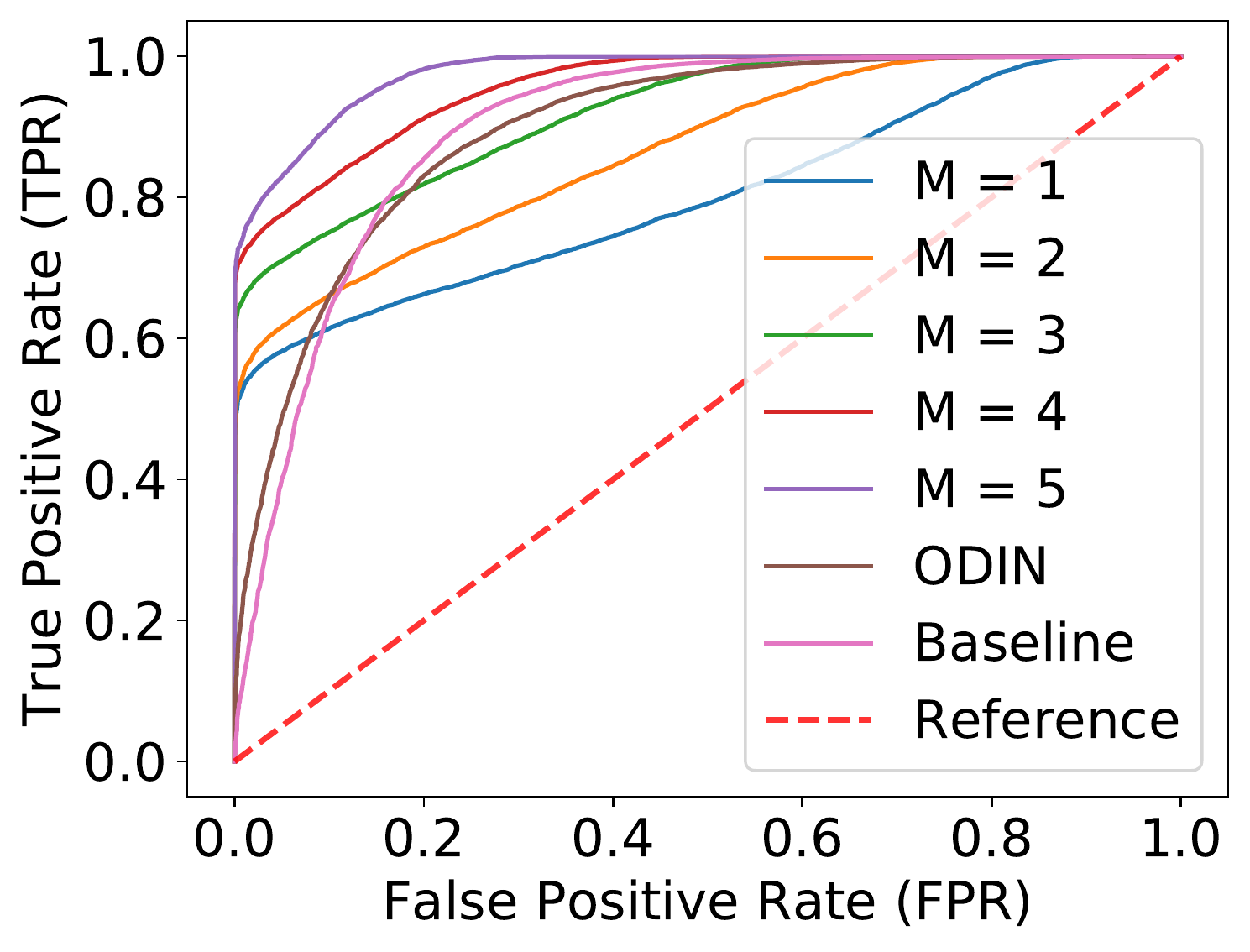}
}
\setcounter{subfigure}{0}
\subfigure[LSUN]{
\includegraphics[width=0.23\textwidth]{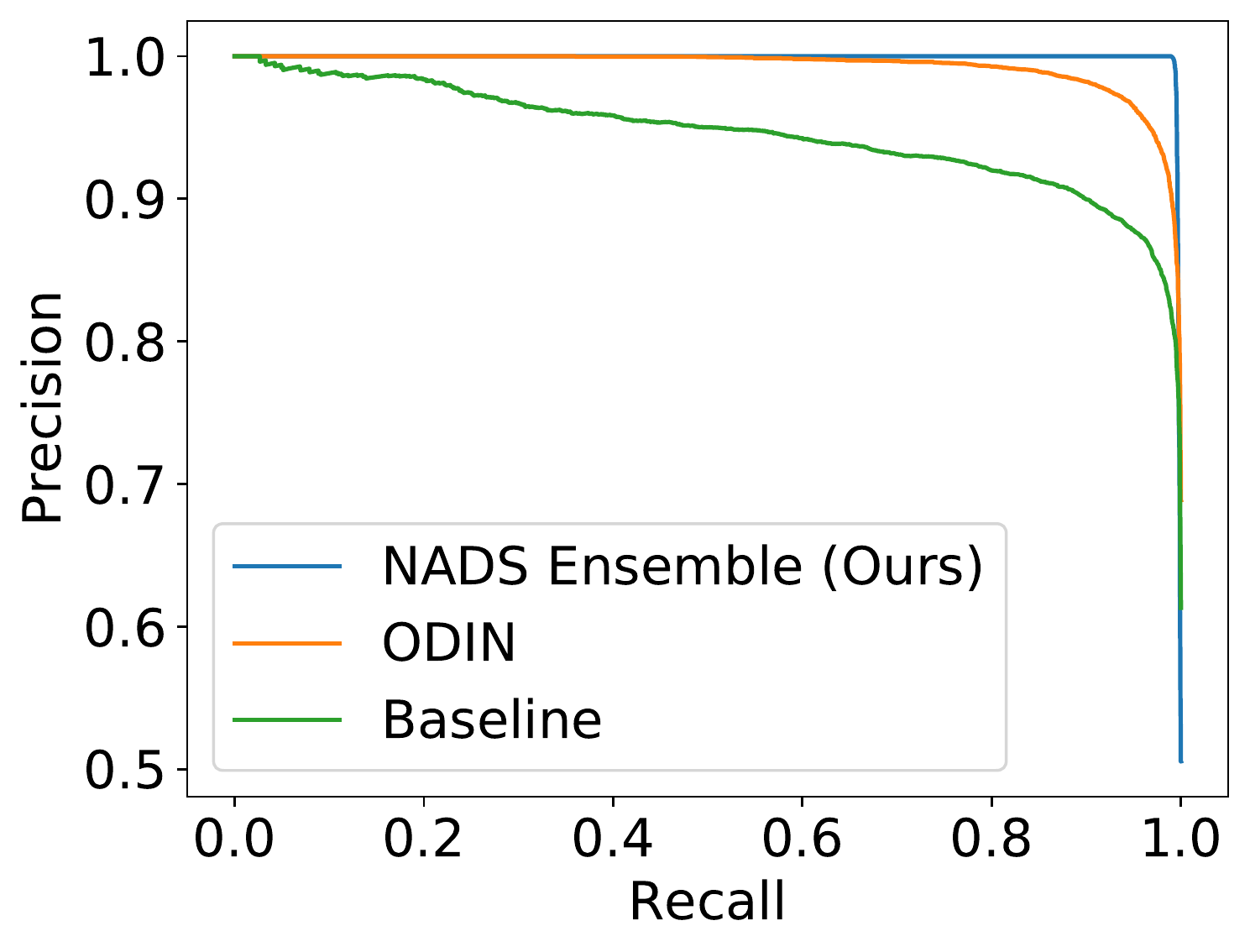}
}
\subfigure[Texture]{
\includegraphics[width=0.23\textwidth]{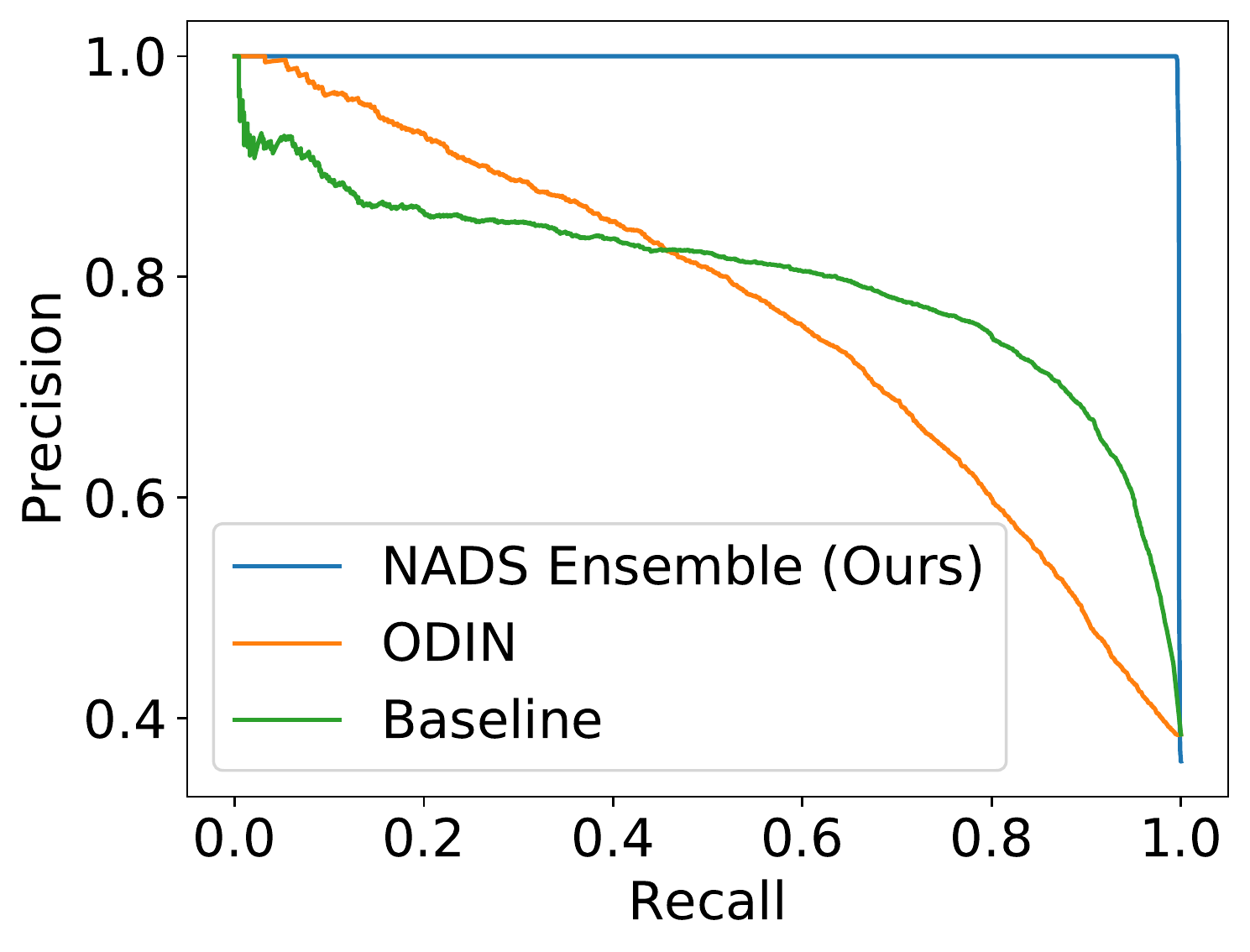}
}
\subfigure[Places]{
\includegraphics[width=0.23\textwidth]{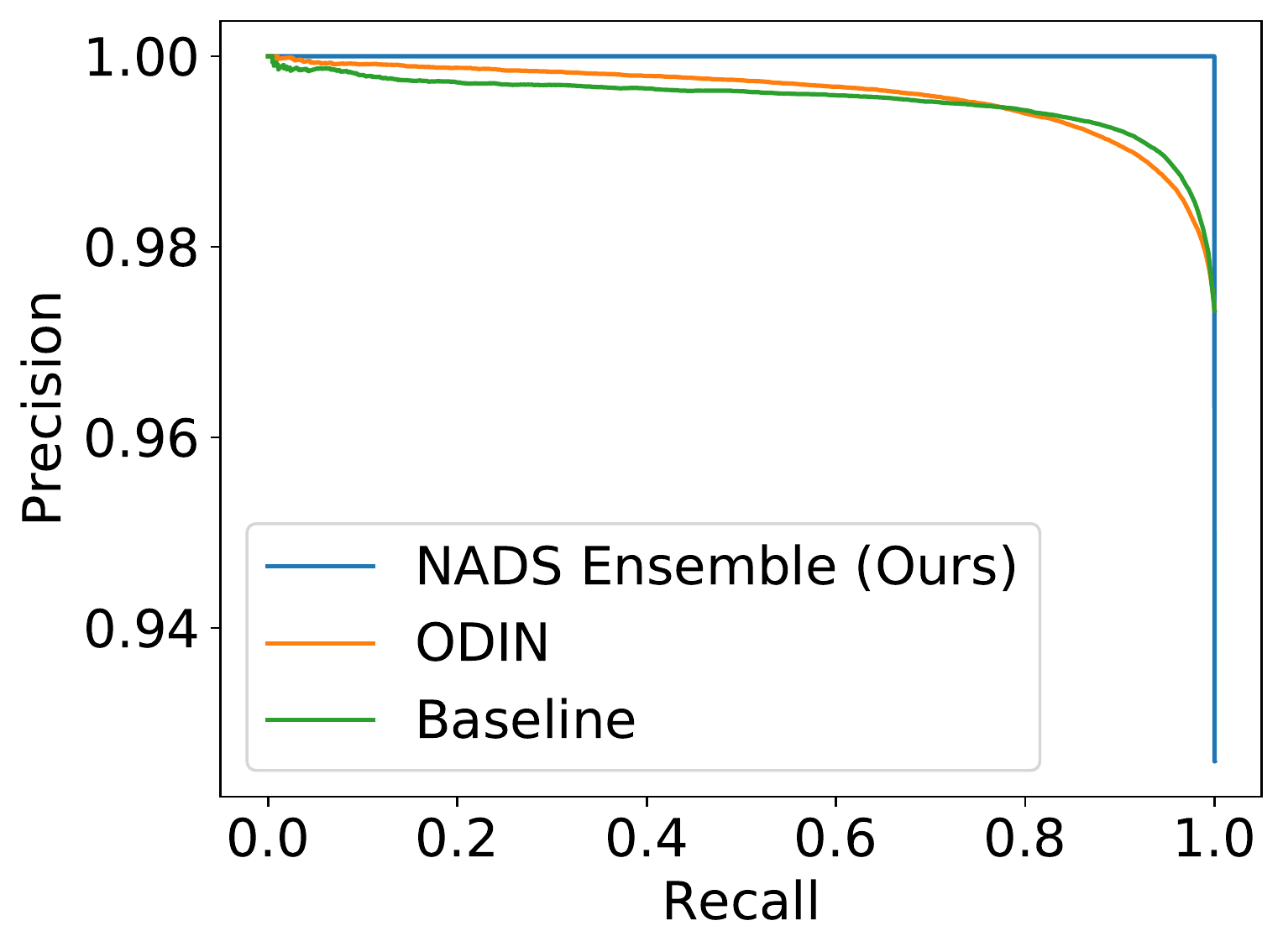}
}
\subfigure[SVHN]{
\includegraphics[width=0.23\textwidth]{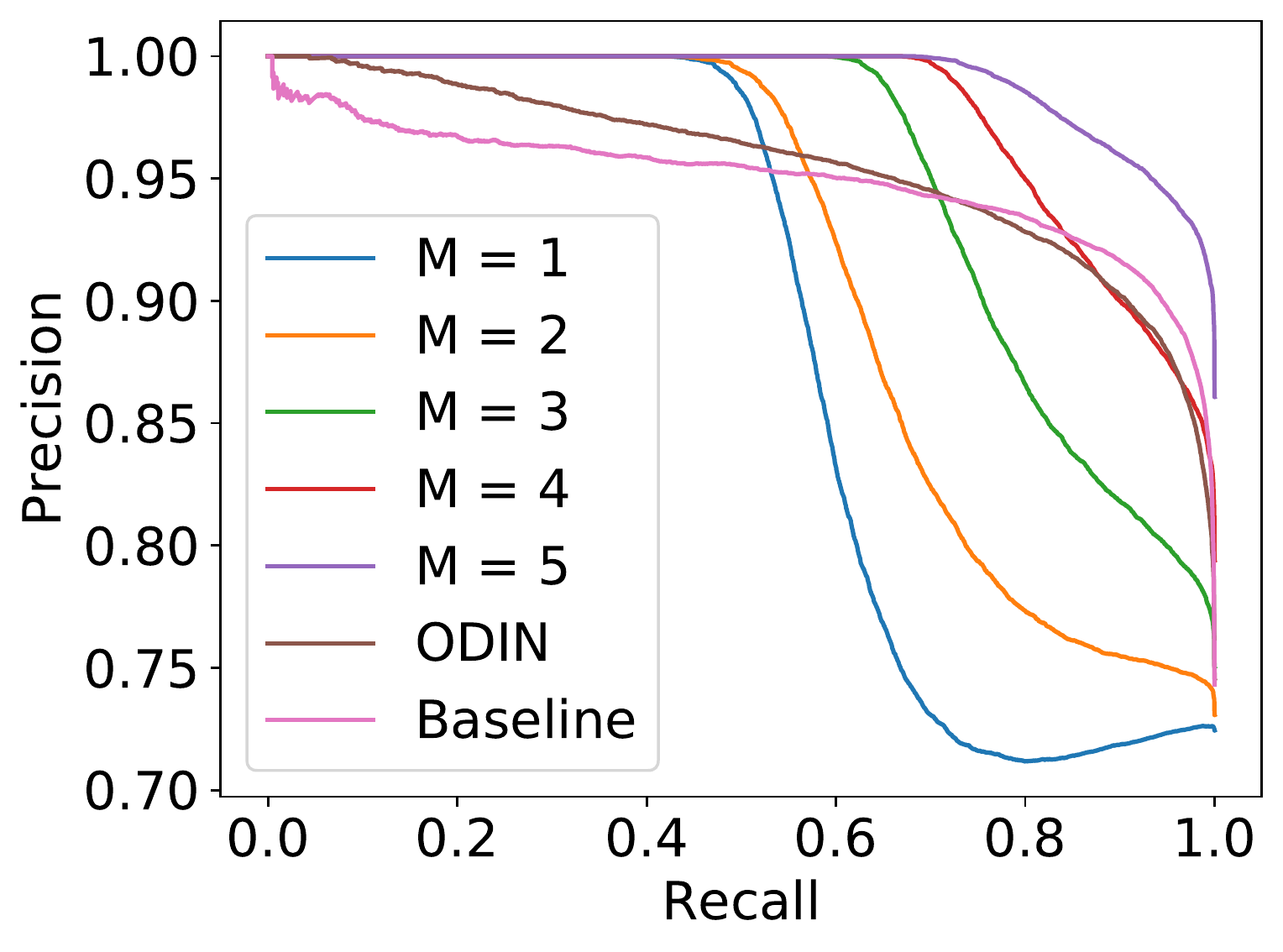}
}
\caption{ROC and PR curve comparison of methods trained on CIFAR-10}
\label{roc_cifar_append}
\end{figure}

\begin{figure}[h!]
\centering
\subfigure{
\includegraphics[width=0.3\textwidth]{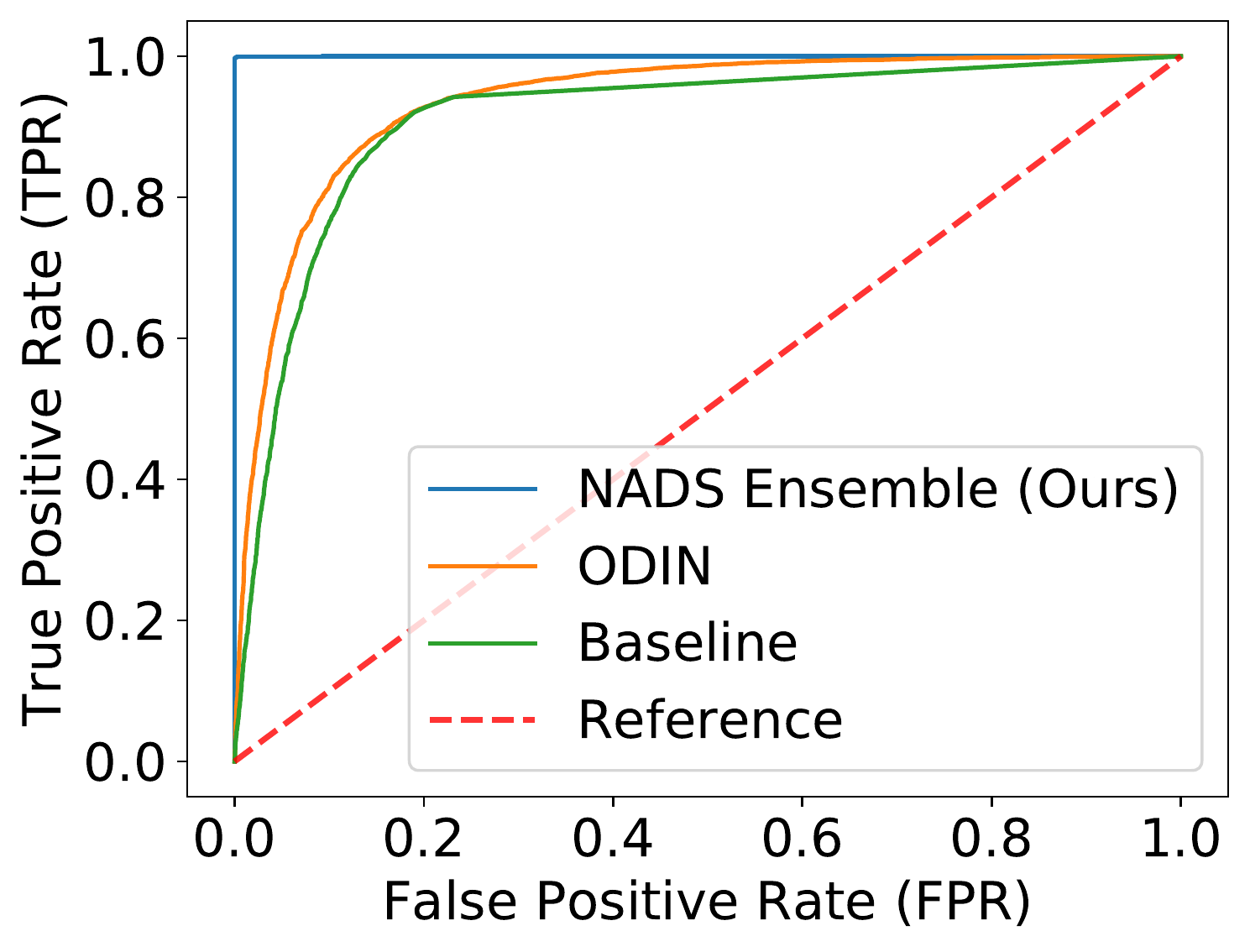}
}
\subfigure{
\includegraphics[width=0.3\textwidth]{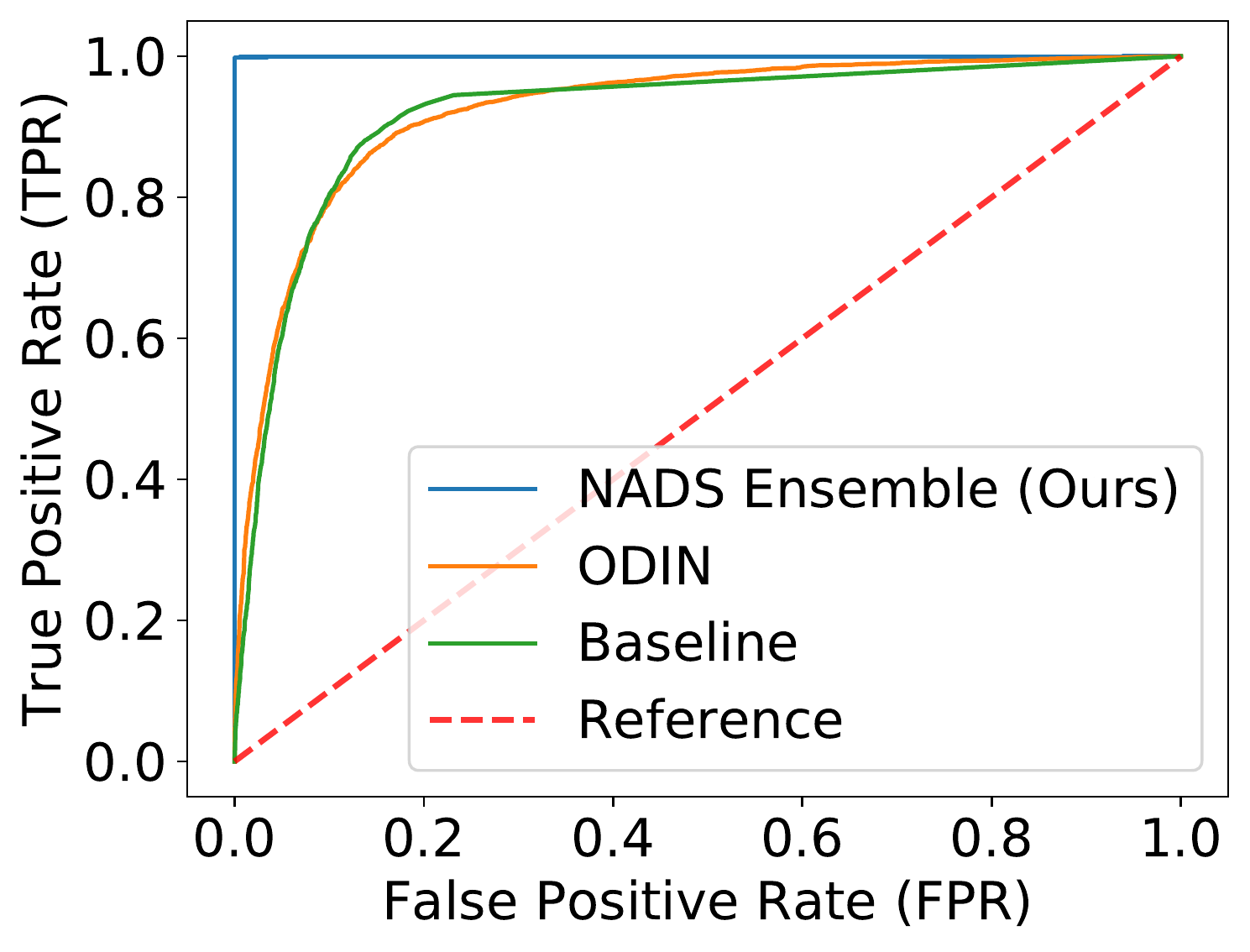}
}
\subfigure{
\includegraphics[width=0.3\textwidth]{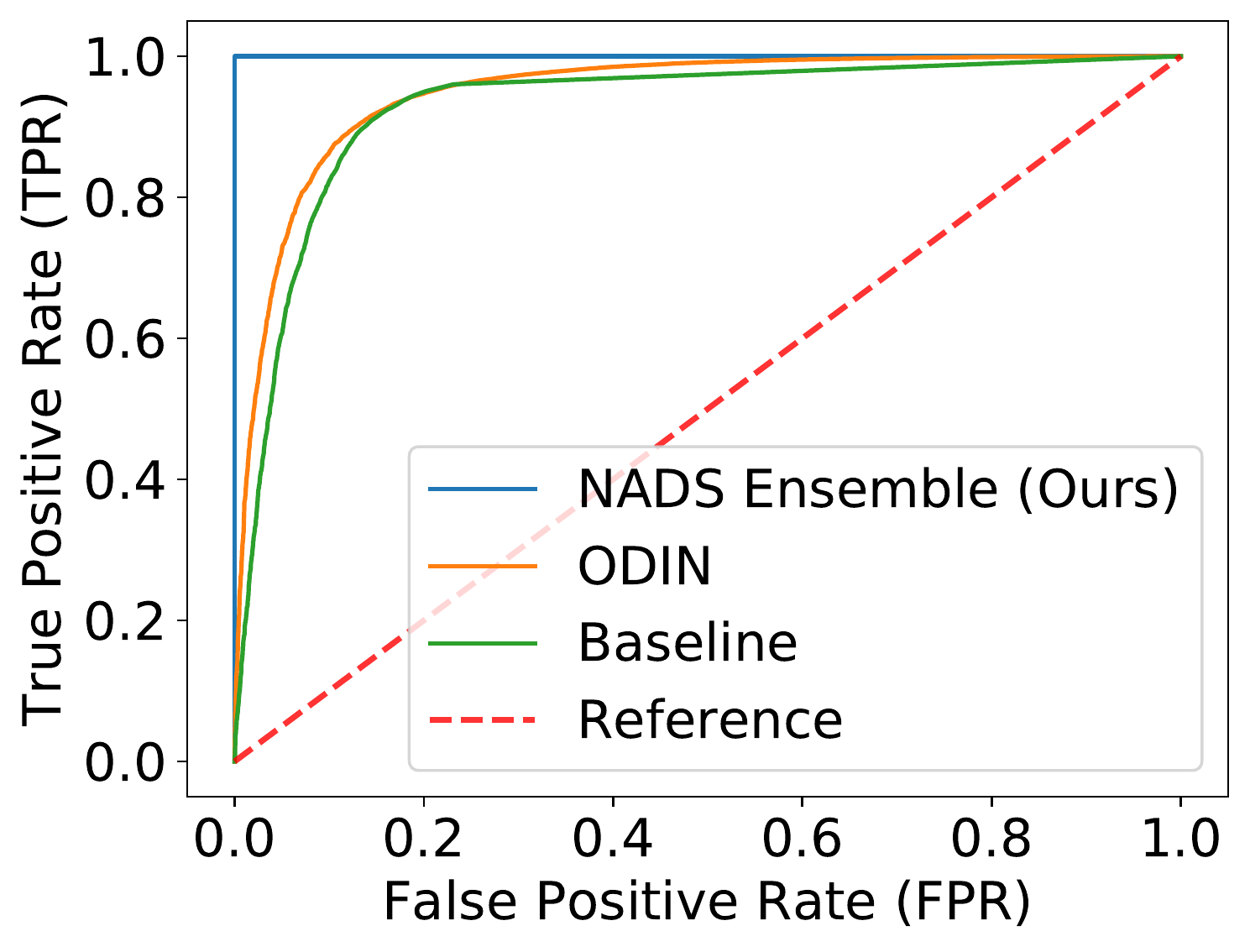}
}
\setcounter{subfigure}{0}
\subfigure[LSUN]{
\includegraphics[width=0.3\textwidth]{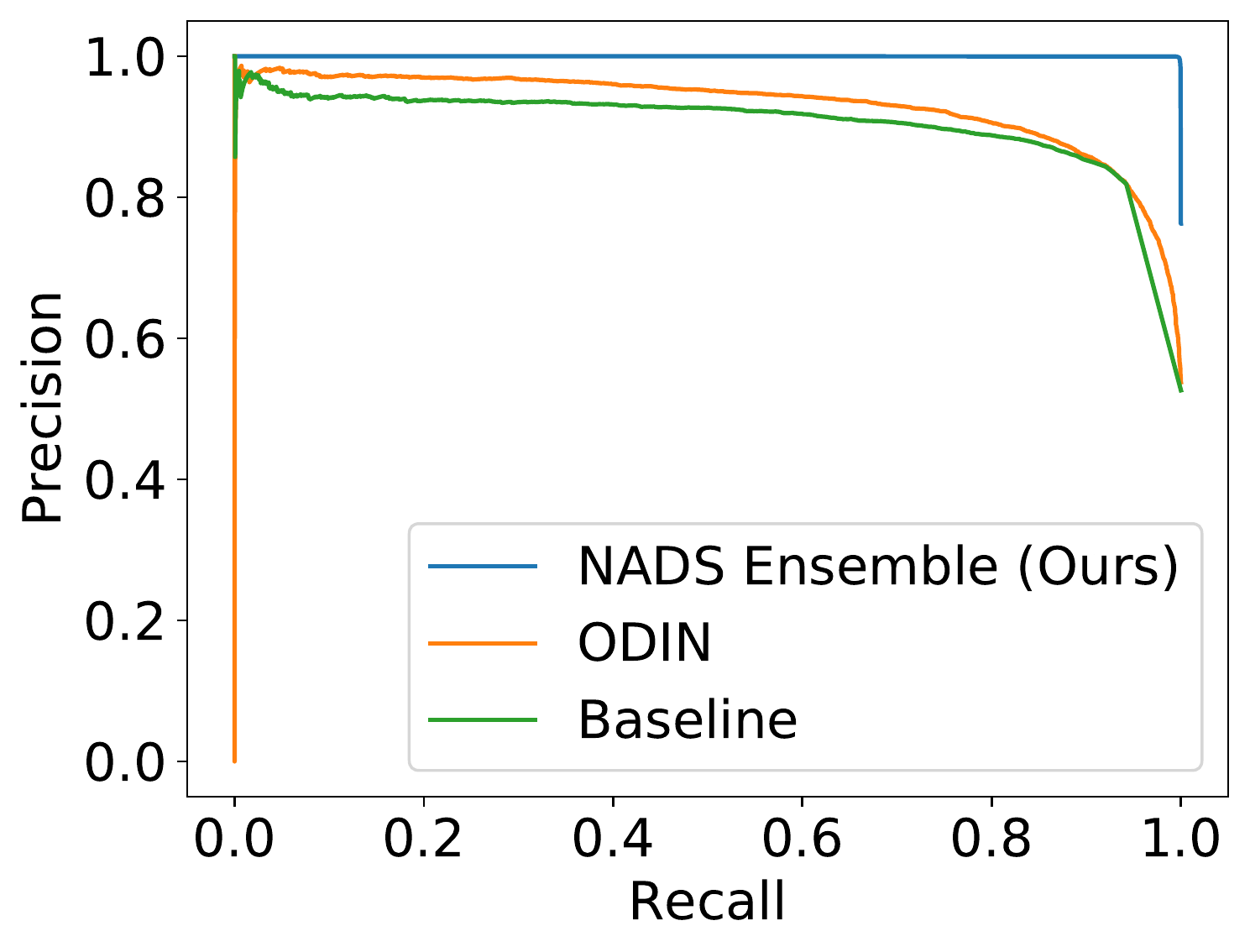}
}
\subfigure[Texture]{
\includegraphics[width=0.3\textwidth]{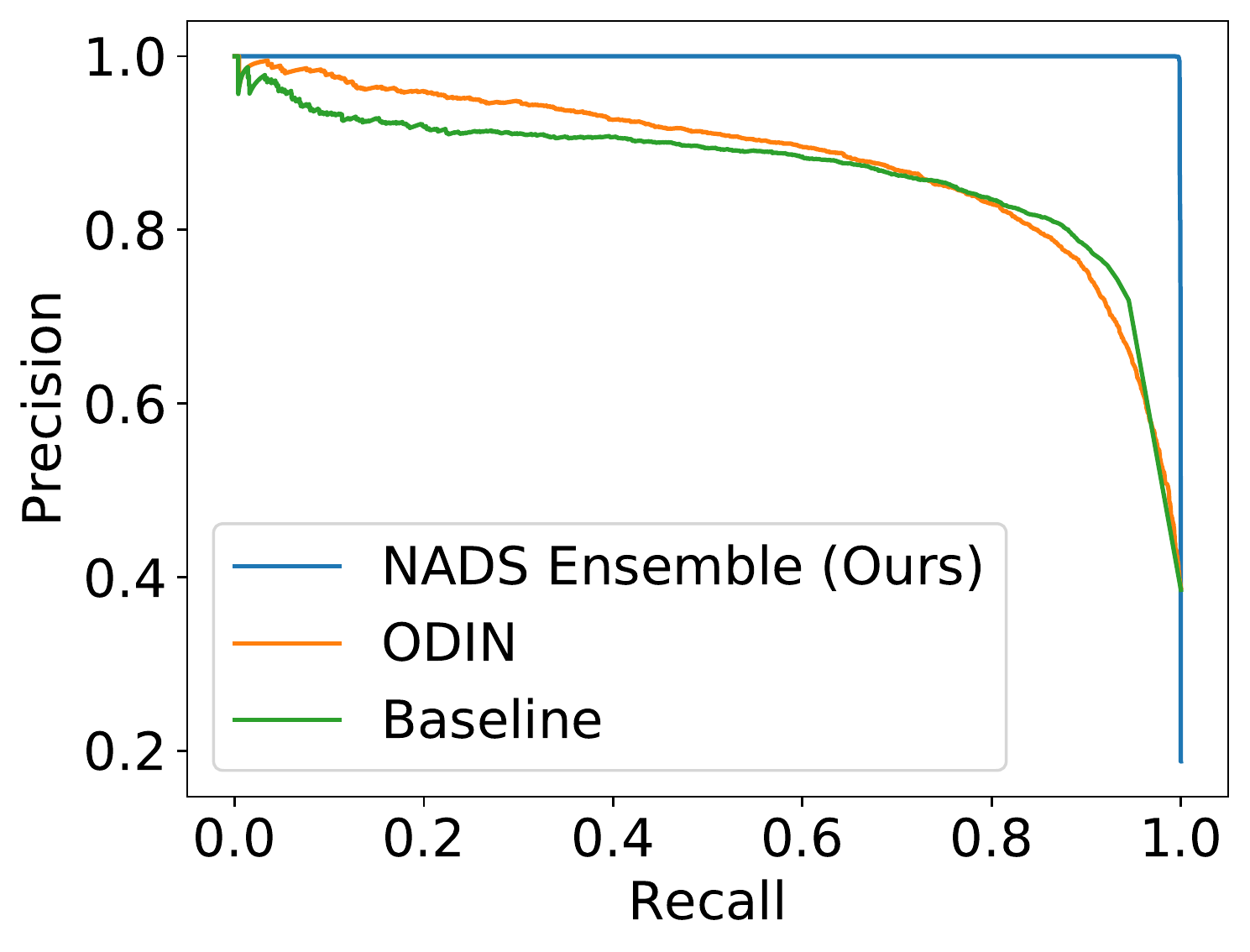}
}
\subfigure[Places]{
\includegraphics[width=0.3\textwidth]{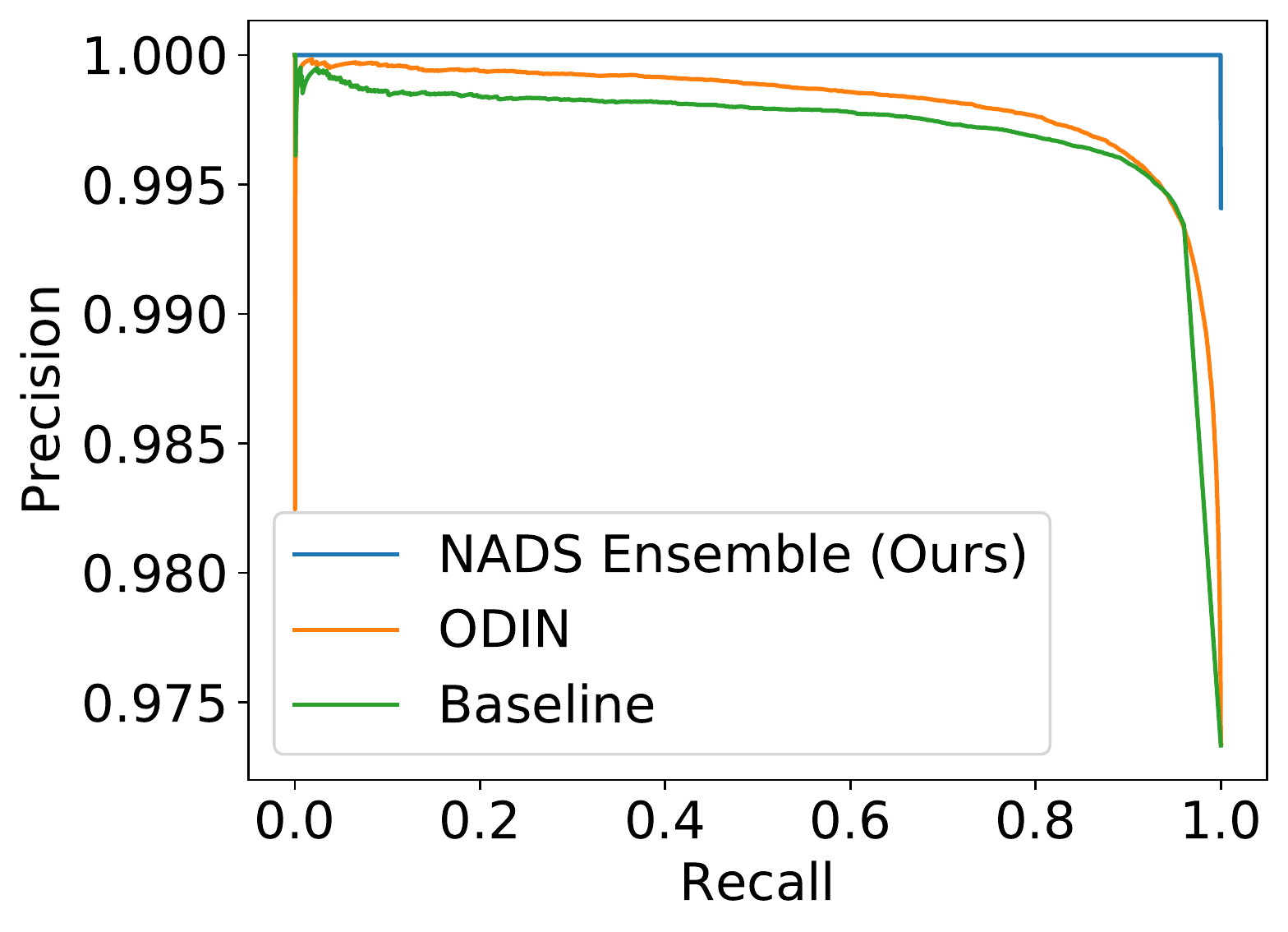}
}
\caption{ROC and PR curve comparison of methods trained on SVHN}
\label{roc_svhn_append}
\end{figure}

\begin{figure}[h!]
\centering
\subfigure{
\includegraphics[width=0.35\textwidth]{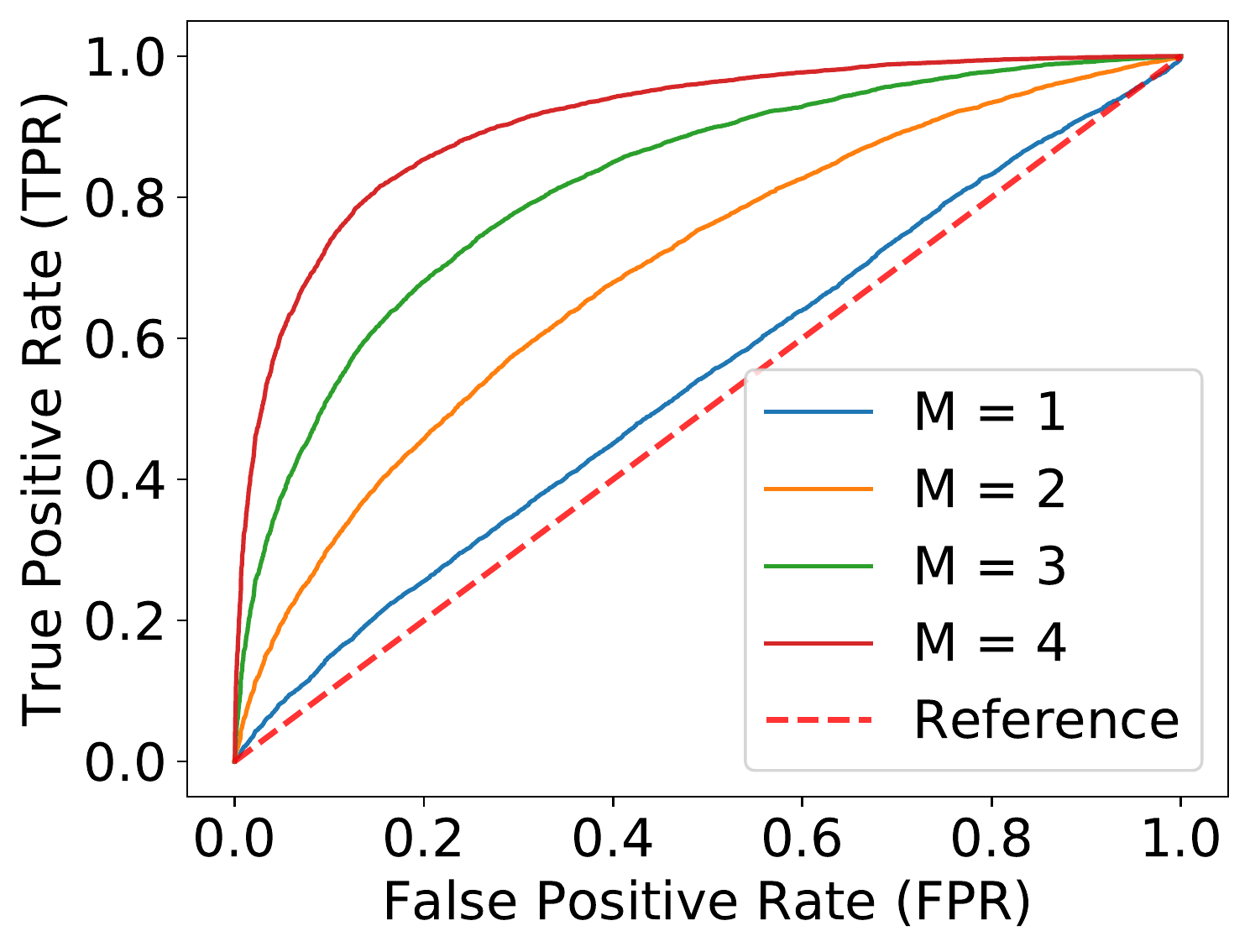}
}
\subfigure{
\includegraphics[width=0.35\textwidth]{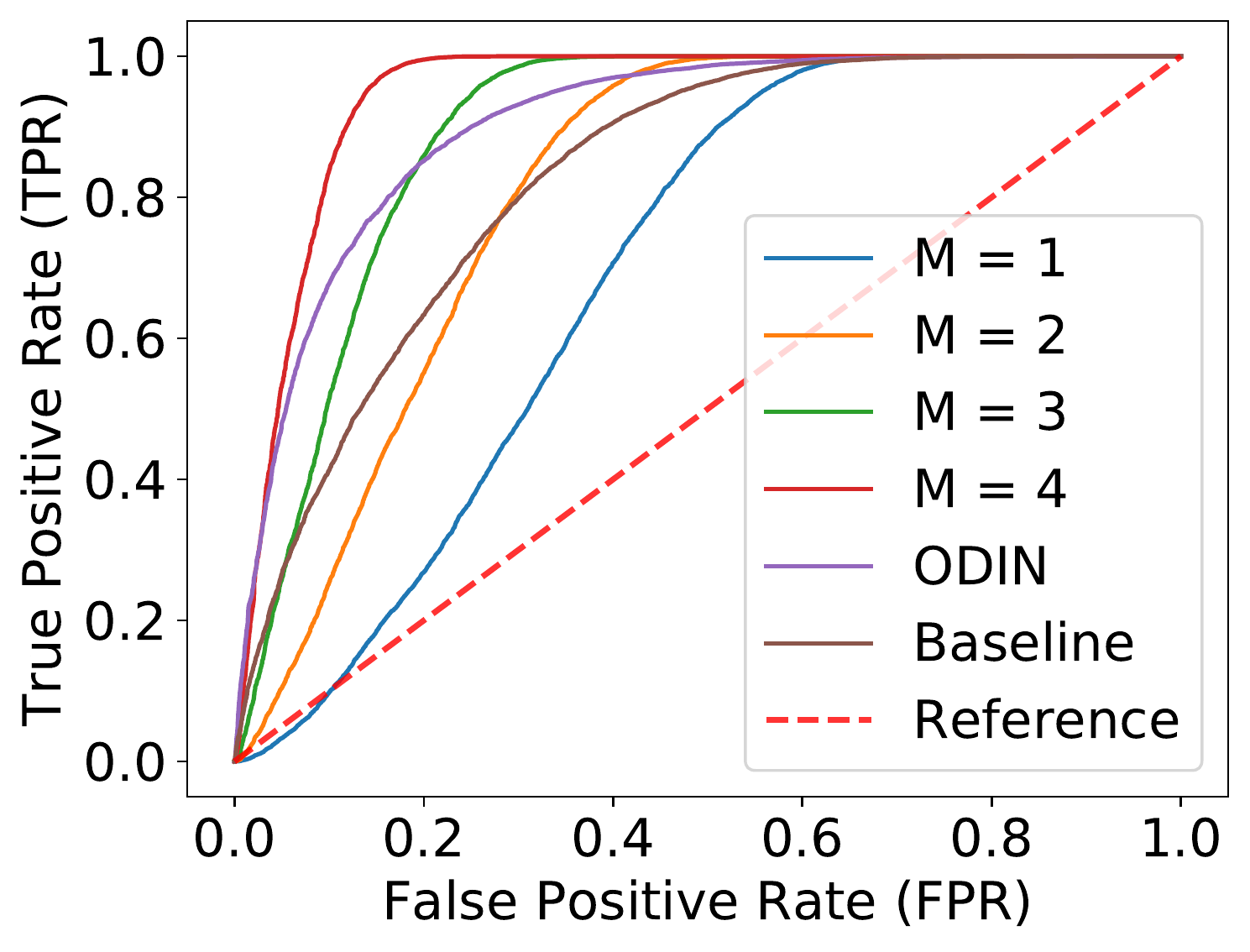}
}
\setcounter{subfigure}{0}
\subfigure[LSUN]{
\includegraphics[width=0.35\textwidth]{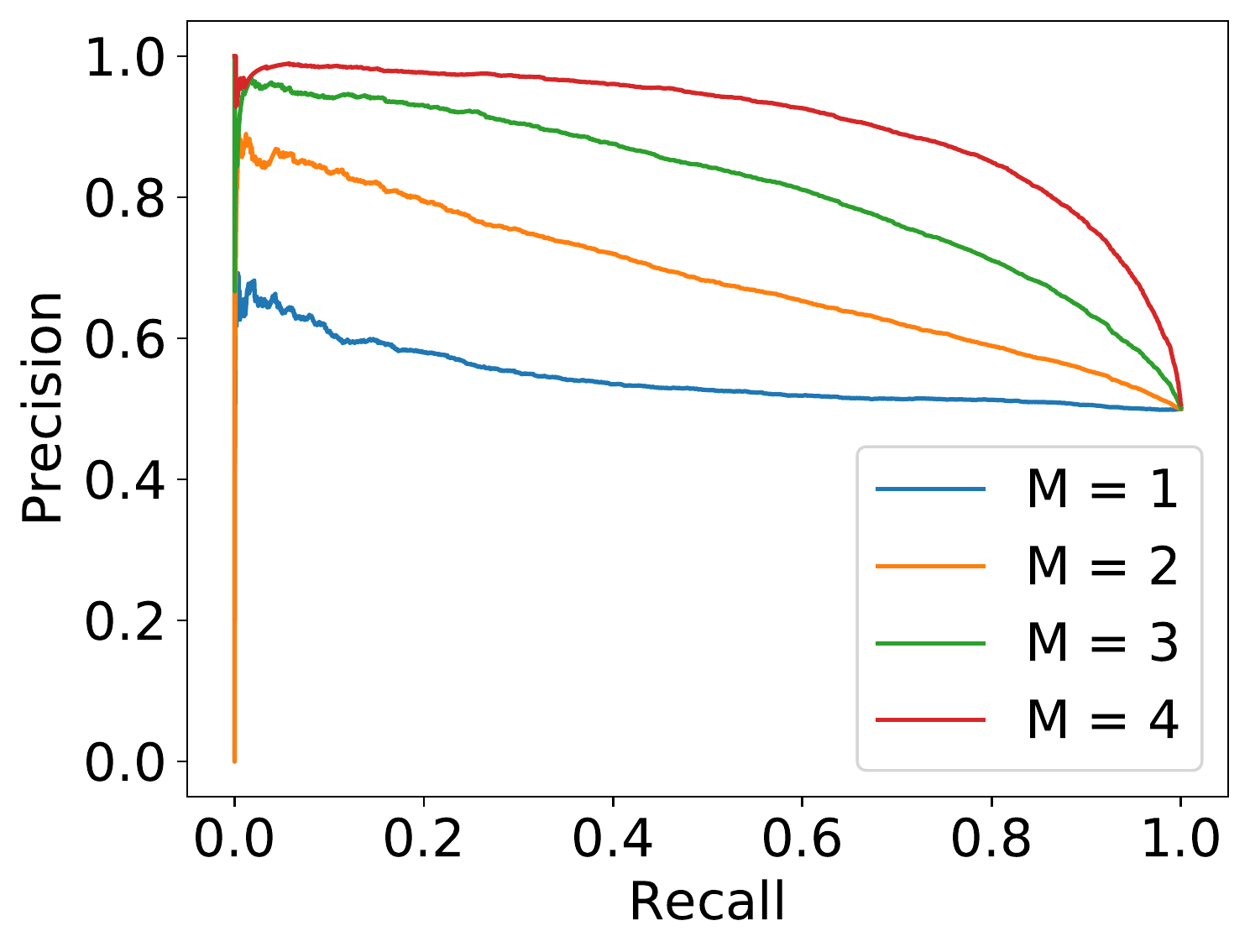}
}
\subfigure[SVHN]{
\includegraphics[width=0.35\textwidth]{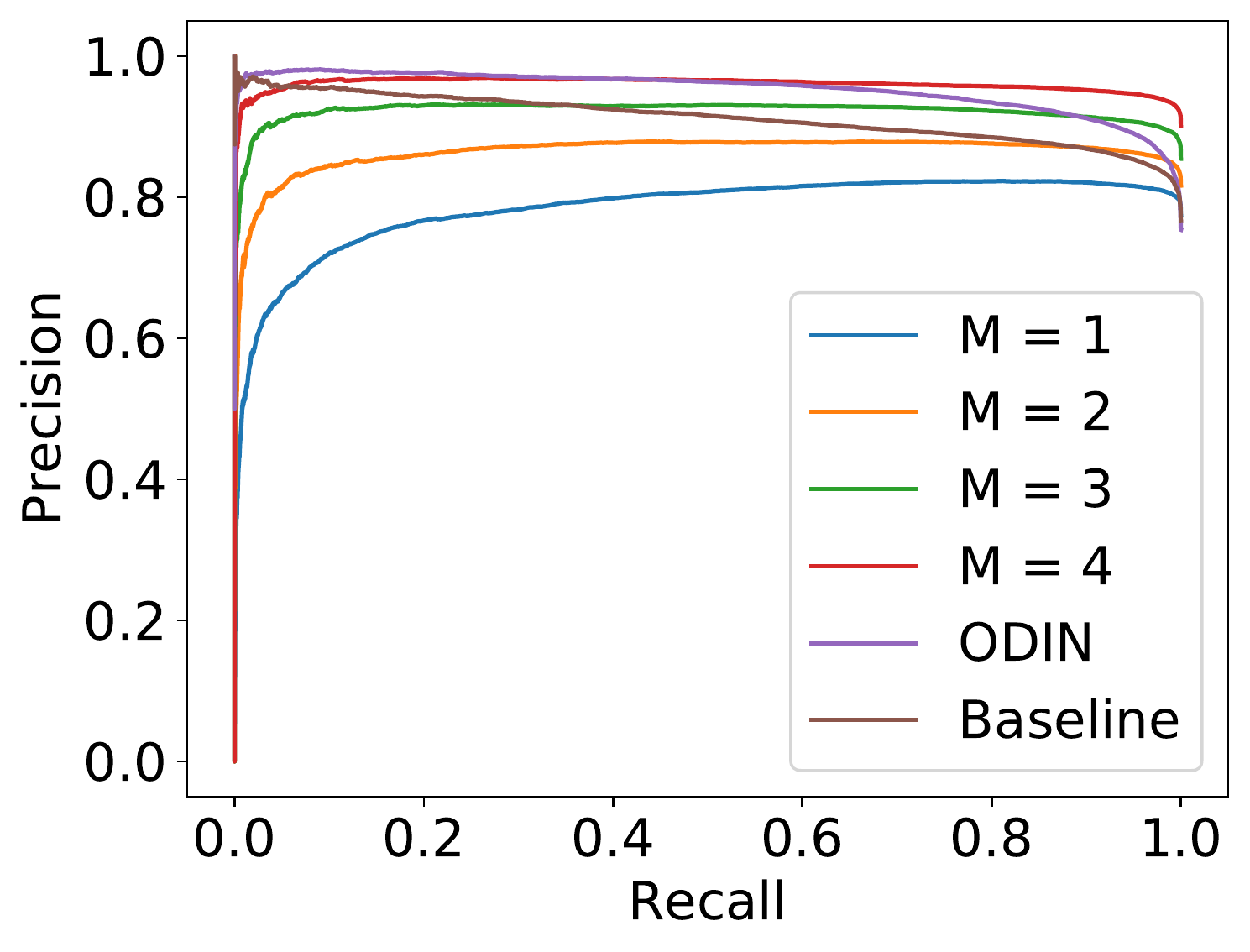}
}
\caption{ROC and PR curve comparison of methods trained on CIFAR-100}
\label{roc_cifar100_append}
\end{figure}

\begin{figure}[h!]
\centering
\subfigure{
\includegraphics[width=0.35\textwidth]{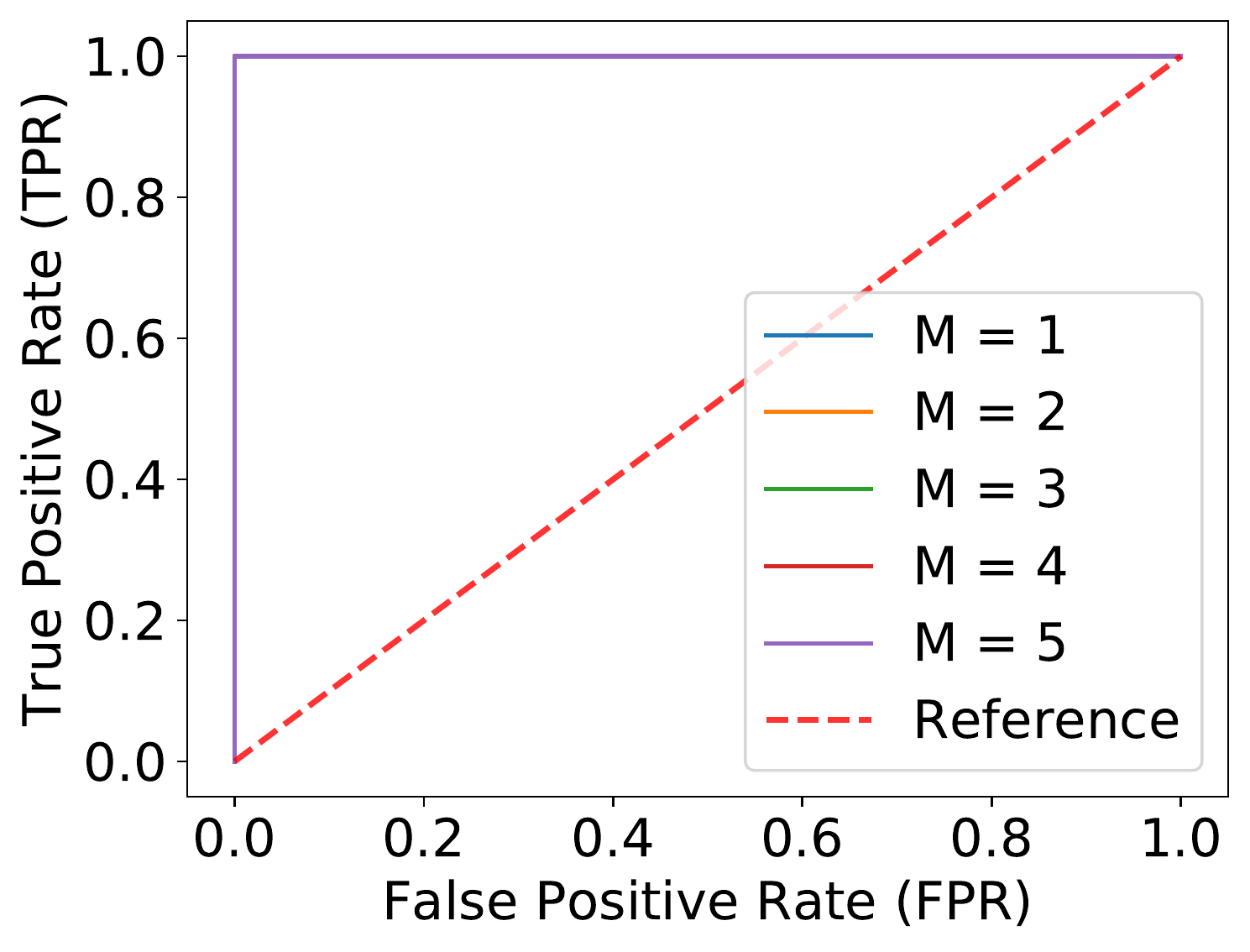}
}
\subfigure{
\includegraphics[width=0.35\textwidth]{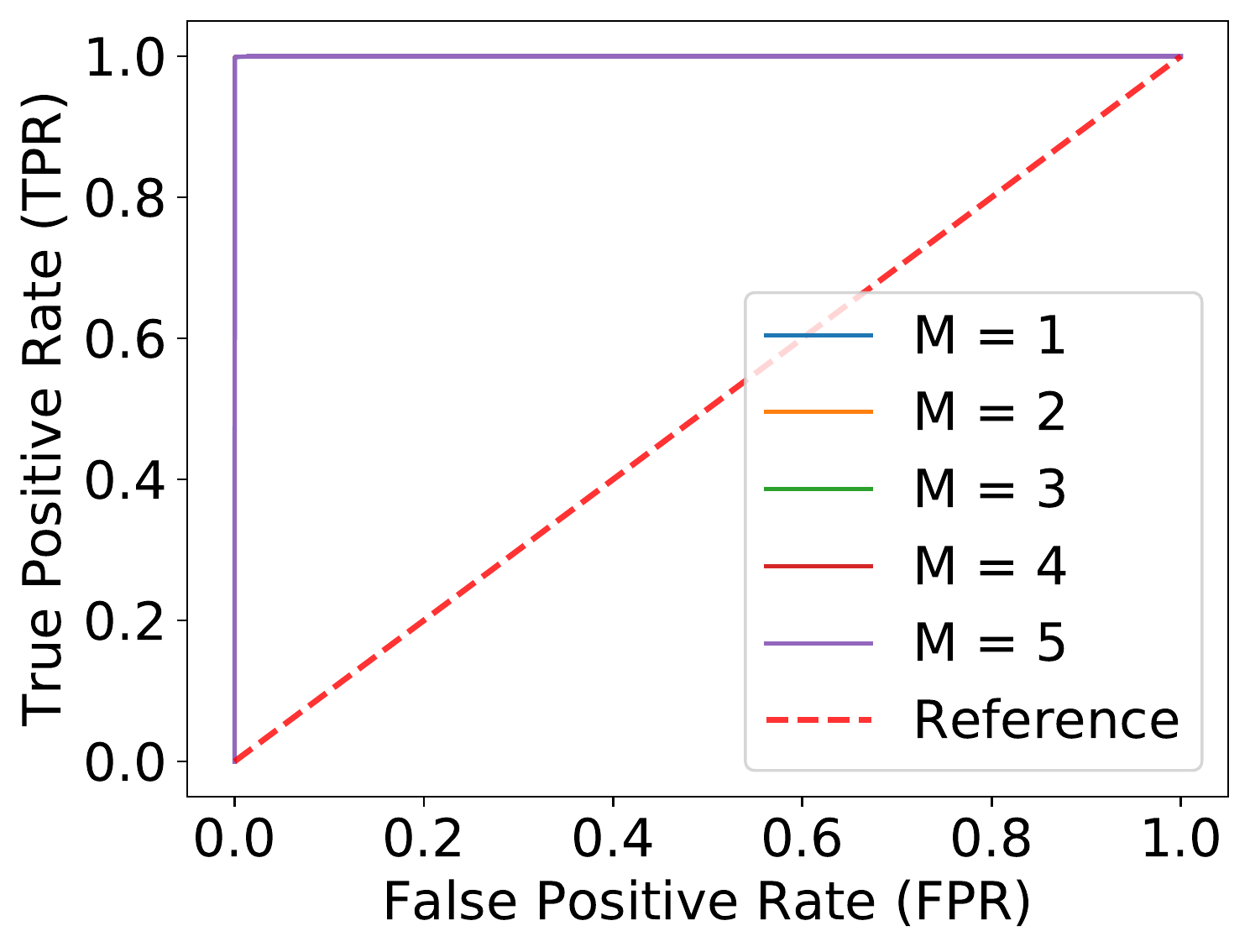}
}
\setcounter{subfigure}{0}
\subfigure[NotMNIST]{
\includegraphics[width=0.35\textwidth]{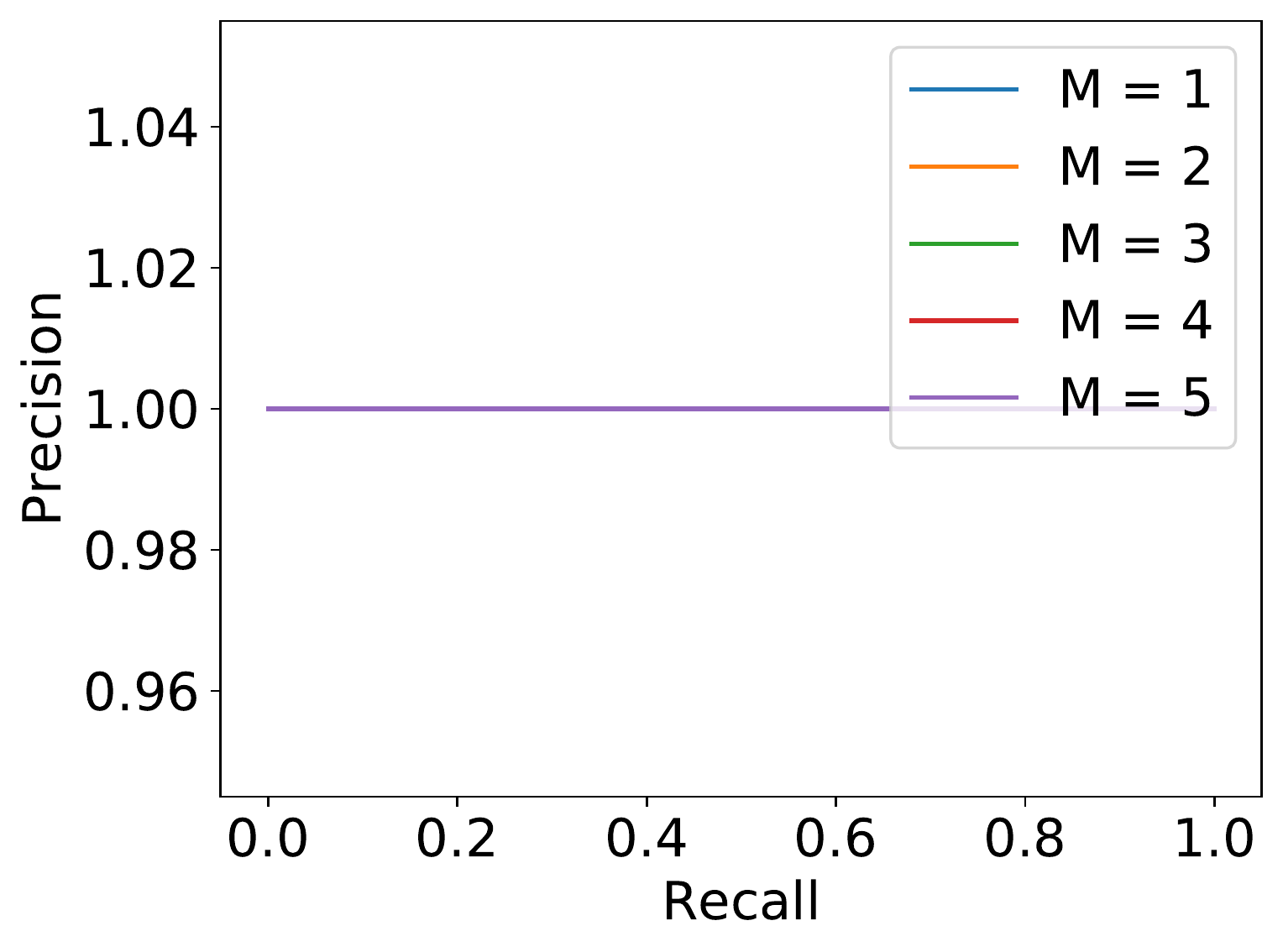}
}
\subfigure[FMNIST]{
\includegraphics[width=0.35\textwidth]{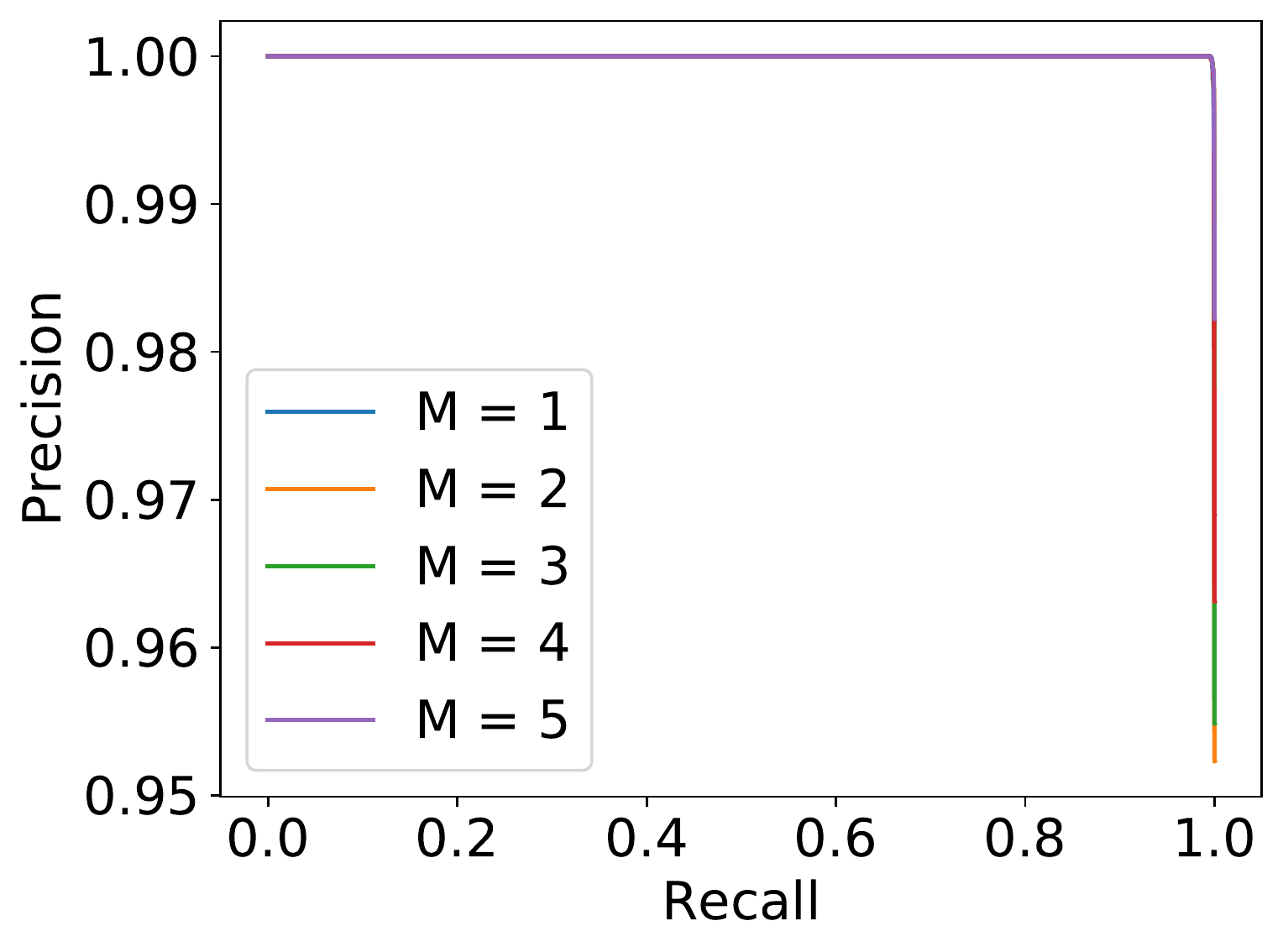}
}
\caption{ROC and PR curve comparison of methods trained on MNIST}
\label{roc_mnist_append}
\end{figure}

\FloatBarrier




\end{document}